\pdfoutput=1
\documentclass{article}
\usepackage{iclr2027_conference,times}
\iclrfinalcopy
\usepackage[utf8]{inputenc}
\usepackage[T1]{fontenc}
\usepackage{amsmath,amsfonts,amssymb,pifont}
\usepackage{graphicx,booktabs,multirow,array,longtable,colortbl}
\usepackage{algorithm,algpseudocode}
\usepackage[font=small,skip=6pt]{caption}
\usepackage{placeins,needspace}
\usepackage{xcolor,microtype,xspace}
\usepackage{hyperref,url}
\definecolor{teacherblue}{HTML}{0072B2}
\definecolor{criticorange}{HTML}{D55E00}
\definecolor{methodgreen}{HTML}{009E73}
\definecolor{ourgray}{gray}{0.90}
\newcommand{\sg}{\operatorname{sg}}
\newcommand{\KL}{\operatorname{KL}}
\newcommand{\E}{\mathbb{E}}
\title{A Better Critic Matters for\\
Distribution Matching Distillation}
\author{Zongze Wu$^{1*}$\quad Xianbang Wang$^{2*\dagger}$\quad Zhengyang Geng$^{1}$\quad Mingdeng Cao$^{1}$\\
\textbf{Zhoutong Zhang$^{1}$\quad Richard Zhang$^{1}$\quad Eli Shechtman$^{1}$\quad Ruiming Cao$^{1}$}\\[4pt]
$^{1}$Adobe\qquad $^{2}$MIT\\
{\small $^{*}$Equal contribution, listing order is random \qquad $^{\dagger}$Work done during an internship at Adobe } \\
{\small Project page \& code: \url{https://peppaking8.github.io/critic-matters/}}}
\hypersetup{pdftitle={A Better Critic Matters for Distribution Matching Distillation},
  pdfauthor={Zongze Wu, Xianbang Wang, Zhengyang Geng, Mingdeng Cao, Zhoutong Zhang, Richard Zhang, Eli Shechtman, Ruiming Cao},pdfsubject={Distribution matching distillation},
  colorlinks=true,citecolor=teacherblue,linkcolor=teacherblue,urlcolor=teacherblue}
\begin{document}
\addtocontents{toc}{\protect\setcounter{tocdepth}{-1}}
\raggedbottom
\maketitle
\lhead{Preprint}
\begin{abstract}

Distribution Matching Distillation (DMD) trains a few-step student using the difference between a teacher's score and an online critic's estimate of the student distribution's score. We investigate what the critic needs, in \emph{capacity} and in \emph{information}, to provide a reliable training signal. On the capacity side, controlled experiments across seven SiT model sizes on ImageNet show that increasing teacher size while keeping a small critic fixed can degrade student quality and cause training collapse. Scaling the critic alongside the teacher, a strategy we call \emph{joint scaling}, prevents collapse in our experiments and allows the student to benefit from larger teachers. On the information side, rather than forcing the critic to model a step-agnostic mixture of student outputs, we introduce a step-aware critic (STAC) that conditions on the student's generation step to estimate step-specific scores. Together, joint scaling and STAC consistently improve few-step distillation across class-conditional ImageNet generation, text-to-image generation with SDXL and FLUX.2, and text-to-video generation with Wan2.1.

\end{abstract}

\section{Introduction}
\label{sec:intro}
\begingroup
\setlength{\parskip}{3pt}

Diffusion and flow matching models produce high-quality images and videos but need many denoising steps for each generation, which makes distilling them into few-step students central to their deployment.
Distribution Matching Distillation (DMD) \citep{yin2024improveddistributionmatchingdistillation,yin2024onestepdiffusiondistributionmatching} approaches this problem by matching distributions of the student and the teacher.
Each student sample is re-noised, and the difference between the teacher and student-distribution scores provides a gradient for updating the student.
Since the student distribution is unknown, DMD estimates it with a \emph{critic} network.
This teacher-critic recipe is now widely adopted in few-step image generation
\citep{imageteam2026zimageefficientimagegeneration,zhao2026qwenimage20technicalreport,liu2026ernieimagetechnicalreport},
real-time autoregressive video
\citep{yin2025slowbidirectionalfastautoregressive,huang2025selfforcingbridgingtraintest,yang2025longliverealtimeinteractivelong},
interactive world models
\citep{he2026matrixgame20opensourcerealtime,sun2026worldplaylongtermgeometricconsistency,robbyantteam2026advancingopensourceworldmodels},
and joint audio-video generation
\citep{su2026omniforcingunleashingrealtimejoint,zhang2026vidus1realtimeinteractive}.

During DMD training, the critic must track the student's evolving distribution while providing the score estimates used to update the student.
Prior work improves critic tracking through a two-time-scale update rule \citep{yin2024improveddistributionmatchingdistillation}.
The critic is typically initialized from a pretrained diffusion model.
When pretrained models of different sizes are available, it remains unclear which critic size is most effective for a given student model.
In practice, compute and memory constraints further complicate this choice.
For example, the DMD recipe of Self Forcing pairs a 14B teacher with a 1.3B critic and a 1.3B student \citep{huang2025selfforcingbridgingtraintest}, a configuration also used by subsequent methods \citep{zhu2026causalforcingautoregressivediffusion,feng2026oneforcingstableonestepautoregressive,yu2026videomiraiautoregressivevideodiffusion}, while other studies pair a large teacher with a teacher-sized critic \citep{zheng2026causalrcmunifiedteacherforcingselfforcing,yang2026incontextforcinguncoveringcontext}.
These differing choices motivate us to revisit critic design: \emph{what does the critic need, in capacity and in information, for DMD?}

First, we study teacher and critic \emph{scaling} while holding the student architecture, the number of function evaluations (NFE) at inference, and the number of training iterations fixed. We extend the SiT \citep{ma2024sitexploringflowdiffusionbased} family to seven pretrained model sizes, from S to H, and systematically evaluate teacher--critic size combinations on ImageNet~\citep{russakovsky2015imagenetlargescalevisual} at $256\times256$ resolution. We also compare generation quality against training FLOPs to assess the computational efficiency of different scaling strategies.
Figure~\ref{fig:scaling_paths} illustrates the resulting scaling behavior.
With a small critic, increasing teacher size can initially improve student quality, but further scaling can degrade performance or lead to training collapse.
Scaling only the critic helps little.
In contrast, increasing both teacher and critic capacity together, which we call \emph{joint scaling}, enables further improvements with larger teachers (Figure~\ref{fig:compute_pareto}).
Across this study, all observed training collapses occur with a critic \emph{smaller} than its teacher.

Beyond network capacity, we further improve the critic's conditioning to facilitate generator learning.
In multi-step DMD, critic training pools student predictions from different generation steps indexed by $k$.
These predictions can follow different distributions.
However, the critic is conditioned on the score-evaluation level $t$ and the class or text prompt, but not on $k$.
It therefore estimates the score of their mixture. We introduce a \emph{step-aware critic} (STAC), which additionally conditions on $k$---represented by the corresponding schedule level $r_k$ in our implementation---to estimate the score of each step-specific student distribution (Figure~\ref{fig:teaser}).

We conduct comprehensive experiments spanning class-conditional image generation with SiT on ImageNet, text-to-image generation with FLUX.2~\citep{blackforestlabs2025flux2} and SDXL~\citep{podell2023sdxlimprovinglatentdiffusion}, and text-to-video generation with Wan2.1~\citep{wan2025wanopenadvancedlargescale}.
At $1024\times1024$, a fixed FLUX.2 4B student improves from 10.58/81.58 to 12.32/85.28 in HPSv3~\citep{ma2025hpsv3widespectrumhumanpreference}/DPG-Bench~\citep{hu2024ella} when scaling from a 4B teacher--critic pair to a 32B pair with STAC (Section~\ref{sec:external}).
We also examine alternative critic designs and teacher scaling in other distillation methods in Appendices~\ref{app:critic_share} and~\ref{app:other_methods}, respectively.
\begin{figure}[t]
\centering
\includegraphics[width=.98\linewidth]{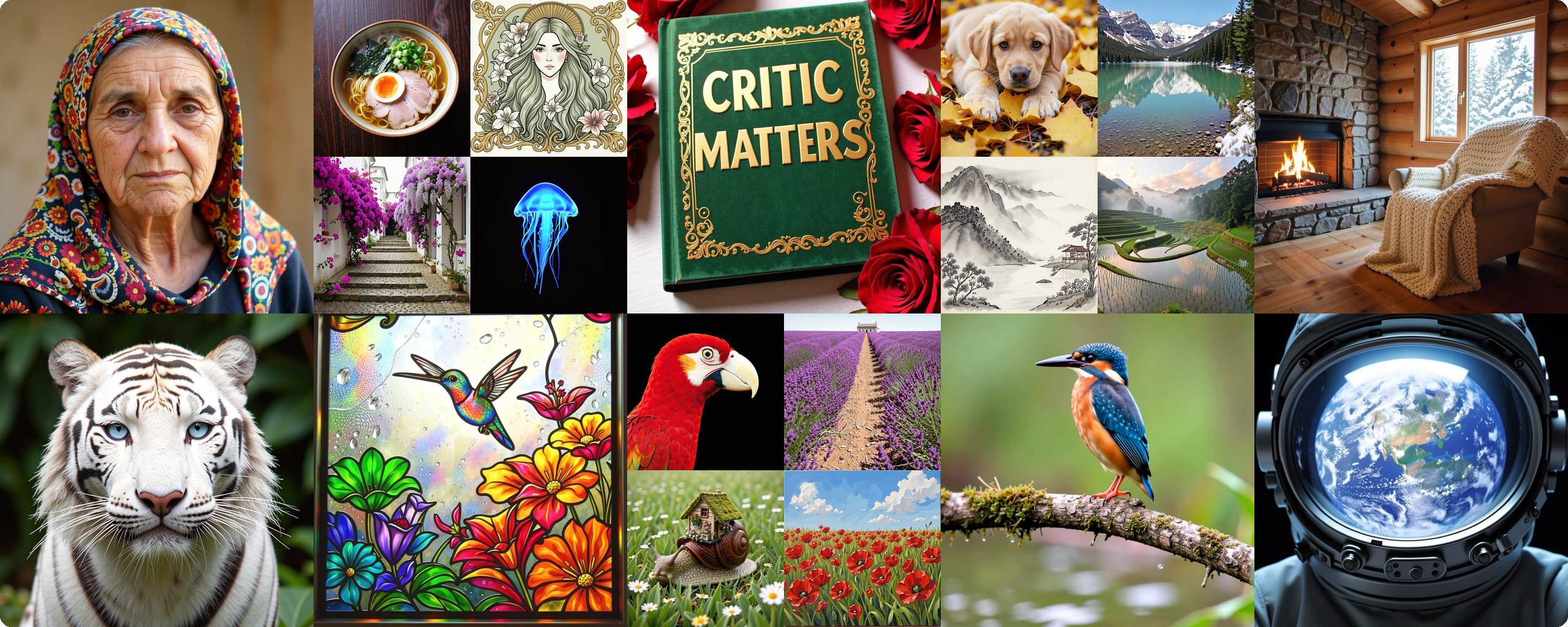}\\[2pt]
\includegraphics[width=.98\linewidth]{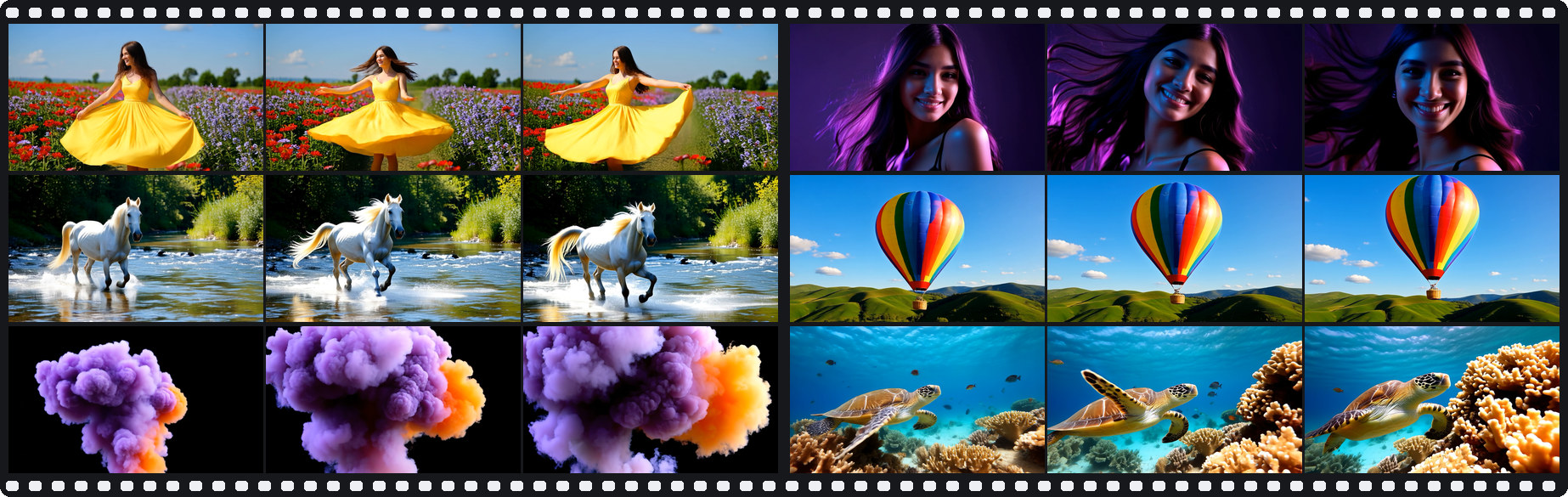}
\caption{\textbf{Few-step students distilled with a critic as large as the teacher.} Top, samples of a 4-step FLUX.2 [klein] 4B student distilled with a 32B teacher and a 32B critic with STAC. Bottom, three frames each of six videos from a 4-step Wan2.1 1.3B student distilled with a 14B teacher and a 14B critic with STAC. (Settings: FLUX.2 $1024\times1024$, Wan2.1 480p.)}
\label{fig:teaser_visual}
\end{figure}
Our contributions are summarized below:
\begin{itemize}
\setlength{\topsep}{2pt}
\setlength{\itemsep}{0pt}
\setlength{\parsep}{0pt}
\item We systematically study teacher--critic scaling in DMD using seven SiT model sizes on ImageNet, examining both generation quality and training compute. We demonstrate the benefits of jointly scaling the teacher and critic at a fixed student model capacity.
\item We introduce a \emph{step-aware critic} (STAC), which conditions on the current student generation step to estimate the score of the corresponding step-specific distribution.
\item We validate our findings across image and video generation with SiT, FLUX.2, SDXL, and Wan2.1.
\end{itemize}
\FloatBarrier

\begin{figure}[t]
\centering
\includegraphics[width=\linewidth]{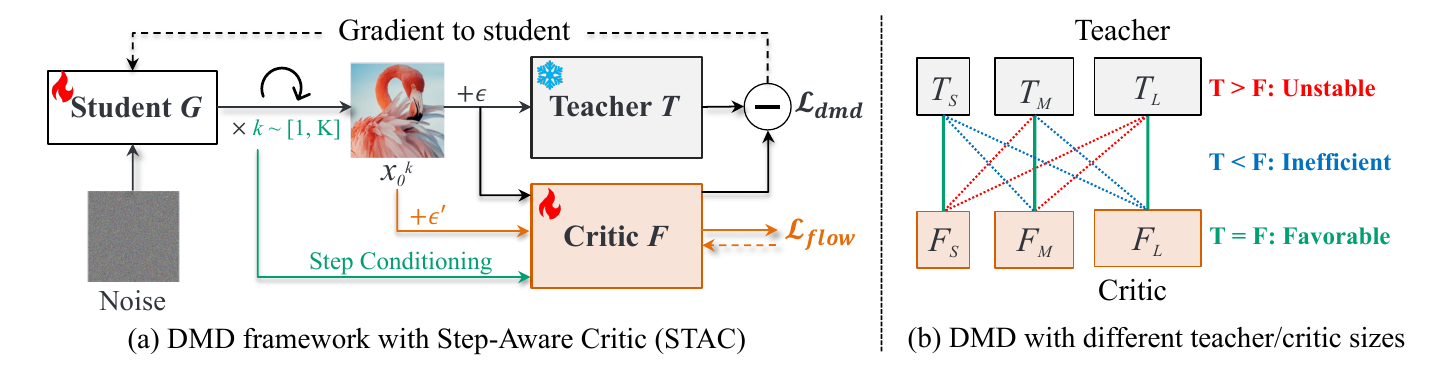}
\caption{\textbf{Overview.} We study (a) step conditioning and (b) joint teacher--critic scaling. In (b), $T$ and $F$ denote teacher/critic capacity: $T>F$ can be unstable, $T<F$ inefficient, and $T=F$ favorable here.}
\label{fig:teaser}
\end{figure}
\endgroup

\section{Related Work}
\label{sec:related}

\paragraph{Distribution matching distillation.}
Distribution-matching approaches accelerate diffusion generators using gradients derived from teacher (real) and student (fake) score estimates. DMD, SwiftBrush, and Diff-Instruct use this structure to train amortized generators whose output distributions approach the teacher's~\citep{yin2024onestepdiffusiondistributionmatching,nguyen2024swiftbrushonesteptexttoimagediffusion,luo2024diffinstructuniversalapproachtransferring}. Related two-score constructions include ProlificDreamer's VSD, which optimizes a particle-based distribution over 3D representations, and Score identity Distillation, which uses a Fisher-divergence objective~\citep{wang2023prolificdreamerhighfidelitydiversetextto3d,zhou2025fewstepdiffusionscoreidentity}. DMD2 extends DMD to few-step generation~\citep{yin2024improveddistributionmatchingdistillation}. A growing body of follow-up work further improves this framework from various perspectives~\citep{liu2025decoupleddmdcfgaugmentation,bai2026optimizingfewstepgenerationadaptive,huang2026reinforcingfewstepgeneratorsrewardtilted,jiang2026distributionmatchingdistillationmeets,kim2026distributionmatchingdistillationfake,lu2025adversarialdistributionmatchingdiffusion,ge2026senseflowscalingdistributionmatching,chen2026dataforcingdistillationrestoringdiversity,zhao2026maskforcingimprovingautoregressive,zheng2026largescalediffusiondistillation}. DMD is now widely used across generation tasks, including text-to-image~\citep{imageteam2026zimageefficientimagegeneration,zhao2026qwenimage20technicalreport,liu2026ernieimagetechnicalreport}, text-to-video and autoregressive video~\citep{yin2025slowbidirectionalfastautoregressive,huang2025selfforcingbridgingtraintest,yang2025longliverealtimeinteractivelong}, and interactive world and audio--video models~\citep{he2026matrixgame20opensourcerealtime,su2026omniforcingunleashingrealtimejoint}. Consistency-distillation systems such as T2V-Turbo-v2 provide a complementary route to fast video generation~\citep{li2025t2vturbov2enhancingvideogeneration}. In this paper, we investigate the size and input conditioning of the DMD critic.

\paragraph{Few-step distribution modeling.}
A multi-step student substantially outperforms a 1-step one by refining its output over several steps, but each step generates from a different intermediate distribution, so few-step methods must decide how to represent these distributions. Within distribution matching, TDM partitions score-evaluation levels into intervals~\citep{luo2025learningfewstepdiffusionmodels}, while Phased DMD uses per-phase experts~\citep{fan2026phaseddmdfewstepdistribution}. Few-Step SiD instead deliberately matches the uniform mixture of outputs across generation steps with a shared fake-score network~\citep{zhou2025fewstepdiffusionscoreidentity}. CMD addresses an orthogonal mismatch in autoregressive video by matching real- and fake-score evaluations to the causal temporal-prefix context available during rollout~\citep{bandyopadhyay2026contextmatcheddistillationteachercausality}. Related acceleration methods expose temporal coordinates in different ways: Shortcut Models explicitly condition on the desired step size~\citep{frans2025stepdiffusionshortcutmodels}; Consistency Models condition on the current noise level and map to a fixed endpoint~\citep{song2023consistencymodels}; and Progressive Distillation repeatedly halves a fixed sampling schedule through trajectory targets, without a separate target-step input~\citep{salimans2022progressivedistillationfastsampling}. In our work, we bring generation-step conditioning into distribution matching: we model each per-step distribution separately with a single critic conditioned on the generation step (STAC), improving multi-step generation quality.

\section{The Critic Matters}

\label{sec:baseline}

We study critic design using a 2-D toy problem (in Appendix~\ref{app:toy}) and $256\times256$ class-conditional ImageNet generation with SiT \citep{ma2024sitexploringflowdiffusionbased}. Here, we focus on ImageNet.
We extend the SiT family to seven sizes, S, S-B, B, M, L, XL, and H, ranging from 33M to 951M parameters, and pretrain each model for 800k updates using a shared flow-matching recipe (Appendix~\ref{app:ladder}).
We use these pretrained base models for DMD with only the distribution matching objective and a common training protocol detailed in Appendix~\ref{app:training_query}. Appendix~\ref{sec:background} reviews DMD and our notation.

\subsection{Scaling up critic capacity}
\label{sec:teacher_student}
\label{sec:scaling}

\begin{figure}[!htb]
\centering
\includegraphics[width=0.32\linewidth]{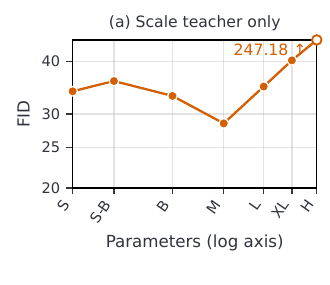}\hfill
\includegraphics[width=0.32\linewidth]{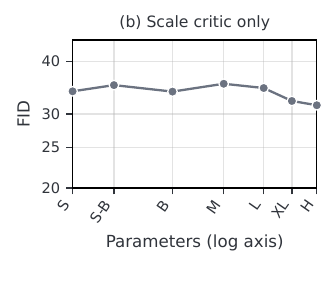}\hfill
\includegraphics[width=0.32\linewidth]{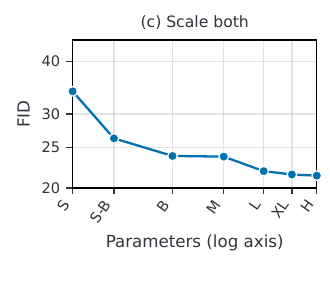}
\vspace{-8pt}
\caption{\textbf{The joint scaling phenomenon.} Starting from an S-sized teacher, critic, and student and fixing the student size, each panel scales (a, left) the teacher, (b, middle) the critic, or (c, right) both along the ladder, with the horizontal axis the parameter count of the scaled network. Scaling only the teacher initially improves FID but eventually degrades it and causes collapse at H; scaling only the critic yields marginal gains; scaling both improves the student at every size. (Settings: 1-step S student, ImageNet $256\times256$, w/o CFG; best FID with the per-cell learning-rate selection of Appendix~\ref{app:training_query}.)}
\label{fig:scaling_paths}
\end{figure}

\begin{figure}[!htb]
\centering
\includegraphics[width=0.9\linewidth]{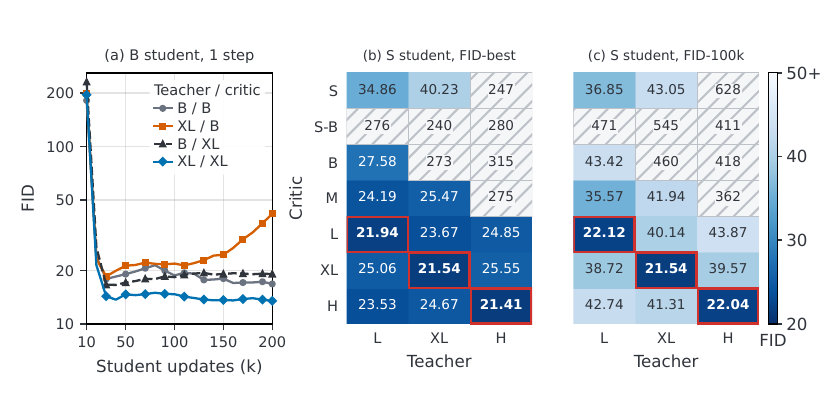}
\caption{\textbf{Scaling the teacher and critic on ImageNet.} Every panel varies the teacher and critic sizes while the student is fixed. \textbf{(a)} FID over training for a 1-step B student under four teacher/critic pairs (B/B, XL/B, B/XL,
XL/XL); the mismatched XL/B pair degrades only late, so (a) runs to 200k updates to
expose it. \textbf{(b, c)} A 1-step S student over a grid of L, XL, and H
teachers and S-to-H critics: (b) reports the best FID over training, and (c) the
FID at a fixed 100k-update budget. Red boxes mark the \emph{diagonal}, matched
teacher/critic pairs. For both students, matched pairs give the lowest FID, while
mismatched pairs often degrade and sometimes collapse (best FID of 100 or higher). (Settings: ImageNet
$256\times256$, w/o CFG; (a) default learning rates; (b, c) the best learning-rate
pair per cell by the lowest FID over training.)}
\label{fig:scaling}
\end{figure}

\textbf{Scaling the teacher.}
Since in standard diffusion models, generation quality generally improves with model size within an architecture family~\citep{peebles2023scalablediffusionmodelstransformers,ma2024sitexploringflowdiffusionbased}, a natural expectation is that a larger teacher produces a better student in DMD. However, with the critic and student fixed at S, as shown in Figure~\ref{fig:scaling_paths}a, FID reaches its lowest value of 28.5 with an M teacher, then worsens to 34.9 with L and 40.2 with XL, before the student collapses with H (FID 247).
The same reversal appears for a 1-step B student: the XL/B pair (teacher/critic) performs worse than B/B (Figure~\ref{fig:scaling}a).
The XL/B run trails B/B from 30k updates onward and reaches FID 41.7 at 200k updates, compared with 16.8 for B/B.
Thus, scaling the teacher while holding the student and critic fixed does not necessarily improve the student and can eventually degrade quality or cause collapse.

\textbf{Scaling the critic.}
Since scaling the teacher alone does not consistently improve the student, we next scale the critic while holding the teacher and student at S.
As shown in Figure~\ref{fig:scaling_paths}b, training remains stable as the critic scales up, but FID changes little across the critic ladder.
Similarly, for a 1-step B student with a B teacher, the B/XL pair reaches a best FID of 16.6, comparable to 16.8 for B/B (Figure~\ref{fig:scaling}a).

\begin{table}[!htb]
\centering
\setlength{\tabcolsep}{5pt}
\begin{tabular}{@{}ccrrrrrr@{}}
\toprule
 & & \multicolumn{3}{c}{FID$\downarrow$} & & & \\
\cmidrule(lr){3-5}
$T$ & $F$ & lowest & 100k & 200k & IS$\uparrow$ & Precision$\uparrow$ & Recall$\uparrow$\\
\midrule
B & B & 16.82 & 18.78 & \underline{16.82} & 99.85$\pm$2.36 & \textbf{0.762} & 0.290\\
XL & B & 18.56 & 21.91 & 41.75 & \underline{107.19$\pm$1.97} & 0.735 & 0.224\\
B & XL & \underline{16.61} & \underline{18.36} & 19.12 & 105.25$\pm$2.38 & 0.698 & \underline{0.333}\\
\rowcolor{ourgray}XL & XL & \textbf{13.50} & \textbf{14.72} & \textbf{13.50} & \textbf{110.14$\pm$2.49} & \underline{0.760} & \textbf{0.388}\\
\bottomrule
\end{tabular}
\caption{\textbf{Training dynamics.} The four runs of Figure~\ref{fig:scaling}a, a 1-step B student under B and XL teachers ($T$) and critics ($F$) with one recipe: the lowest FID over training, the FID at 100k and 200k updates, and IS, precision, and recall at the lowest-FID checkpoint. XL/XL leads on every metric except precision, where it trails the best by 0.002, and only the two matched pairs reach their lowest FID at the end of training. (Settings: ImageNet $256\times 256$, w/o CFG, default learning rates, 200k updates.)}
\label{tab:b1_metrics}
\end{table}

\begin{figure}[!htb]
\centering
\includegraphics[width=.92\linewidth]{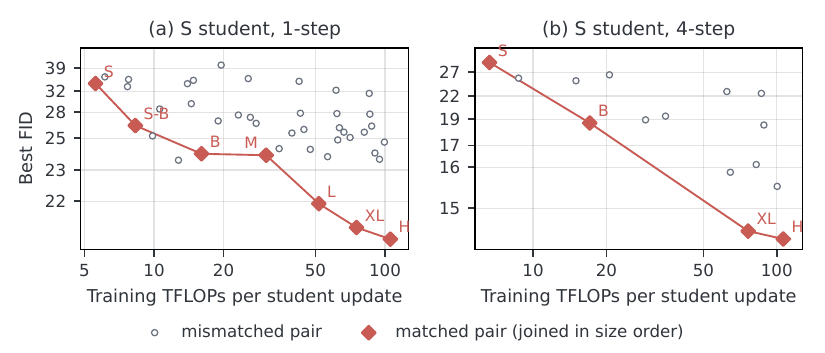}
\caption{\textbf{Per-update training compute versus FID.} Best FID versus TFLOPs per student update for every non-collapsed teacher/critic pair of the S-student grids at (a) 1 and (b) 4 steps. Matched pairs are labeled by size and joined in size order; the nonlinear FID axis expands the low-FID region. At comparable per-update compute, matched pairs sit at or near the lowest FID, while most mismatched pairs lie above them. (Settings: ImageNet $256\times256$, w/o CFG; learning rates as in Figure~\ref{fig:full_grids}; FLOPs from Table~\ref{tab:ladder}.)}
\label{fig:compute_pareto}
\end{figure}

\textbf{Joint scaling.}
We therefore scale the teacher and critic together at matched sizes, which we call \emph{joint scaling}.
As shown in Figure~\ref{fig:scaling_paths}c, for the 1-step S student, this improves FID at every size, from 34.0 to 21.4 along the ladder.
For the 1-step B student, XL/XL achieves the best FID of 13.5, ahead of B/B (16.8) and the mismatched XL/B and B/XL pairs (Figure~\ref{fig:scaling}a).
Figures~\ref{fig:scaling}b and~c extend the comparison to a 1-step S student with L, XL, and H teachers and S-to-H critics.
In both figure panels, the matched critic gives the lowest FID for each teacher. Examining FID throughout training in Table~\ref{tab:b1_metrics}, matched pairs remain close to their best FID over the long term, whereas most mismatched pairs degrade by 100k or 200k updates.

Figure~\ref{fig:full_grids} shows that every observed collapse occurs with a critic smaller than its teacher, and Table~\ref{tab:lr_ablation} further examines the effect of the critic learning rate. Across these experiments, joint scaling remains among the best-performing configurations and offers a favorable compute--quality trade-off (Figure~\ref{fig:compute_pareto}).

\textbf{Further Analysis. }
Unlike the frozen teacher, the critic must track the student's changing distribution throughout training.
A natural remedy is to update the critic more often, as suggested in DMD2~\citep{yin2024improveddistributionmatchingdistillation}.
Five critic updates per student update speed up early training, but the student-size critic still settles at an FID well above that of an XL critic (Figure~\ref{fig:update_ratio} and Appendix~\ref{app:update_ratio}).
Warming up a small critic on student samples before distillation also fails to prevent collapse (Figure~\ref{fig:critic_warmup} and Appendix~\ref{app:critic_warmup}).
A large critic need not be initialized from the teacher's weights.
For a 1-step B student distilled from the XL teacher pretrained for 7M updates by \citet{ma2024sitexploringflowdiffusionbased}, our XL critic pretrained for 800k updates reaches FID 12.01, close to initializing the critic from the teacher, while a B critic reaches only 35.96 (Table~\ref{tab:critic_checkpoints} and Appendix~\ref{app:initialization}).
Moreover, low-rank adapters on our frozen XL teacher, with under 1\% of the trainable parameters of a student-size critic, already outperform that full critic and thus offer a cheap form of joint scaling (Table~\ref{tab:lora_critic} and Appendix~\ref{app:lora_critic}).
Velocity knowledge distillation and MeanFlow \citep{geng2025meanflowsonestepgenerative, geng2026improvedmeanflowschallenges} distillation do not show the same collapse under larger teachers (Appendix~\ref{app:other_methods}), suggesting that the collapse is tied to distribution matching.
Together, these results suggest that in DMD, a larger critic helps track the student's evolving distribution when using a larger teacher, without requiring the teacher's exact pretrained weights.

\subsection{Step-aware Critic (STAC)}

\begin{figure}[!htb]
\centering
\includegraphics[width=\linewidth,trim=0 26 0 0,clip]{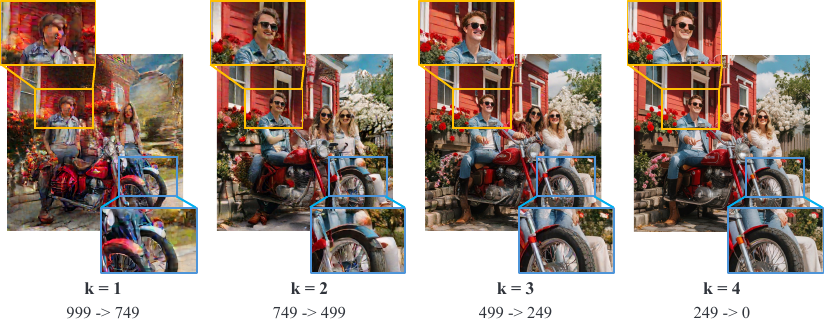}\\[3pt]
\footnotesize
\makebox[.25\linewidth]{$\mathbf{k=1}$}\makebox[.25\linewidth]{$\mathbf{k=2}$}\makebox[.25\linewidth]{$\mathbf{k=3}$}\makebox[.25\linewidth]{$\mathbf{k=4}$}\\[1pt]
\makebox[.25\linewidth]{$999\rightarrow749$}\makebox[.25\linewidth]{$749\rightarrow499$}\makebox[.25\linewidth]{$499\rightarrow249$}\makebox[.25\linewidth]{$249\rightarrow0$}
\normalsize
\caption{\textbf{Generation quality varies sharply across steps.}
Clean predictions of a 4-step SDXL student along one generation
path with a shared prompt and seed. Colored boxes and insets magnify the
regions discussed below. The first step already fixes the overall color
and layout but still carries significant artifacts; the second step
refines the image substantially, yet artifacts remain in detailed regions
such as the face and the wheel; the third step retains only small residual
artifacts around the mouth, which the last step removes.}
\label{fig:steps}
\end{figure}

\label{sec:step}

\textbf{Intuition. }
A $K$-step student produces clean $x_0$ predictions with different characteristics at each sampling step.
Early predictions often contain blur and artifacts, while later predictions recover finer detail, as illustrated in Figure~\ref{fig:steps}.
During multi-step DMD training, these predictions are pooled to train a critic that receives the score-evaluation level $t$ but not the generation step $k$.
The critic therefore estimates the score of a mixture across steps, which can differ from the score of each step's distribution even at the same $t$.

\textbf{Step-aware critic.}
We introduce a \emph{step-aware critic} (STAC) that conditions on the generation step during both critic training and student updates.
This allows it to estimate the score of each step's distribution.
The loss retains its DMD2 form, while the teacher target, rollout, and sampler remain unchanged.
STAC therefore adds no inference cost and little training overhead (detailed in Appendix~\ref{sec:external_overhead}).
The DMD algorithm with STAC is shown in Algorithm~\ref{alg:training_query}.
The critic now estimates step-specific scores rather than a mixture score, so early and late predictions receive supervision matched to their respective distributions.
On ImageNet, as a validation, we train a classifier to predict which generation step produced each student prediction (Figure~\ref{fig:step_separability} and Appendix~\ref{app:step_info}).
Its accuracy decreases with STAC, suggesting that predictions from different steps become more similar.
This is consistent with STAC providing more effective step-specific score estimates that guide all steps toward the same teacher distribution.
We note that TDM also addresses the issue of sharing a critic across steps, but restricts each step to a disjoint range of score-evaluation levels \citep{luo2025learningfewstepdiffusionmodels}, whereas STAC uses the same range for every step.

\begin{figure}[!tb]
\centering
\begin{minipage}[c]{.48\linewidth}
\centering
\includegraphics[width=\linewidth]{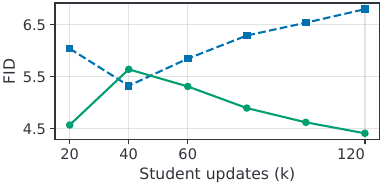}
\end{minipage}\hfill
\begin{minipage}[c]{.48\linewidth}
\caption{\textbf{Training dynamics of STAC.} FID during training
of a 4-step student without STAC (dashed) and with STAC (solid).
With STAC, FID improves steadily after 40k updates.
(Settings: XL student, teacher, and critic, ImageNet $256\times256$,
CFG $=1.3$.)}
\label{fig:stac_dynamics}
\end{minipage}
\end{figure}

Under the SiT ImageNet setting used in Section~\ref{sec:scaling}, we encode the generation noise level $r_k$ of step $k$ as four additional in-context tokens~\citep{geng2026improvedmeanflowschallenges}.
As these tokens are learned from scratch rather than inherited from pretraining, we train them at a much higher rate than the critic (Appendix~\ref{app:imagenet_settings}).
We evaluate STAC with a 4-step B student and B/B or XL/XL teacher-critic pairs, keeping all other settings fixed (Table~\ref{tab:cumulative}, left).
STAC lowers FID from 13.91 to 11.77 with B/B and from 8.08 to 6.90 with XL/XL, complementing joint scaling.
IS and recall improve in both settings, while precision changes by less than 0.03.
The gain also persists late in training (Figure~\ref{fig:stac_dynamics}).
For a 4-step XL student with an XL/XL pair at CFG $1.3$, FID without STAC worsens after 40k updates, whereas FID with STAC continues to improve and reaches its lowest value at the end of training.
STAC also reduces fluctuations in the critic and distribution-matching losses (Figure~\ref{fig:stac_cv} and Appendix~\ref{app:step_info}).

\paragraph{Ablation.}
To separate the effects of step information and learned tokens, we compare three controls for the 4-step B student with an XL/XL teacher-critic pair (Table~\ref{tab:cumulative}, right).
Learned tokens shared across all steps (\emph{const tokens}) reach FID 7.95, compared with 6.90 for STAC, indicating that extra tokens alone explain little of the gain.
Step-specific random tokens kept fixed during training (\emph{frozen tokens}) reach 7.76.
An \emph{AdaLN branch}~\citep{peebles2023scalablediffusionmodelstransformers} that injects the step through adaptive normalization reaches 7.68.
Its step embedder uses the same learning rate as STAC's conditioning parameters.
Both step-aware controls improve on shared tokens, but learned in-context tokens perform best.

\begin{table}[!htb]
\centering\small
\begin{minipage}[t]{.58\linewidth}
\centering
\setlength{\tabcolsep}{4pt}
\begin{tabular}{@{}lrrrr@{}}
\toprule
$T$ / $F$ & FID$\downarrow$ & IS$\uparrow$ & Prec.$\uparrow$ & Rec.$\uparrow$\\
\midrule
B / B & 13.91 & 116.81$\pm$1.98 & 0.792 & 0.334\\
\quad + STAC & 11.77 & 127.90$\pm$1.84 & \textbf{0.815} & 0.340\\
XL / XL & 8.08 & 161.01$\pm$3.75 & 0.739 & 0.452\\
\rowcolor{ourgray}\quad + STAC & \textbf{6.90} & \textbf{165.16$\pm$3.13} & 0.731 & \textbf{0.470}\\
\bottomrule
\end{tabular}
\end{minipage}\hfill
\begin{minipage}[t]{.38\linewidth}
\centering
\setlength{\tabcolsep}{4pt}
\begin{tabular}{@{}lr@{}}
\toprule
Step input to $F$ & FID$\downarrow$\\
\midrule
\rowcolor{ourgray}In-context tokens (STAC) & \textbf{6.90}\\
Const tokens & 7.95\\
Frozen tokens & 7.76\\
AdaLN branch & 7.68\\
\bottomrule
\end{tabular}
\end{minipage}
\caption{\textbf{Cumulative results and step-input ablations on ImageNet.}
Left: a 4-step B student under B/B and XL/XL teacher--critic pairs,
without and with STAC (``+ STAC''); joint scaling and STAC each lower FID,
and together they give the best student.
Right: comparison of the XL/XL student with STAC against three controls that replace its
learned in-context step tokens, namely \emph{const tokens} shared by all
steps, step-specific but random and untrained \emph{frozen tokens}, and
an \emph{AdaLN branch} that injects the step through adaptive
normalization. Relative to the XL/XL baseline without STAC (FID 8.08),
all three controls recover a smaller FID gain than STAC (6.90).
Every run shares the recipe, and IS is given with its standard deviation
over splits.
(Settings: ImageNet $256\times256$, w/o CFG, default learning rates,
FID at 100k updates.)}
\label{tab:cumulative}
\end{table}

\section{Additional Experiments}
\label{sec:external}

Beyond the controlled ImageNet setting, we run experiments on text-to-image and text-to-video generation.
For text-to-image, we use FLUX.2 \citep{blackforestlabs2025flux2},
whose 4B and 9B [klein] base models and 32B [dev] model let the teacher be
scaled alone or together with the critic; we also validate STAC on
SDXL~\citep{podell2023sdxlimprovinglatentdiffusion}, a denoising
UNet-based text-to-image model.
For video, we pair a 14B Wan2.1 teacher with a 1.3B student~\citep{wan2025wanopenadvancedlargescale}, so the critic can grow from the student's size to the teacher's.
Within each comparison the student and its number of steps are fixed.
We score SDXL images with FID, CLIP score (CLIP-S)~\citep{radford2021learningtransferablevisualmodels}, ImageReward
\citep{xu2023imagerewardlearningevaluatinghuman}, and HPSv3
\citep{ma2025hpsv3widespectrumhumanpreference}; FLUX.2 images with HPSv3
and DPG-Bench~\citep{hu2024ella}; and videos on
VBench~\citep{huang2023vbenchcomprehensivebenchmarksuite}.
Detailed training and evaluation protocols are in Appendices~\ref{app:sdxl_settings}--\ref{app:wan_settings}.

\subsection{Text-to-Image Experiments}
\label{sec:external_flux}
\textbf{FLUX.2}~\citep{blackforestlabs2025flux2}. We keep a 4B FLUX.2 [klein] base student fixed and distill it with the DMD2 recipe. We evaluate the 4B/4B, 9B/4B, and 9B/9B teacher/critic pairs from FLUX.2 [klein] base and the 32B/32B pair from FLUX.2 [dev]. We additionally apply STAC to the 9B/9B and 32B/32B pairs (Table~\ref{tab:flux_scaling}, Figure~\ref{fig:flux_qualitative}).
With the 4B critic, the 9B teacher lowers HPSv3 from 10.58 to 10.32
and raises DPG-Bench by only 0.50, so a larger teacher alone gives
little gain.
Scaling the critic to 9B as well raises both scores of the 9B-teacher
student, to 10.92 and 83.70, and adding STAC raises them further to
11.86 and 84.22.
At $1024\times1024$, the 32B teacher and critic give the strongest students: with STAC, the
student reaches 12.32 HPSv3 and 85.28 DPG-Bench, above both official
50-step base models (right columns) in HPSv3 and on par with the 9B one
in DPG-Bench.
The CFG value used in DMD often differs from that used by the base
model; we discuss its effect in Appendix~\ref{app:guidance}.

\begin{table}[!htb]
\centering
\small
\setlength{\tabcolsep}{5pt}
\begin{tabular}{ccccc@{\hspace{2em}}ccc}
\toprule
\multicolumn{5}{c}{\textit{Few-step: ours, 4B student, 4 steps}} &
\multicolumn{3}{c}{\textit{Multi-step: base model, 50 steps}} \\
\cmidrule(r{1em}){1-5} \cmidrule{6-8}
$T$ & $F$ & STAC & HPSv3$\uparrow$ & DPG-Bench$\uparrow$ &
Model & HPSv3$\uparrow$ & DPG-Bench$\uparrow$ \\
\midrule
4B  & 4B  & w/o & 10.58 & 81.58 & 4B & 10.51 & 82.88 \\
\addlinespace[2pt]
9B  & 4B  & w/o & 10.32 & 82.08 & & & \\
9B  & 9B  & w/o & 10.92 & 83.70 & & & \\
\cellcolor{ourgray}9B & \cellcolor{ourgray}9B & \cellcolor{ourgray}w/ &
\cellcolor{ourgray}11.86 & \cellcolor{ourgray}84.22 &
\multirow{-3}{*}{9B} & \multirow{-3}{*}{10.58} & \multirow{-3}{*}{85.28} \\
\addlinespace[2pt]
32B & 32B & w/o & 12.10 & 84.98 & & & \\
\cellcolor{ourgray}32B & \cellcolor{ourgray}32B & \cellcolor{ourgray}w/ &
\cellcolor{ourgray}\textbf{12.32} & \cellcolor{ourgray}\textbf{85.28} &
\multirow{-2}{*}{32B} & \multirow{-2}{*}{12.14} & \multirow{-2}{*}{86.79} \\
\bottomrule
\end{tabular}
\caption{\textbf{FLUX.2 benchmark at $1024\times1024$.} Left: our 4B
students with 4-step sampling, distilled with different teacher ($T$) and
critic ($F$) sizes. The 4B and 9B models used as teachers or critics come
from FLUX.2 [klein] base; the 32B models come from FLUX.2 [dev]. Right: the official base model
of the teacher's size, sampled with 50 steps and CFG 4.0.
Details in Appendix~\ref{app:flux_settings}.}
\label{tab:flux_scaling}
\end{table}

\label{sec:external_sdxl}
\textbf{SDXL}~\citep{podell2023sdxlimprovinglatentdiffusion}. We append the four step tokens to the critic's text sequence.
Under pure DMD with a 4-step student, STAC improves FID from 34.71 to 32.73 at an unchanged CLIP score (Appendix~\ref{sec:external_implementation}).
We then add STAC to the full DMD2 recipe, with its GAN loss and
backward simulation, by continuing training from the official DMD2
checkpoint (Table~\ref{tab:sdxl_systems} and Figure~\ref{fig:sdxl_qualitative_all}).
In this recipe, the step tokens also enter the critic's GAN branch,
while the pure-DMD result above, which has no GAN loss, isolates their
effect on the score estimate.
With STAC, the continued student improves FID from 18.95 to 16.88 and
ImageReward from 71.01 to 80.00, whereas an equally long continuation
without STAC worsens all four metrics.
This further supports the usefulness of conditioning the critic on the student's generation step.

\begin{table}[!htb]
\centering
\small
\setlength{\tabcolsep}{6pt}
\renewcommand{\arraystretch}{0.92}
\begin{tabular}{@{}lrrrr@{}}
\toprule
Method & FID$\downarrow$ & CLIP-S$\uparrow$ & ImageReward$\uparrow$ & HPSv3$\uparrow$ \\
\midrule
LCM-LoRA~\citep{luo2023lcmlorauniversalstablediffusionacceleration} & 22.27 & 31.71 & 39.56 & 6.45 \\
SDXL-Turbo~\citep{sauer2023adversarialdiffusiondistillation} & 27.27 & 32.16 & 46.09 & 9.09 \\
SDXL-Lightning~\citep{lin2024sdxllightningprogressiveadversarialdiffusion} & 24.49 & 32.31 & 57.48 & 9.48 \\
Flash~\citep{chadebec2024flashdiffusionacceleratingconditional} & 22.96 & 31.84 & 19.04 & 6.49 \\
PCM~\citep{wang2024phasedconsistencymodels} & 24.13 & 32.52 & 64.73 & 9.46 \\
DMD2, official~\citep{yin2024improveddistributionmatchingdistillation} & 18.95 & 33.14 & 71.01 & 9.64 \\
Decoupled DMD~\citep{liu2025decoupleddmdcfgaugmentation} & 17.80 & \textbf{33.62} & 78.61 & 9.79 \\
\midrule
DMD2, continued w/o STAC & 19.49 & 33.04 & 70.10 & 9.15 \\
\rowcolor{ourgray}\textbf{DMD2 + STAC (Ours)} & \textbf{16.88} & 33.60 & \textbf{80.00} & \textbf{9.84} \\
\bottomrule
\end{tabular}
\caption{\textbf{Comparison with distilled SDXL systems.} SDXL
systems with 4-step sampling on 10k COCO prompts at $512\times512$. The first seven
rows are the values reported by \citet{liu2025decoupleddmdcfgaugmentation},
and our rows follow the same protocol. CLIP-S and ImageReward are
multiplied by 100. Continuing DMD2 training with STAC improves clearly over the
official checkpoint on every metric, whereas the same continuation
without STAC falls below it. Detailed settings in Appendix~\ref{app:sdxl_settings}.}
\label{tab:sdxl_systems}
\end{table}

\subsection{Text-to-Video Experiments}
\label{sec:external_wan}
\textbf{Wan2.1}~\citep{wan2025wanopenadvancedlargescale} provides two model sizes, a 14B and a 1.3B model. We scale the teacher and the critic while keeping the 1.3B student fixed.
We evaluate the distilled four-step 1.3B student models on VBench~\citep{huang2023vbenchcomprehensivebenchmarksuite}.

\begin{figure}[!htb]
\centering
\includegraphics[width=\linewidth]{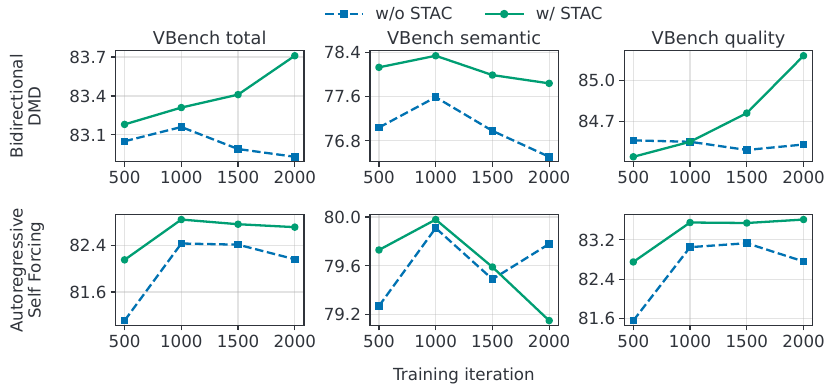}
\caption{\textbf{Controlled comparison of STAC on Wan2.1.}
VBench scores during training for bidirectional DMD (top) and autoregressive
Self Forcing (bottom), without and with STAC. STAC leads in VBench total at
every evaluation in both regimes. (Settings: 1.3B teacher, student, and critic;
ODE initialization from the CausVid repository~\citep{yin2025slowbidirectionalfastautoregressive};
4 steps, 480p, 2k iterations; bidirectional DMD uses no backward
simulation, while the autoregressive runs use an online Self Forcing rollout.)}
\label{fig:wan_step_conditioning}
\end{figure}

To isolate the effect of STAC, we first adopt a deliberately simple
bidirectional DMD setting (Figure~\ref{fig:wan_step_conditioning}, top).
The teacher, critic, and student are all 1.3B, the student is initialized from
the ODE checkpoint released in the CausVid repository~\citep{yin2025slowbidirectionalfastautoregressive},
and we disable backward simulation. Thus, the only difference between the two
runs is whether the critic is conditioned on the generation step through STAC.
Under this controlled setup, the multi-step DMD baseline peaks
near 1k iterations and then degrades, whereas STAC keeps improving and
attains a higher VBench total throughout training. In a separate
autoregressive test, we use an online Self Forcing rollout while retaining the
same model sizes and initialization
(Figure~\ref{fig:wan_step_conditioning}, bottom). STAC similarly attains a
higher VBench total at every evaluated iteration and degrades less after
prolonged training than the step-agnostic baseline.

Finally, we study joint scaling with an autoregressive student initialized from the Causal Forcing ODE checkpoint. We train this student using the Self Forcing rollout.
Across the controlled configurations in Table~\ref{tab:wan_main}, we hold the student initialization and the remaining training settings fixed while varying the teacher and critic sizes and adding STAC.
The 1.3B/1.3B teacher/critic pair yields the highest Dynamic Degree and the lowest semantic score, with excessive motion and visible artifacts (Figure~\ref{fig:wan_qualitative}).
Scaling only the teacher to 14B leaves the VBench total almost unchanged.
Joint scaling to a 14B teacher and critic raises the VBench total to 84.26.
Compared with the default 14B/1.3B teacher/critic pair, joint scaling improves the semantic and quality scores and raises Dynamic Degree from 52.06 to 57.52, without the excessive motion of the 1.3B/1.3B pair.
Adding STAC yields small further gains. Unlike Figure~\ref{fig:wan_step_conditioning}, these runs use the Causal Forcing ODE initialization and train for 1.5k rather than 2k iterations, so their absolute VBench values are not directly comparable.

\begin{table}[!htb]
\centering
\setlength{\tabcolsep}{5pt}
\begin{tabular}{@{}lrrrr@{}}
\toprule
Configuration & Total$\uparrow$ & Quality$\uparrow$ & Semantic$\uparrow$ & Dynamic$\uparrow$ \\
\midrule
CausVid (reproduced) & 82.76 & 83.49 & 79.85 & -- \\
Self Forcing (our eval.) & 83.87 & 84.59 & 81.00 & 50.12 \\
\midrule
Causal Forcing~\citep{zhu2026causalforcingautoregressivediffusion} & & & & \\
$T=14$B, $F=1.3$B (Default) & 83.94 & 84.73 & 80.79 & 52.06 \\
$T=1.3$B, $F=1.3$B & 83.96 & 85.14 & 79.26 & 81.39 \\
$T=14$B, $F=14$B & 84.26 & 85.04 & 81.16 & 57.52 \\
\rowcolor{ourgray}\quad + STAC (Ours) & 84.32 & 85.08 & 81.32 & 58.06 \\
\bottomrule
\end{tabular}
\caption{\textbf{Critic size and STAC for 1.3B Wan2.1 students.}
The first rows are Self Forcing's CausVid reproduction and our Self Forcing
evaluation~\citep{yin2025slowbidirectionalfastautoregressive,huang2025selfforcingbridgingtraintest}.
Controlled rows share a Causal Forcing ODE initialization and Self Forcing
rollout; the default is 14B/1.3B, and the last adds STAC to 14B/14B. A 14B
critic improves all metrics (Dynamic: 52.06 to 57.52); STAC improves them
further. (4 steps, 480p VBench; Appendix~\ref{app:wan_settings}.)}
\label{tab:wan_main}
\end{table}

\FloatBarrier

\section{Discussion}

This work revisits a component often treated as secondary in Distribution Matching Distillation: the critic. We examine what the critic needs in both capacity and conditioning to effectively guide a few-step student. Across controlled ImageNet scaling ladders and large-scale image and video models, pairing a larger teacher with a small critic frequently degrades distillation or triggers training collapse. Scaling the critic alongside the teacher (\emph{joint scaling}) unlocks the benefits of larger teachers and remains compute-efficient. Beyond capacity, standard multi-step DMD forces the critic to estimate the score of a mixture distribution pooled across different generation steps. Conditioning the critic on the generation step via STAC disentangles these step-specific distributions, improving both sample quality and training stability.

While matching critic and teacher capacities is a reliable empirical rule across our tested architectures, training budgets, and hyperparameter sweep protocols, why a stronger teacher alone fails to stably guide the student remains an open theoretical question. Our diagnostics characterize how teacher--critic capacity mismatch destabilizes the score difference, and developing methods that can safely distill from a massive teacher using a lightweight critic is a promising direction for reducing training memory and compute. Finally, because step conditioning breaks the structural symmetry between the critic and the teacher, exploring broader asymmetric designs in architecture or objective may further advance distribution-matching distillation.

\label{maintextend}
\flushbottom
\bibliographystyle{iclr2027_conference}
\bibliography{bibliography/references_display}

\begin{thebibliography}{75}
\providecommand{\natexlab}[1]{#1}
\providecommand{\url}[1]{\texttt{#1}}
\expandafter\ifx\csname urlstyle\endcsname\relax
  \providecommand{\doi}[1]{doi: #1}\else
  \providecommand{\doi}{doi: \begingroup \urlstyle{rm}\Url}\fi

\bibitem[Bai et~al.(2026)Bai, Zhou, Shao, Zhong, Yang, Chen, Chen, and
  Xie]{bai2026optimizingfewstepgenerationadaptive}
Lichen Bai, Zikai Zhou, Shitong Shao, Wenliang Zhong, Shuo Yang, Shuo Chen,
  Bojun Chen, and Zeke Xie.
\newblock {Optimizing Few-Step Generation with Adaptive Matching Distillation},
  2026.
\newblock URL \url{https://arxiv.org/abs/2602.07345}.

\bibitem[Bandyopadhyay et~al.(2026)Bandyopadhyay, Ren, Huang, Wu, Cao, Li, Chu,
  Fidler, Song, and
  Wang]{bandyopadhyay2026contextmatcheddistillationteachercausality}
Hmrishav Bandyopadhyay, Xuanchi Ren, Zijian Huang, Jay~Zhangjie Wu, Tianshi
  Cao, Ruilong Li, Bryan Chu, Sanja Fidler, Yi-Zhe Song, and Zian Wang.
\newblock {Context-Matched Distillation: Teacher Causality for Autoregressive
  Video Distillation}, 2026.
\newblock URL \url{https://arxiv.org/abs/2608.13391}.

\bibitem[{Black Forest Labs}(2025)]{blackforestlabs2025flux2}
{Black Forest Labs}.
\newblock {FLUX.2: Frontier Visual Intelligence}.
\newblock \url{https://github.com/black-forest-labs/flux2}, 2025.

\bibitem[Chadebec et~al.(2024)Chadebec, Tasar, Benaroche, and
  Aubin]{chadebec2024flashdiffusionacceleratingconditional}
Clément Chadebec, Onur Tasar, Eyal Benaroche, and Benjamin Aubin.
\newblock {Flash Diffusion: Accelerating Any Conditional Diffusion Model for
  Few Steps Image Generation}, 2024.
\newblock URL \url{https://arxiv.org/abs/2406.02347}.

\bibitem[Chen et~al.(2026)Chen, Liu, Jia, Wang, Ling, Qu, and
  Gao]{chen2026dataforcingdistillationrestoringdiversity}
Siyi Chen, Shaowei Liu, Yixuan Jia, Zian Wang, Huan Ling, Qing Qu, and Jun Gao.
\newblock {Data-Forcing Distillation: Restoring Diversity and Fidelity in
  Few-Step Video Generation}, 2026.
\newblock URL \url{https://arxiv.org/abs/2606.18478}.

\bibitem[Dhariwal \& Nichol(2021)Dhariwal and
  Nichol]{dhariwal2021diffusionmodelsbeatgans}
Prafulla Dhariwal and Alex Nichol.
\newblock {Diffusion Models Beat GANs on Image Synthesis}, 2021.
\newblock URL \url{https://arxiv.org/abs/2105.05233}.

\bibitem[Esser et~al.(2024)Esser, Kulal, Blattmann, Entezari, Müller, Saini,
  Levi, Lorenz, Sauer, Boesel, Podell, Dockhorn, English, Lacey, Goodwin,
  Marek, and Rombach]{esser2024scalingrectifiedflowtransformers}
Patrick Esser, Sumith Kulal, Andreas Blattmann, Rahim Entezari, Jonas Müller,
  Harry Saini, Yam Levi, Dominik Lorenz, Axel Sauer, Frederic Boesel, Dustin
  Podell, Tim Dockhorn, Zion English, Kyle Lacey, Alex Goodwin, Yannik Marek,
  and Robin Rombach.
\newblock {Scaling Rectified Flow Transformers for High-Resolution Image
  Synthesis}, 2024.
\newblock URL \url{https://arxiv.org/abs/2403.03206}.

\bibitem[Fan et~al.(2026)Fan, Qiu, Wu, Wang, Lin, Ren, Lin, Gong, and
  Yang]{fan2026phaseddmdfewstepdistribution}
Xiangyu Fan, Zesong Qiu, Zhuguanyu Wu, Fanzhou Wang, Zhiqian Lin, Tianxiang
  Ren, Dahua Lin, Ruihao Gong, and Lei Yang.
\newblock {Phased DMD: Few-step Distribution Matching Distillation via Score
  Matching within Subintervals}, 2026.
\newblock URL \url{https://arxiv.org/abs/2510.27684}.

\bibitem[Fang et~al.(2025)Fang, Yu, Duan, Huang, Bai, Cai, Wang, Liu, Liu, and
  Li]{fang2025fluxreason6mprismbenchmillionscale}
Rongyao Fang, Aldrich Yu, Chengqi Duan, Linjiang Huang, Shuai Bai, Yuxuan Cai,
  Kun Wang, Si~Liu, Xihui Liu, and Hongsheng Li.
\newblock {FLUX-Reason-6M \& PRISM-Bench: A Million-Scale Text-to-Image
  Reasoning Dataset and Comprehensive Benchmark}, 2025.
\newblock URL \url{https://arxiv.org/abs/2509.09680}.

\bibitem[Feng et~al.(2026)Feng, Cui, Ban, and
  Hsieh]{feng2026oneforcingstableonestepautoregressive}
Jiaqi Feng, Justin Cui, Yuanhao Ban, and Cho-Jui Hsieh.
\newblock {One-Forcing: Towards Stable One-Step Autoregressive Video
  Generation}, 2026.
\newblock URL \url{https://arxiv.org/abs/2605.23458}.

\bibitem[Frans et~al.(2025)Frans, Hafner, Levine, and
  Abbeel]{frans2025stepdiffusionshortcutmodels}
Kevin Frans, Danijar Hafner, Sergey Levine, and Pieter Abbeel.
\newblock {One Step Diffusion via Shortcut Models}, 2025.
\newblock URL \url{https://arxiv.org/abs/2410.12557}.

\bibitem[Ge et~al.(2026)Ge, Zhang, Xu, Zhang, Zhang, Wang, and
  Zhang]{ge2026senseflowscalingdistributionmatching}
Xingtong Ge, Xin Zhang, Tongda Xu, Yi~Zhang, Xinjie Zhang, Yan Wang, and Jun
  Zhang.
\newblock {SenseFlow: Scaling Distribution Matching for Flow-based
  Text-to-Image Distillation}, 2026.
\newblock URL \url{https://arxiv.org/abs/2506.00523}.

\bibitem[Geng et~al.(2025)Geng, Deng, Bai, Kolter, and
  He]{geng2025meanflowsonestepgenerative}
Zhengyang Geng, Mingyang Deng, Xingjian Bai, J.~Zico Kolter, and Kaiming He.
\newblock {Mean Flows for One-step Generative Modeling}, 2025.
\newblock URL \url{https://arxiv.org/abs/2505.13447}.

\bibitem[Geng et~al.(2026)Geng, Lu, Wu, Shechtman, Kolter, and
  He]{geng2026improvedmeanflowschallenges}
Zhengyang Geng, Yiyang Lu, Zongze Wu, Eli Shechtman, J.~Zico Kolter, and
  Kaiming He.
\newblock {Improved Mean Flows: On the Challenges of Fastforward Generative
  Models}, 2026.
\newblock URL \url{https://arxiv.org/abs/2512.02012}.

\bibitem[He et~al.(2026)He, Peng, Liu, Wang, Zhang, Cui, Kang, Jiang, An, Ren,
  Xu, Guo, Gong, Wu, Li, Song, Liu, Li, and
  Zhou]{he2026matrixgame20opensourcerealtime}
Xianglong He, Chunli Peng, Zexiang Liu, Boyang Wang, Yifan Zhang, Qi~Cui, Fei
  Kang, Biao Jiang, Mengyin An, Yangyang Ren, Baixin Xu, Hao-Xiang Guo,
  Kaixiong Gong, Size Wu, Wei Li, Xuchen Song, Yang Liu, Yangguang Li, and
  Yahui Zhou.
\newblock {Matrix-Game 2.0: An Open-Source Real-Time and Streaming Interactive
  World Model}, 2026.
\newblock URL \url{https://arxiv.org/abs/2508.13009}.

\bibitem[Heusel et~al.(2017)Heusel, Ramsauer, Unterthiner, Nessler, and
  Hochreiter]{heusel2018ganstrainedtimescaleupdate}
Martin Heusel, Hubert Ramsauer, Thomas Unterthiner, Bernhard Nessler, and Sepp
  Hochreiter.
\newblock {GANs Trained by a Two Time-Scale Update Rule Converge to a Local
  Nash Equilibrium}, 2017.
\newblock URL \url{https://arxiv.org/abs/1706.08500}.

\bibitem[Ho \& Salimans(2022)Ho and
  Salimans]{ho2022classifierfreediffusionguidance}
Jonathan Ho and Tim Salimans.
\newblock {Classifier-Free Diffusion Guidance}, 2022.
\newblock URL \url{https://arxiv.org/abs/2207.12598}.

\bibitem[Hu et~al.(2021)Hu, Shen, Wallis, Allen-Zhu, Li, Wang, Wang, and
  Chen]{hu2021loralowrankadaptationlarge}
Edward~J. Hu, Yelong Shen, Phillip Wallis, Zeyuan Allen-Zhu, Yuanzhi Li, Shean
  Wang, Lu~Wang, and Weizhu Chen.
\newblock {LoRA: Low-Rank Adaptation of Large Language Models}, 2021.
\newblock URL \url{https://arxiv.org/abs/2106.09685}.

\bibitem[Hu et~al.(2024)Hu, Wang, Fang, Fu, Cheng, and Yu]{hu2024ella}
Xiwei Hu, Rui Wang, Yixiao Fang, Bin Fu, Pei Cheng, and Gang Yu.
\newblock {ELLA: Equip Diffusion Models with LLM for Enhanced Semantic
  Alignment}, 2024.
\newblock URL \url{https://arxiv.org/abs/2403.05135}.

\bibitem[Huang et~al.(2025)Huang, Li, He, Zhou, and
  Shechtman]{huang2025selfforcingbridgingtraintest}
Xun Huang, Zhengqi Li, Guande He, Mingyuan Zhou, and Eli Shechtman.
\newblock {Self Forcing: Bridging the Train-Test Gap in Autoregressive Video
  Diffusion}, 2025.
\newblock URL \url{https://arxiv.org/abs/2506.08009}.

\bibitem[Huang et~al.(2026)Huang, Zhou, Wang, Zhang, Zhang, and
  Pang]{huang2026reinforcingfewstepgeneratorsrewardtilted}
Yushi Huang, Xiangxin Zhou, Ruoyu Wang, Chi Zhang, Jun Zhang, and Tianyu Pang.
\newblock {Reinforcing Few-step Generators via Reward-Tilted Distribution
  Matching}, 2026.
\newblock URL \url{https://arxiv.org/abs/2605.26108}.

\bibitem[Huang et~al.(2023)Huang, He, Yu, Zhang, Si, Jiang, Zhang, Wu, Jin,
  Chanpaisit, Wang, Chen, Wang, Lin, Qiao, and
  Liu]{huang2023vbenchcomprehensivebenchmarksuite}
Ziqi Huang, Yinan He, Jiashuo Yu, Fan Zhang, Chenyang Si, Yuming Jiang, Yuanhan
  Zhang, Tianxing Wu, Qingyang Jin, Nattapol Chanpaisit, Yaohui Wang, Xinyuan
  Chen, Limin Wang, Dahua Lin, Yu~Qiao, and Ziwei Liu.
\newblock {VBench: Comprehensive Benchmark Suite for Video Generative Models},
  2023.
\newblock URL \url{https://arxiv.org/abs/2311.17982}.

\bibitem[Jiang et~al.(2026)Jiang, Liu, Wang, Wu, Li, Li, Jin, Liu, Lu, Li,
  Zhang, Wang, Hoi, Gao, and
  Yang]{jiang2026distributionmatchingdistillationmeets}
Dengyang Jiang, Dongyang Liu, Zanyi Wang, Qilong Wu, Liuzhuozheng Li,
  Hengzhuang Li, Xin Jin, David Liu, Changsheng Lu, Zhen Li, Bo~Zhang, Mengmeng
  Wang, Steven Hoi, Peng Gao, and Harry Yang.
\newblock {Distribution Matching Distillation Meets Reinforcement Learning},
  2026.
\newblock URL \url{https://arxiv.org/abs/2511.13649}.

\bibitem[Kalajdzievski(2023)]{kalajdzievski2023rankstabilizationscalingfactor}
Damjan Kalajdzievski.
\newblock {A Rank Stabilization Scaling Factor for Fine-Tuning with LoRA},
  2023.
\newblock URL \url{https://arxiv.org/abs/2312.03732}.

\bibitem[Karras et~al.(2024)Karras, Aittala, Kynkäänniemi, Lehtinen, Aila,
  and Laine]{karras2024guidingdiffusionmodelbad}
Tero Karras, Miika Aittala, Tuomas Kynkäänniemi, Jaakko Lehtinen, Timo Aila,
  and Samuli Laine.
\newblock {Guiding a Diffusion Model with a Bad Version of Itself}, 2024.
\newblock URL \url{https://arxiv.org/abs/2406.02507}.

\bibitem[Kim et~al.(2026)Kim, Lee, and
  Park]{kim2026distributionmatchingdistillationfake}
Youngjoong Kim, Deokyeong Lee, and Jaesik Park.
\newblock {Distribution Matching Distillation without Fake Score Network},
  2026.
\newblock URL \url{https://arxiv.org/abs/2605.19256}.

\bibitem[Kynkäänniemi et~al.(2019)Kynkäänniemi, Karras, Laine, Lehtinen,
  and Aila]{kynkaanniemi2019improvedprecisionrecallmetric}
Tuomas Kynkäänniemi, Tero Karras, Samuli Laine, Jaakko Lehtinen, and Timo
  Aila.
\newblock {Improved Precision and Recall Metric for Assessing Generative
  Models}, 2019.
\newblock URL \url{https://arxiv.org/abs/1904.06991}.

\bibitem[Li et~al.(2025)Li, Long, Zheng, Gao, Piramuthu, Chen, and
  Wang]{li2025t2vturbov2enhancingvideogeneration}
Jiachen Li, Qian Long, Jian Zheng, Xiaofeng Gao, Robinson Piramuthu, Wenhu
  Chen, and William~Yang Wang.
\newblock {T2V-Turbo-v2: Enhancing Video Generation Model Post-Training through
  Data, Reward, and Conditional Guidance Design}, 2025.
\newblock URL \url{https://arxiv.org/abs/2410.05677}.

\bibitem[Lin et~al.(2024)Lin, Wang, and
  Yang]{lin2024sdxllightningprogressiveadversarialdiffusion}
Shanchuan Lin, Anran Wang, and Xiao Yang.
\newblock {SDXL-Lightning: Progressive Adversarial Diffusion Distillation},
  2024.
\newblock URL \url{https://arxiv.org/abs/2402.13929}.

\bibitem[Lin et~al.(2015)Lin, Maire, Belongie, Bourdev, Girshick, Hays, Perona,
  Ramanan, Zitnick, and Dollár]{lin2015microsoftcococommonobjects}
Tsung-Yi Lin, Michael Maire, Serge Belongie, Lubomir Bourdev, Ross Girshick,
  James Hays, Pietro Perona, Deva Ramanan, C.~Lawrence Zitnick, and Piotr
  Dollár.
\newblock {Microsoft COCO: Common Objects in Context}, 2015.
\newblock URL \url{https://arxiv.org/abs/1405.0312}.

\bibitem[Lipman et~al.(2023)Lipman, Chen, Ben-Hamu, Nickel, and
  Le]{lipman2023flowmatchinggenerativemodeling}
Yaron Lipman, Ricky T.~Q. Chen, Heli Ben-Hamu, Maximilian Nickel, and Matt Le.
\newblock {Flow Matching for Generative Modeling}, 2023.
\newblock URL \url{https://arxiv.org/abs/2210.02747}.

\bibitem[Liu et~al.(2025)Liu, Gao, Liu, Du, Li, Wu, Jin, Cao, Zhang, Li, and
  Hoi]{liu2025decoupleddmdcfgaugmentation}
Dongyang Liu, Peng Gao, David Liu, Ruoyi Du, Zhen Li, Qilong Wu, Xin Jin, Sihan
  Cao, Shifeng Zhang, Hongsheng Li, and Steven Hoi.
\newblock {Decoupled DMD: CFG Augmentation as the Spear, Distribution Matching
  as the Shield}, 2025.
\newblock URL \url{https://arxiv.org/abs/2511.22677}.

\bibitem[Liu et~al.(2026)Liu, Feng, Zou, Qian, Zhu, Xia, Dong, Lin, Xiong,
  Chen, Ding, Duan, Gao, Han, He, Hu, Hua, Jiang, Kong, Lei, Li, Liu, Liu, Liu,
  Liu, Liu, Ma, Pan, Ren, Sheng, Sun, Sun, Tu, Wan, Wang, Wang, Wu, Yang, Yang,
  Yang, Yang, Zhang, Zhang, Zhang, Zhang, Zhang, Zhang, Zhao, and
  Zhou]{liu2026ernieimagetechnicalreport}
Jiaxiang Liu, Zhida Feng, Pengyu Zou, Zhenyu Qian, Tianrui Zhu, Jun Xia, Yuehu
  Dong, Yanzheng Lin, Honglin Xiong, Anqi Chen, Yunpeng Ding, Jinghui Duan, Lin
  Gao, Chao Han, Tiechao He, Jiakang Hu, Ranjun Hua, Xueming Jiang, Qingli
  Kong, Yuting Lei, Tianyu Li, Yunlin Liu, Changling Liu, Yaxin Liu, Yi~Liu,
  Xuguang Liu, Xiaolong Ma, Yan Pan, Yiran Ren, Nan Sheng, Yu~Sun, Siyang Sun,
  Yixiang Tu, Yang Wan, Huanai Wang, Siqi Wang, Yang Wu, Youzhi Yang, Xiaowen
  Yang, Jianwen Yang, Yehua Yang, Quanwen Zhang, Xinmin Zhang, Haoxin Zhang,
  Xiang Zhang, Jun Zhang, Qian Zhang, Qiao Zhao, and Qi~Zhou.
\newblock {ERNIE-Image Technical Report}, 2026.
\newblock URL \url{https://arxiv.org/abs/2605.25347}.

\bibitem[Lu et~al.(2025)Lu, Ren, Xia, Lin, Wang, Xiao, Ma, Xie, and
  Lai]{lu2025adversarialdistributionmatchingdiffusion}
Yanzuo Lu, Yuxi Ren, Xin Xia, Shanchuan Lin, Xing Wang, Xuefeng Xiao, Andy~J.
  Ma, Xiaohua Xie, and Jian-Huang Lai.
\newblock {Adversarial Distribution Matching for Diffusion Distillation Towards
  Efficient Image and Video Synthesis}, 2025.
\newblock URL \url{https://arxiv.org/abs/2507.18569}.

\bibitem[Lu et~al.(2026)Lu, Lu, Sun, Zhao, Jiang, Wang, Li, Geng, and
  He]{lu2026onesteplatentfreeimagegeneration}
Yiyang Lu, Susie Lu, Qiao Sun, Hanhong Zhao, Zhicheng Jiang, Xianbang Wang,
  Tianhong Li, Zhengyang Geng, and Kaiming He.
\newblock {One-step Latent-free Image Generation with Pixel Mean Flows}, 2026.
\newblock URL \url{https://arxiv.org/abs/2601.22158}.

\bibitem[Luo et~al.(2023{\natexlab{a}})Luo, Tan, Patil, Gu, von Platen, Passos,
  Huang, Li, and Zhao]{luo2023lcmlorauniversalstablediffusionacceleration}
Simian Luo, Yiqin Tan, Suraj Patil, Daniel Gu, Patrick von Platen, Apolinário
  Passos, Longbo Huang, Jian Li, and Hang Zhao.
\newblock {LCM-LoRA: A Universal Stable-Diffusion Acceleration Module},
  2023{\natexlab{a}}.
\newblock URL \url{https://arxiv.org/abs/2311.05556}.

\bibitem[Luo et~al.(2023{\natexlab{b}})Luo, Hu, Zhang, Sun, Li, and
  Zhang]{luo2024diffinstructuniversalapproachtransferring}
Weijian Luo, Tianyang Hu, Shifeng Zhang, Jiacheng Sun, Zhenguo Li, and Zhihua
  Zhang.
\newblock {Diff-Instruct: A Universal Approach for Transferring Knowledge From
  Pre-trained Diffusion Models}, 2023{\natexlab{b}}.
\newblock URL \url{https://arxiv.org/abs/2305.18455}.

\bibitem[Luo et~al.(2025)Luo, Hu, Sun, Cai, and
  Tang]{luo2025learningfewstepdiffusionmodels}
Yihong Luo, Tianyang Hu, Jiacheng Sun, Yujun Cai, and Jing Tang.
\newblock {Learning Few-Step Diffusion Models by Trajectory Distribution
  Matching}, 2025.
\newblock URL \url{https://arxiv.org/abs/2503.06674}.

\bibitem[Ma et~al.(2024)Ma, Goldstein, Albergo, Boffi, Vanden-Eijnden, and
  Xie]{ma2024sitexploringflowdiffusionbased}
Nanye Ma, Mark Goldstein, Michael~S. Albergo, Nicholas~M. Boffi, Eric
  Vanden-Eijnden, and Saining Xie.
\newblock {SiT: Exploring Flow and Diffusion-based Generative Models with
  Scalable Interpolant Transformers}, 2024.
\newblock URL \url{https://arxiv.org/abs/2401.08740}.

\bibitem[Ma et~al.(2025)Ma, Shui, Wu, Sun, and
  Li]{ma2025hpsv3widespectrumhumanpreference}
Yuhang Ma, Yunhao Shui, Xiaoshi Wu, Keqiang Sun, and Hongsheng Li.
\newblock {HPSv3: Towards Wide-Spectrum Human Preference Score}, 2025.
\newblock URL \url{https://arxiv.org/abs/2508.03789}.

\bibitem[Nguyen \& Tran(2024)Nguyen and
  Tran]{nguyen2024swiftbrushonesteptexttoimagediffusion}
Thuan~Hoang Nguyen and Anh Tran.
\newblock {SwiftBrush: One-Step Text-to-Image Diffusion Model with Variational
  Score Distillation}, 2024.
\newblock URL \url{https://arxiv.org/abs/2312.05239}.

\bibitem[Peebles \& Xie(2023)Peebles and
  Xie]{peebles2023scalablediffusionmodelstransformers}
William Peebles and Saining Xie.
\newblock {Scalable Diffusion Models with Transformers}, 2023.
\newblock URL \url{https://arxiv.org/abs/2212.09748}.

\bibitem[Podell et~al.(2023)Podell, English, Lacey, Blattmann, Dockhorn,
  Müller, Penna, and Rombach]{podell2023sdxlimprovinglatentdiffusion}
Dustin Podell, Zion English, Kyle Lacey, Andreas Blattmann, Tim Dockhorn, Jonas
  Müller, Joe Penna, and Robin Rombach.
\newblock {SDXL: Improving Latent Diffusion Models for High-Resolution Image
  Synthesis}, 2023.
\newblock URL \url{https://arxiv.org/abs/2307.01952}.

\bibitem[Radford et~al.(2021)Radford, Kim, Hallacy, Ramesh, Goh, Agarwal,
  Sastry, Askell, Mishkin, Clark, Krueger, and
  Sutskever]{radford2021learningtransferablevisualmodels}
Alec Radford, Jong~Wook Kim, Chris Hallacy, Aditya Ramesh, Gabriel Goh,
  Sandhini Agarwal, Girish Sastry, Amanda Askell, Pamela Mishkin, Jack Clark,
  Gretchen Krueger, and Ilya Sutskever.
\newblock {Learning Transferable Visual Models From Natural Language
  Supervision}, 2021.
\newblock URL \url{https://arxiv.org/abs/2103.00020}.

\bibitem[{Robbyant Team} et~al.(2026){Robbyant Team}, Gao, Wang, Zeng, Zhu,
  Cheng, Li, Wang, Xu, Ma, Chen, Liu, Cheng, Yao, Zhu, Meng, Zheng, Bai, Chen,
  Shen, Yu, Zhu, Shen, and
  Ouyang]{robbyantteam2026advancingopensourceworldmodels}
{Robbyant Team}, Zelin Gao, Qiuyu Wang, Yanhong Zeng, Jiapeng Zhu, Ka~Leong
  Cheng, Yixuan Li, Hanlin Wang, Yinghao Xu, Shuailei Ma, Yihang Chen, Jie Liu,
  Yansong Cheng, Yao Yao, Jiayi Zhu, Yihao Meng, Kecheng Zheng, Qingyan Bai,
  Jingye Chen, Zehong Shen, Yue Yu, Xing Zhu, Yujun Shen, and Hao Ouyang.
\newblock {Advancing Open-source World Models}, 2026.
\newblock URL \url{https://arxiv.org/abs/2601.20540}.

\bibitem[Rombach et~al.(2022)Rombach, Blattmann, Lorenz, Esser, and
  Ommer]{rombach2022highresolutionimagesynthesislatent}
Robin Rombach, Andreas Blattmann, Dominik Lorenz, Patrick Esser, and Björn
  Ommer.
\newblock {High-Resolution Image Synthesis with Latent Diffusion Models}, 2022.
\newblock URL \url{https://arxiv.org/abs/2112.10752}.

\bibitem[Russakovsky et~al.(2015)Russakovsky, Deng, Su, Krause, Satheesh, Ma,
  Huang, Karpathy, Khosla, Bernstein, Berg, and
  Fei-Fei]{russakovsky2015imagenetlargescalevisual}
Olga Russakovsky, Jia Deng, Hao Su, Jonathan Krause, Sanjeev Satheesh, Sean Ma,
  Zhiheng Huang, Andrej Karpathy, Aditya Khosla, Michael Bernstein,
  Alexander~C. Berg, and Li~Fei-Fei.
\newblock {ImageNet Large Scale Visual Recognition Challenge}, 2015.
\newblock URL \url{https://arxiv.org/abs/1409.0575}.

\bibitem[Salimans \& Ho(2022)Salimans and
  Ho]{salimans2022progressivedistillationfastsampling}
Tim Salimans and Jonathan Ho.
\newblock {Progressive Distillation for Fast Sampling of Diffusion Models},
  2022.
\newblock URL \url{https://arxiv.org/abs/2202.00512}.

\bibitem[Salimans et~al.(2016)Salimans, Goodfellow, Zaremba, Cheung, Radford,
  and Chen]{salimans2016improvedtechniquestraininggans}
Tim Salimans, Ian Goodfellow, Wojciech Zaremba, Vicki Cheung, Alec Radford, and
  Xi~Chen.
\newblock {Improved Techniques for Training GANs}, 2016.
\newblock URL \url{https://arxiv.org/abs/1606.03498}.

\bibitem[Sauer et~al.(2023)Sauer, Lorenz, Blattmann, and
  Rombach]{sauer2023adversarialdiffusiondistillation}
Axel Sauer, Dominik Lorenz, Andreas Blattmann, and Robin Rombach.
\newblock {Adversarial Diffusion Distillation}, 2023.
\newblock URL \url{https://arxiv.org/abs/2311.17042}.

\bibitem[Schuhmann et~al.(2022)Schuhmann, Beaumont, Vencu, Gordon, Wightman,
  Cherti, Coombes, Katta, Mullis, Wortsman, Schramowski, Kundurthy, Crowson,
  Schmidt, Kaczmarczyk, and Jitsev]{schuhmann2022laion5bopenlargescaledataset}
Christoph Schuhmann, Romain Beaumont, Richard Vencu, Cade Gordon, Ross
  Wightman, Mehdi Cherti, Theo Coombes, Aarush Katta, Clayton Mullis, Mitchell
  Wortsman, Patrick Schramowski, Srivatsa Kundurthy, Katherine Crowson, Ludwig
  Schmidt, Robert Kaczmarczyk, and Jenia Jitsev.
\newblock {LAION-5B: An open large-scale dataset for training next generation
  image-text models}, 2022.
\newblock URL \url{https://arxiv.org/abs/2210.08402}.

\bibitem[Sejdinovic et~al.(2013)Sejdinovic, Sriperumbudur, Gretton, and
  Fukumizu]{sejdinovic2013equivalencedistancebasedrkhsbasedstatistics}
Dino Sejdinovic, Bharath Sriperumbudur, Arthur Gretton, and Kenji Fukumizu.
\newblock {Equivalence of distance-based and RKHS-based statistics in
  hypothesis testing}, 2013.
\newblock URL \url{https://arxiv.org/abs/1207.6076}.

\bibitem[Song et~al.(2023)Song, Dhariwal, Chen, and
  Sutskever]{song2023consistencymodels}
Yang Song, Prafulla Dhariwal, Mark Chen, and Ilya Sutskever.
\newblock {Consistency Models}, 2023.
\newblock URL \url{https://arxiv.org/abs/2303.01469}.

\bibitem[Su et~al.(2026)Su, Li, Xue, Huang, Fu, Li, Li, Qian, Huang, and
  Duan]{su2026omniforcingunleashingrealtimejoint}
Yaofeng Su, Yuming Li, Zeyue Xue, Jie Huang, Siming Fu, Haoran Li, Ying Li,
  Zezhong Qian, Haoyang Huang, and Nan Duan.
\newblock {OmniForcing: Unleashing Real-time Joint Audio-Visual Generation},
  2026.
\newblock URL \url{https://arxiv.org/abs/2603.11647}.

\bibitem[Sun et~al.(2026)Sun, Zhang, Wang, Wu, Wang, Wang, Wang, Zhang, Wang,
  and Guo]{sun2026worldplaylongtermgeometricconsistency}
Wenqiang Sun, Haiyu Zhang, Haoyuan Wang, Junta Wu, Zehan Wang, Zhenwei Wang,
  Yunhong Wang, Jun Zhang, Tengfei Wang, and Chunchao Guo.
\newblock {WorldPlay: Towards Long-Term Geometric Consistency for Real-Time
  Interactive World Modeling}, 2026.
\newblock URL \url{https://arxiv.org/abs/2512.14614}.

\bibitem[Vaswani et~al.(2017)Vaswani, Shazeer, Parmar, Uszkoreit, Jones, Gomez,
  Kaiser, and Polosukhin]{vaswani2023attentionneed}
Ashish Vaswani, Noam Shazeer, Niki Parmar, Jakob Uszkoreit, Llion Jones,
  Aidan~N. Gomez, Lukasz Kaiser, and Illia Polosukhin.
\newblock {Attention Is All You Need}, 2017.
\newblock URL \url{https://arxiv.org/abs/1706.03762}.

\bibitem[Wan et~al.(2025)Wan, Wang, Ai, Wen, Mao, Xie, Chen, Yu, Zhao, Yang,
  Zeng, Wang, Zhang, Zhou, Wang, Chen, Zhu, Zhao, Yan, Huang, Feng, Zhang, Li,
  Wu, Chu, Feng, Zhang, Sun, Fang, Wang, Gui, Weng, Shen, Lin, Wang, Wang,
  Zhou, Wang, Shen, Yu, Shi, Huang, Xu, Kou, Lv, Li, Liu, Wang, Zhang, Huang,
  Li, Wu, Liu, Pan, Zheng, Hong, Shi, Feng, Jiang, Han, Wu, and
  Liu]{wan2025wanopenadvancedlargescale}
Team Wan, Ang Wang, Baole Ai, Bin Wen, Chaojie Mao, Chen-Wei Xie, Di~Chen,
  Feiwu Yu, Haiming Zhao, Jianxiao Yang, Jianyuan Zeng, Jiayu Wang, Jingfeng
  Zhang, Jingren Zhou, Jinkai Wang, Jixuan Chen, Kai Zhu, Kang Zhao, Keyu Yan,
  Lianghua Huang, Mengyang Feng, Ningyi Zhang, Pandeng Li, Pingyu Wu, Ruihang
  Chu, Ruili Feng, Shiwei Zhang, Siyang Sun, Tao Fang, Tianxing Wang, Tianyi
  Gui, Tingyu Weng, Tong Shen, Wei Lin, Wei Wang, Wei Wang, Wenmeng Zhou, Wente
  Wang, Wenting Shen, Wenyuan Yu, Xianzhong Shi, Xiaoming Huang, Xin Xu, Yan
  Kou, Yangyu Lv, Yifei Li, Yijing Liu, Yiming Wang, Yingya Zhang, Yitong
  Huang, Yong Li, You Wu, Yu~Liu, Yulin Pan, Yun Zheng, Yuntao Hong, Yupeng
  Shi, Yutong Feng, Zeyinzi Jiang, Zhen Han, Zhi-Fan Wu, and Ziyu Liu.
\newblock {Wan: Open and Advanced Large-Scale Video Generative Models}, 2025.
\newblock URL \url{https://arxiv.org/abs/2503.20314}.

\bibitem[Wang et~al.(2024)Wang, Huang, Bergman, Shen, Gao, Lingelbach, Sun,
  Bian, Song, Liu, Wang, and Li]{wang2024phasedconsistencymodels}
Fu-Yun Wang, Zhaoyang Huang, Alexander~William Bergman, Dazhong Shen, Peng Gao,
  Michael Lingelbach, Keqiang Sun, Weikang Bian, Guanglu Song, Yu~Liu, Xiaogang
  Wang, and Hongsheng Li.
\newblock {Phased Consistency Models}, 2024.
\newblock URL \url{https://arxiv.org/abs/2405.18407}.

\bibitem[Wang et~al.(2023)Wang, Lu, Wang, Bao, Li, Su, and
  Zhu]{wang2023prolificdreamerhighfidelitydiversetextto3d}
Zhengyi Wang, Cheng Lu, Yikai Wang, Fan Bao, Chongxuan Li, Hang Su, and Jun
  Zhu.
\newblock {ProlificDreamer: High-Fidelity and Diverse Text-to-3D Generation
  with Variational Score Distillation}, 2023.
\newblock URL \url{https://arxiv.org/abs/2305.16213}.

\bibitem[Xu et~al.(2023)Xu, Liu, Wu, Tong, Li, Ding, Tang, and
  Dong]{xu2023imagerewardlearningevaluatinghuman}
Jiazheng Xu, Xiao Liu, Yuchen Wu, Yuxuan Tong, Qinkai Li, Ming Ding, Jie Tang,
  and Yuxiao Dong.
\newblock {ImageReward: Learning and Evaluating Human Preferences for
  Text-to-Image Generation}, 2023.
\newblock URL \url{https://arxiv.org/abs/2304.05977}.

\bibitem[Yang et~al.(2026)Yang, Liu, Li, Feng, Cao, Zhang, and
  Shi]{yang2026incontextforcinguncoveringcontext}
Lingxiao Yang, Liu Liu, Moran Li, Han Feng, Wenjian Cao, Jiangning Zhang, and
  Ye~Shi.
\newblock {In-Context Forcing: Uncovering Context Effects in Autoregressive
  Video Diffusion}, 2026.
\newblock URL \url{https://arxiv.org/abs/2608.05237}.

\bibitem[Yang et~al.(2025)Yang, Huang, Chu, Xiao, Zhao, Wang, Li, Xie, Chen,
  Lu, Han, and Chen]{yang2025longliverealtimeinteractivelong}
Shuai Yang, Wei Huang, Ruihang Chu, Yicheng Xiao, Yuyang Zhao, Xianbang Wang,
  Muyang Li, Enze Xie, Yingcong Chen, Yao Lu, Song Han, and Yukang Chen.
\newblock {LongLive: Real-time Interactive Long Video Generation}, 2025.
\newblock URL \url{https://arxiv.org/abs/2509.22622}.

\bibitem[Yin et~al.(2024{\natexlab{a}})Yin, Gharbi, Park, Zhang, Shechtman,
  Durand, and Freeman]{yin2024improveddistributionmatchingdistillation}
Tianwei Yin, Michaël Gharbi, Taesung Park, Richard Zhang, Eli Shechtman, Fredo
  Durand, and William~T. Freeman.
\newblock {Improved Distribution Matching Distillation for Fast Image
  Synthesis}, 2024{\natexlab{a}}.
\newblock URL \url{https://arxiv.org/abs/2405.14867}.

\bibitem[Yin et~al.(2024{\natexlab{b}})Yin, Gharbi, Zhang, Shechtman, Durand,
  Freeman, and Park]{yin2024onestepdiffusiondistributionmatching}
Tianwei Yin, Michaël Gharbi, Richard Zhang, Eli Shechtman, Fredo Durand,
  William~T. Freeman, and Taesung Park.
\newblock {One-step Diffusion with Distribution Matching Distillation},
  2024{\natexlab{b}}.
\newblock URL \url{https://arxiv.org/abs/2311.18828}.

\bibitem[Yin et~al.(2025)Yin, Zhang, Zhang, Freeman, Durand, Shechtman, and
  Huang]{yin2025slowbidirectionalfastautoregressive}
Tianwei Yin, Qiang Zhang, Richard Zhang, William~T. Freeman, Fredo Durand, Eli
  Shechtman, and Xun Huang.
\newblock {From Slow Bidirectional to Fast Autoregressive Video Diffusion
  Models}, 2025.
\newblock URL \url{https://arxiv.org/abs/2412.07772}.

\bibitem[Yu et~al.(2026)Yu, Huang, Li, Wang, and
  Yamasaki]{yu2026videomiraiautoregressivevideodiffusion}
Yonghao Yu, Lang Huang, Runyi Li, Zerun Wang, and Toshihiko Yamasaki.
\newblock {Video-Mirai: Autoregressive Video Diffusion Models Need Foresight},
  2026.
\newblock URL \url{https://arxiv.org/abs/2606.03971}.

\bibitem[{Z-Image Team} et~al.(2026){Z-Image Team}, Cai, Cao, Du, Gao, Hao,
  Hoi, Hou, Huang, Jiang, Jiang, Jin, Li, Li, Li, Liu, Liu, Wu, Yu, Zhan,
  Zhang, Zhang, Zhou, and Zhou]{imageteam2026zimageefficientimagegeneration}
{Z-Image Team}, Huanqia Cai, Sihan Cao, Ruoyi Du, Peng Gao, Aiming Hao, Steven
  Hoi, Zhaohui Hou, Shijie Huang, Dengyang Jiang, Yuming Jiang, Xin Jin,
  Liangchen Li, Zhen Li, Zhong-Yu Li, David Liu, Dongyang Liu, Qilong Wu, Feng
  Yu, Zechao Zhan, Chi Zhang, Shifeng Zhang, Ruikai Zhou, and Shilin Zhou.
\newblock {Z-Image: An Efficient Image Generation Foundation Model with
  Single-Stream Diffusion Transformer}, 2026.
\newblock URL \url{https://arxiv.org/abs/2511.22699}.

\bibitem[Zhang et~al.(2026)Zhang, Jiang, Chen, Wang, Luo, Wang, Chen, Li, Ye,
  Chen, Zhu, Zhao, Jiang, Huang, Xiang, Zheng, Wang, Wang, Jia, Chen, Chen,
  Jiang, Fu, Deng, Bao, Chen, and Zhu]{zhang2026vidus1realtimeinteractive}
Jintao Zhang, Kai Jiang, Jintao Chen, Xu~Wang, Yang Luo, Yuji Wang, Dechuang
  Chen, Jungang Li, Chengyang Ye, Marco Chen, Hongzhou Zhu, Min Zhao, Yuxuan
  Jiang, Zhengkun Huang, Chendong Xiang, Kaiwen Zheng, Haoxu Wang, Xiaohang
  Wang, Qi~Jia, Xin Chen, Yimin Chen, Youhe Jiang, Fangcheng Fu, Zhijie Deng,
  Fan Bao, Jianfei Chen, and Jun Zhu.
\newblock {Vidu S1: A Real-Time Interactive Video Generation Model}, 2026.
\newblock URL \url{https://arxiv.org/abs/2607.03118}.

\bibitem[Zhao et~al.(2026{\natexlab{a}})Zhao, Wu, Li, Meng, Li, Zhang, Zhou,
  Lin, Gao, Cao, Yan, Peng, Jiang, Li, Tang, Yin, Wu, Xu, Chen, Wang, Shu,
  Zhang, Wang, Chen, Ba, Xu, Wu, Chen, Tang, Zhang, Wang, Liu, Zhou, Yang,
  Cheng, Lv, Liu, Zhou, Xiong, Shi, Wei, Zhao, Liu, Zhang, Zhang, Chen, He,
  Xue, Qu, Tang, Feng, Wu, Sun, Ni, Men, Bai, Zheng, Lan, Zhang, Wen, Wang,
  Qiao, Lu, Zhou, Deng, Xu, Fang, Chen, Wang, Fan, Zhang, Xu, Wu, Ma, and
  Cai]{zhao2026qwenimage20technicalreport}
Bing Zhao, Chenfei Wu, Deqing Li, Hao Meng, Jiahao Li, Jie Zhang, Jingren Zhou,
  Junyang Lin, Kaiyuan Gao, Kuan Cao, Kun Yan, Liang Peng, Lihan Jiang,
  Niantong Li, Ningyuan Tang, Shengming Yin, Tianhe Wu, Xiao Xu, Xiaoyue Chen,
  Xihua Wang, Yan Shu, Yanran Zhang, Yi~Wang, Yilei Chen, Ying Ba, Yixian Xu,
  Yujia Wu, Yuxiang Chen, Zecheng Tang, Zekai Zhang, Zhendong Wang, Zihao Liu,
  Zikai Zhou, An~Yang, Chen Cheng, Chenxu Lv, Dayiheng Liu, Fan Zhou, Hantian
  Xiong, Hongzhu Shi, Hu~Wei, Huihong Zhao, Ivy Liu, Jianwei Zhang, Jiawei
  Zhang, Kai Chen, Kang He, Levon Xue, Lin Qu, Linhan Tang, Luwen Feng,
  Minggang Wu, Minmin Sun, Na~Ni, Rui Men, Shuai Bai, Sishou Zheng, Tao Lan,
  Tianqi Zhang, Tingkun Wen, Wei Wang, Weixu Qiao, Weiyi Lu, Wenmeng Zhou,
  Xiaodong Deng, Xiaoxiao Xu, Xinlei Fang, Xionghui Chen, Yanan Wang, Yang Fan,
  Yichang Zhang, Yixuan Xu, Yu~Wu, Zhiyuan Ma, and Zhizhi Cai.
\newblock {Qwen-Image-2.0 Technical Report}, 2026{\natexlab{a}}.
\newblock URL \url{https://arxiv.org/abs/2605.10730}.

\bibitem[Zhao et~al.(2026{\natexlab{b}})Zhao, Qian, Liang, Kong, Zhang, Huang,
  Fang, Wang, Hui, and Rao]{zhao2026maskforcingimprovingautoregressive}
Zhuoran Zhao, Shengju Qian, Tongtong Liang, Xianghao Kong, Songchun Zhang,
  Junchao Huang, Guian Fang, Xin Wang, Pan Hui, and Anyi Rao.
\newblock {Mask Forcing: Improving Autoregressive Video Diffusion Distillation
  via Dual-Noise Masking Rollout}, 2026{\natexlab{b}}.
\newblock URL \url{https://arxiv.org/abs/2609.09123}.

\bibitem[Zheng et~al.(2026{\natexlab{a}})Zheng, He, Zhao, Zhang, Chen, Chen,
  Lin, Liu, Zhu, and Ma]{zheng2026causalrcmunifiedteacherforcingselfforcing}
Kaiwen Zheng, Guande He, Min Zhao, Jintao Zhang, Huayu Chen, Jianfei Chen,
  Chen-Hsuan Lin, Ming-Yu Liu, Jun Zhu, and Qianli Ma.
\newblock {Causal-rCM: A Unified Teacher-Forcing and Self-Forcing Open Recipe
  for Autoregressive Diffusion Distillation in Streaming Video Generation and
  Interactive World Models}, 2026{\natexlab{a}}.
\newblock URL \url{https://arxiv.org/abs/2606.25473}.

\bibitem[Zheng et~al.(2026{\natexlab{b}})Zheng, Wang, Ma, Chen, Zhang, Balaji,
  Chen, Liu, Zhu, and Zhang]{zheng2026largescalediffusiondistillation}
Kaiwen Zheng, Yuji Wang, Qianli Ma, Huayu Chen, Jintao Zhang, Yogesh Balaji,
  Jianfei Chen, Ming-Yu Liu, Jun Zhu, and Qinsheng Zhang.
\newblock {Large Scale Diffusion Distillation via Score-Regularized
  Continuous-Time Consistency}, 2026{\natexlab{b}}.
\newblock URL \url{https://arxiv.org/abs/2510.08431}.

\bibitem[Zhou et~al.(2024)Zhou, Zheng, Wang, Yin, and
  Huang]{zhou2024scoreidentitydistillationexponentially}
Mingyuan Zhou, Huangjie Zheng, Zhendong Wang, Mingzhang Yin, and Hai Huang.
\newblock {Score identity Distillation: Exponentially Fast Distillation of
  Pretrained Diffusion Models for One-Step Generation}, 2024.
\newblock URL \url{https://arxiv.org/abs/2404.04057}.

\bibitem[Zhou et~al.(2025)Zhou, Gu, and
  Wang]{zhou2025fewstepdiffusionscoreidentity}
Mingyuan Zhou, Yi~Gu, and Zhendong Wang.
\newblock {Few-Step Diffusion via Score identity Distillation}, 2025.
\newblock URL \url{https://arxiv.org/abs/2505.12674}.

\bibitem[Zhu et~al.(2026)Zhu, Zhao, He, Su, Li, and
  Zhu]{zhu2026causalforcingautoregressivediffusion}
Hongzhou Zhu, Min Zhao, Guande He, Hang Su, Chongxuan Li, and Jun Zhu.
\newblock {Causal Forcing: Autoregressive Diffusion Distillation Done Right for
  High-Quality Real-Time Interactive Video Generation}, 2026.
\newblock URL \url{https://arxiv.org/abs/2602.02214}.

\end{thebibliography}
\appendix
\addtocontents{toc}{\protect\setcounter{tocdepth}{2}}
\renewcommand{\contentsname}{Appendix: Table of Contents}
\clearpage
\tableofcontents
\raggedbottom
\makeatletter
\setlength{\@fptop}{0pt}
\setlength{\@fpsep}{12pt}
\setlength{\@fpbot}{0pt plus 1fil}
\makeatother
\section{Background}
\label{sec:background}

\paragraph{Distribution matching.}
Distribution Matching Distillation (DMD) trains an efficient student $G_\theta$ to match the target distribution learned by a frozen diffusion teacher $s_T$. Because the exact score of the student's own generated distribution is unknown, DMD introduces an auxiliary network—the fake score model, or the \emph{critic}—to estimate it online. Throughout the paper, we denote the \emph{sizes} (i.e., parameter counts) of the student, teacher, and critic as $G$, $T$, and $F$, respectively.

The training process operates as follows. First, the student generates a sample $x_0=G_\theta(z,c)$ from noise $z$ and condition $c$. To supervise this prediction, we corrupt it to a \emph{score-evaluation level} $t$ by adding noise: $x_t=\alpha_t x_0+\sigma_t\epsilon$, where $\epsilon\sim\mathcal N(0,I)$. The student is then updated by moving its prediction along the difference between two scores at $x_t$: the teacher's score $s_T$ and the student score $s_q$, which the critic estimates. With exact scores, this gradient minimizes a weighted KL divergence between the student distribution $q_t$ and the teacher distribution $p_t$~\citep{luo2024diffinstructuniversalapproachtransferring,yin2024onestepdiffusiondistributionmatching}. Specifically, the student parameters $\theta$ are updated via:
\begin{equation}
 \Delta\theta\;\propto\;
 \E_{z,c,t,\epsilon}
 \left[w(t)\,\alpha_t\,J_\theta(x_0)^\top
 \bigl(s_T(x_t,t,c)-s_q(x_t,t,c)\bigr)\right],
 \label{eq:dm}
\end{equation}
where $J_\theta(x_0)=\partial x_0/\partial\theta$ is the Jacobian and $w(t)$ is a nonnegative weight. To obtain $s_q$, the critic acts as a denoiser $D_F(x_t,t,c)$ with parameters $\phi$, trained online in alternation with the student to reconstruct the clean student sample $x_0$ from its corruption $x_t$. Its prediction is analytically converted to the score $s_F$, which substitutes $s_q$ in practical DMD~\citep{yin2024improveddistributionmatchingdistillation,yin2024onestepdiffusiondistributionmatching}:
\begin{equation}
 \mathcal L_F=\E\!\left[
 \lambda(t)\|D_F(x_t,t,c)-\sg(x_0)\|^2\right],
 \qquad
 s_F=\frac{\alpha_tD_F-x_t}{\sigma_t^2}.
 \label{eq:critic}
\end{equation}
Here, $\sg$ stops gradients and $\lambda(t)$ weights the fitting loss. (Note that noise- and velocity-prediction networks use their corresponding targets and conversions.) This two-score formulation is also foundational to Variational Score Distillation (VSD) and Score Identity Distillation (SiD)~\citep{wang2023prolificdreamerhighfidelitydiversetextto3d,zhou2024scoreidentitydistillationexponentially,zhou2025fewstepdiffusionscoreidentity}, extending the critic's relevance beyond DMD.

\paragraph{Multi-step students.}
Instead of generating an image in a single pass, a $K$-step student iteratively refines the sample over $K$ distinct calls. During training (e.g., in DMD2), we simulate this sequential process. At step $k$, the student receives an input at noise level $r_k$ and predicts a fully clean sample $x_0^k=G_\theta(x_{r_k},r_k,c)$. To proceed to the next step, this clean prediction is analytically re-noised to a lower noise level $r_{k+1}$, providing the input for step $k+1$. This progressive rollout is known as \emph{backward simulation}.

Importantly, the objective in Equation~\ref{eq:dm} can be used to supervise \emph{every} step $k$. To do so, the step's clean prediction $x_0^k$ is corrupted with fresh noise, $x_t^k=\alpha_t x_0^k+\sigma_t\epsilon$, inducing a distribution $q_t^k$. A crucial distinction here is that this score-evaluation level $t$ is drawn \emph{independently} of the simulation level $r_k$. Consequently, two distinct time variables operate simultaneously: $r_k$ acts as a ``schedule timestamp'' that tracks the step's position within the generation rollout, whereas $t$ determines the intensity of the noise used to evaluate the score-matching gradient from the teacher.

\section{Implementation Details}
\label{app:protocol}

\begin{table}[!htb]
\centering
\setlength{\tabcolsep}{7pt}
\begin{tabular}{@{}lrrrrrrrr@{}}
\toprule
& & & & & & \multicolumn{2}{c}{GFLOPs} & \\
\cmidrule(lr){7-8}
Model & Depth & Width & Heads & Head dim. & Params. (M) & Fwd. & Bwd. & FID$\downarrow$\\
\midrule
S & 12 & 384 & 6 & 64 & 33.0 & 10.9 & 21.8 & 50.99\\
S-B & 12 & 512 & 8 & 64 & 58.3 & 19.4 & 38.8 & 40.64\\
B & 12 & 768 & 12 & 64 & 130.5 & 43.6 & 87.2 & 29.50\\
M & 18 & 896 & 14 & 64 & 264.2 & 89.0 & 178.0 & 20.66\\
L & 24 & 1024 & 16 & 64 & 458.1 & 155.0 & 309.9 & 15.91\\
XL & 28 & 1152 & 16 & 72 & 675.1 & 228.8 & 457.6 & 14.43\\
H & 32 & 1280 & 16 & 80 & 951.3 & 322.8 & 645.6 & 12.56\\
\bottomrule
\end{tabular}
\caption{\textbf{The teacher and critic ladder.} Architecture, forward and backward GFLOPs per sample, and teacher FID of the seven SiT sizes after 800k pretraining updates. Teacher FID falls steadily from S to H. (Settings: ImageNet $256\times256$, w/o CFG; teacher sampling as in Table~\ref{tab:teacher_pretraining}.)}
\label{tab:ladder}
\end{table}

\subsection{Architecture and Teacher Pretraining on ImageNet}
\label{app:ladder}

We use seven sizes, the S, B, L, and XL models of SiT~\citep{ma2024sitexploringflowdiffusionbased} and three added sizes, S-B between S and B, M between B and L, and H above XL. Table~\ref{tab:ladder} gives the full architecture, and Table~\ref{tab:teacher_pretraining} gives the shared pretraining recipe. We conduct all ImageNet experiments at a resolution of $256\times 256$. All models use patch size 2, MLP ratio 4, and $32\times32\times4$ latents from the SD-VAE latent space~\citep{rombach2022highresolutionimagesynthesislatent}. We evaluate metrics on exponential moving average (EMA) checkpoints. The 800k-step training budget is chosen to compare all sizes under one shared recipe rather than to reach state-of-the-art FID.

\begin{table}[!htb]
\centering
\begin{tabular}{@{}>{\raggedright\arraybackslash}p{0.31\linewidth}>{\raggedright\arraybackslash}p{0.63\linewidth}@{}}
\toprule
Setting & Value\\
\midrule
Training objective & Flow Matching~\citep{lipman2023flowmatchinggenerativemodeling}\\
Optimizer & AdamW; $\beta=(0.9,0.999)$; weight decay 0\\
Learning rate & $10^{-4}$, constant; no warmup or schedule\\
Batch size & 256\\
Training length & 800k updates\\
EMA decay & 0.9999\\
Class dropout & 0.1\\
VAE type & \texttt{sd-vae-ft-ema}\\
Teacher sampling & Dormand--Prince (dopri5), 250 steps\\
\bottomrule
\end{tabular}
\caption{\textbf{Hyperparameter choices of ImageNet teacher pretraining.}}
\label{tab:teacher_pretraining}
\end{table}

\FloatBarrier
\subsection{Distillation Implementation on ImageNet}
\label{app:training_query}
\label{app:system_protocols}
\label{app:imagenet_settings}
\paragraph{Pseudocode for one training iteration.}
Algorithm~\ref{alg:training_query} lists one training iteration of our method, with a detached DMD2-style rollout over the generation schedule $r_1,\ldots,r_K$ with step weights $\pi$. On the conditioning side, the critic receives the step $k$ in both updates, whereas the teacher sees the same corrupted prediction without it. The critic update reuses the detached student predictions of the same iteration. Notably, the training procedure decouples the architectures of the student, critic, and teacher, which enables the study of different scaling strategies.

\begin{algorithm}[!htb]
\caption{One DMD training iteration with STAC.}
\label{alg:training_query}
\begin{algorithmic}[1]
\Require student $G_\theta$; frozen teacher denoiser $D_T$; critic $D_F$ with parameters $\phi$; schedule $r_1,\ldots,r_K$; step weights $\pi$
\State Sample condition $c$, step $k\sim\pi$, and initial noise
\State Roll the student out through steps $1,\ldots,k-1$ to obtain $x_{r_k}$
\State $x_0^k\gets G_\theta(x_{r_k},r_k,c)$
\State Sample the score-evaluation level $t$ and $\epsilon\sim\mathcal N(0,I)$;
  $x_t^k\gets\alpha_t x_0^k+\sigma_t\epsilon$
\State Query $D_T(x_t^k,t,c)$ and $D_F(x_t^k,t,k,c)$; form $u$ by Equation~\ref{eq:normalized_update}
\State Update $\theta$ with $\mathcal L_G$ of Equation~\ref{eq:normalized_update}
\State Update $\phi$ with $\mathcal L_F$ of Equation~\ref{eq:critic}, fitting $D_F(x_t^k,t,k,c)$ to $\sg(x_0^k)$
\end{algorithmic}
\end{algorithm}

\paragraph{A pure DMD baseline.}
Table~\ref{tab:pure_baseline} shows the two recipe changes that make the distribution matching (DM) loss of Equation~\ref{eq:normalized_update} alone a strong baseline on ImageNet. We raise the student learning rate to eight times the critic rate and draw the score-evaluation level $t$ from a logit-normal distribution~\citep{esser2024scalingrectifiedflowtransformers}, $t=\operatorname{sigmoid}(n)$ with $n\sim\mathcal N(0,1)$, instead of a uniform one. Each change lowers FID at every size tested, and we keep both as the ImageNet default, listed with the other shared settings in Table~\ref{tab:imagenet_protocol_detail}.

\begin{table}[!htb]
\centering
\setlength{\tabcolsep}{7pt}
\begin{tabular}{@{}lccrrr@{}}
\toprule
& & & \multicolumn{3}{c@{}}{FID$\downarrow$}\\
\cmidrule(l){4-6}
$t$-sampler & $\eta_G$ & $\eta_F$ & S & B & L\\
\midrule
Uniform & $10^{-6}$ & $10^{-6}$ & 51.16 & 30.77 & 12.28\\
Uniform & $8\times10^{-6}$ & $10^{-6}$ & 37.13 & 19.11 & 8.91\\
\rowcolor{ourgray}Logit-normal & $8\times10^{-6}$ & $10^{-6}$ & \textbf{33.96} & \textbf{16.82} & \textbf{8.28}\\
\bottomrule
\end{tabular}
\caption{\textbf{The pure DMD baseline.} FID of 1-step students after 200k updates at three sizes, with student, critic, and teacher of equal size, for the uniform and logit-normal samplers of the score-evaluation level and two student learning rates. Each change lowers FID at every size, and the last row is our default ImageNet recipe. (Settings: ImageNet $256\times256$, w/o CFG, one critic update per student update.)}
\label{tab:pure_baseline}
\end{table}

\paragraph{Learning rates.} Unless stated otherwise, ImageNet runs use the default pair $(\eta_G,\eta_F)=(8\times10^{-6},10^{-6})$ (Table~\ref{tab:pure_baseline}). The teacher--critic grids of Figures~\ref{fig:scaling}b,c and~\ref{fig:full_grids}, and the summaries drawn from them in Figures~\ref{fig:scaling_paths} and~\ref{fig:compute_pareto}, are the exception. For each of their cells separately, we search $\eta_G\in\{4\times10^{-6},8\times10^{-6},1.2\times10^{-5}\}$ and $\eta_F\in\{5\times10^{-7},10^{-6},2\times10^{-6}\}$ and keep the pair with the lowest FID over training. The step tokens of STAC, and the step embedder of the AdaLN-branch control in Table~\ref{tab:cumulative}, are always trained at \emph{20 times} the critic rate $\eta_F$.

\paragraph{Prediction-space implementation.}
As shown in Equation~\ref{eq:dm}, DMD moves the student along the difference between the teacher's and the critic's scores. Substituting the critic's score $s_F$ for $s_q$ and converting both scores to clean predictions by Equation~\ref{eq:critic}, $s_T-s_F=\alpha_t(D_T-D_F)/\sigma_t^2$, the update becomes
\begin{equation}
 \Delta\theta\;\propto\;\E\!\left[w(t)\,\frac{\alpha_t^2}{\sigma_t^2}\,J_\theta(x_0)^\top\bigl(D_T-D_F\bigr)\right].
\end{equation}
This is, up to a constant factor, the negative gradient of a regression of the student prediction onto a detached target,
\begin{equation}
 \mathcal L_G=\tfrac12\operatorname{MSE}\bigl(x_0,\sg(x_0-\tilde u)\bigr),
 \qquad
 \tilde u=w(t)\,\frac{\alpha_t^2}{\sigma_t^2}\,(D_F-D_T),
\end{equation}
since $\nabla_\theta\mathcal L_G\propto J_\theta(x_0)^\top\tilde u$. This $\mathcal L_G$ is the actual code implementation of DMD. In this form the weight $w(t)$ does nothing but rescale the residual $D_F-D_T$ of each sample, so choosing it is essentially choosing a loss weight. Following DMD \citep{yin2024improveddistributionmatchingdistillation,yin2024onestepdiffusiondistributionmatching}, we set $w(t)=\sigma_t^2/(\alpha_t^2\cdot \operatorname{avg}|x_0-D_T|)$, which divides the residual by the teacher's own average absolute residual on the same sample across dimensions and thereby normalizes the update magnitude across score-evaluation levels and stabilizes optimization. The implemented update is therefore
\begin{equation}
 u=\frac{D_F-D_T}{\max(\operatorname{avg}|x_0-D_T|,10^{-6})},
 \qquad
 \mathcal L_G=\tfrac12\operatorname{MSE}
 \left(x_0,\sg(x_0-u)\right),
 \label{eq:normalized_update}
\end{equation}
where $\operatorname{avg}$ averages over the coordinates of each sample, and importantly every quantity forming $u$ is detached. On ImageNet the networks predict velocity on the linear path $x_\tau=\tau x_0+(1-\tau)\epsilon$ with $\tau=1-t$, and $D=x_\tau+(1-\tau)v$ converts each velocity output to the clean prediction used above.

\begin{table}[!htb]
\centering
\begin{tabular}{@{}>{\raggedright\arraybackslash}p{0.25\linewidth}>{\raggedright\arraybackslash}p{0.7\linewidth}@{}}
\toprule
Setting & Value\\
\midrule
Objective & Equation~\ref{eq:normalized_update}; no auxiliary regression or GAN loss\\
Initialization & Pretrained 800k EMA checkpoint of each network's own size\\
Batch size & 64\\
Optimizer & AdamW with $\beta=(0.9,0.999)$\\
Learning rates ($\eta_G$ / $\eta_F$) & $8\times10^{-6}$ / $10^{-6}$ unless stated otherwise\\
Weight decay & 0.01\\
Gradient clipping norm & 10\\
Critic updates per student update & 1\\
Score-evaluation level & $t=\operatorname{sigmoid}(n)$, $n\sim\mathcal N(0,1)$\\
EMA decay & 0.9999\\
Step weights $\pi$ & Uniform over the $K$ steps\\
Step schedules & 2 steps: 999, 499; 4 steps: 999, 666, 333, 124\\
\bottomrule
\end{tabular}
\caption{\textbf{Hyperparameter choices of ImageNet distillation experiments.} Sizes, learning rates, and budgets of individual comparisons are given with each experiment.}
\label{tab:imagenet_protocol_detail}
\end{table}
\FloatBarrier

\paragraph{Evaluation metrics.}
We report Fr\'echet Inception Distance (FID)~\citep{heusel2018ganstrainedtimescaleupdate}
between 50k generated and 50k real images and, where stated, Inception Score
(IS)~\citep{salimans2016improvedtechniquestraininggans} and the improved
precision and recall~\citep{kynkaanniemi2019improvedprecisionrecallmetric} of
the same samples (Tables~\ref{tab:b1_metrics} and~\ref{tab:cumulative}). Unless noted otherwise, we take either the lowest FID over
training or the FID at a fixed number of updates, as indicated for each
experiment, and compute the other metrics at the same checkpoint.

\subsection{Distillation Experiments on SDXL}
\label{app:sdxl_settings}
We conduct our experiments on SDXL 1.0~\citep{podell2023sdxlimprovinglatentdiffusion} in two settings. The first is a pure DMD ablation (Table~\ref{tab:sdxl_conditioning}) that compares how the critic receives the step. The second is our main comparison (Table~\ref{tab:sdxl_systems}), which continues training the official DMD2 checkpoint under DMD2's full recipe with and without STAC. Both settings train on LAION prompts with aesthetic score above 6.25~\citep{schuhmann2022laion5bopenlargescaledataset} at $1024\times1024$ and are evaluated on 10k COCO 2014 prompts~\citep{lin2015microsoftcococommonobjects} at $512\times512$. Table~\ref{tab:sdxl_protocol_detail} lists the hyperparameter settings.

\begin{table}[!htb]
\centering
\setlength{\tabcolsep}{6pt}
\begin{tabular}{@{}>{\raggedright\arraybackslash}p{0.355\linewidth}>{\raggedright\arraybackslash}p{0.30\linewidth}>{\raggedright\arraybackslash}p{0.28\linewidth}@{}}
\toprule
Setting & Pure DMD ablation & \textbf{Main results} \\
\midrule
Step schedule & \multicolumn{2}{>{\centering\arraybackslash}p{\dimexpr0.58\linewidth+2\tabcolsep}@{}}{$\{999,749,499,249\}$} \\
Step input to critic & See Table~\ref{tab:sdxl_conditioning} & 4 in-context tokens (w/~STAC only) \\
Learning rates & \multicolumn{2}{>{\centering\arraybackslash}p{\dimexpr0.58\linewidth+2\tabcolsep}@{}}{Student and critic $5\times10^{-7}$; step tokens $10^{-4}$} \\
Optimizer & \multicolumn{2}{>{\centering\arraybackslash}p{\dimexpr0.58\linewidth+2\tabcolsep}@{}}{AdamW, $\beta=(0.9,0.999)$; gradient clipping at norm 10} \\
Batch size & \multicolumn{2}{>{\centering\arraybackslash}p{\dimexpr0.58\linewidth+2\tabcolsep}@{}}{128} \\
Critic updates per student update & \multicolumn{2}{>{\centering\arraybackslash}p{\dimexpr0.58\linewidth+2\tabcolsep}@{}}{5} \\
EMA & \multicolumn{2}{>{\centering\arraybackslash}p{\dimexpr0.58\linewidth+2\tabcolsep}@{}}{None} \\
Teacher / critic CFG & \multicolumn{2}{>{\centering\arraybackslash}p{\dimexpr0.58\linewidth+2\tabcolsep}@{}}{8.0 / 1.0} \\
Maximum score-evaluation level & \multicolumn{2}{>{\centering\arraybackslash}p{\dimexpr0.58\linewidth+2\tabcolsep}@{}}{0.98} \\
GAN loss weights (student / critic) & None & $5\times10^{-3}$ / $10^{-2}$ \\
Backward simulation & No & Yes \\
Training length & 8k updates & 20k updates \\
\bottomrule
\end{tabular}
\caption{\textbf{Hyperparameter choices of SDXL experiments.} Our core comparison table (Table~\ref{tab:sdxl_systems}) uses the configurations on the right, while ablations in Table~\ref{tab:sdxl_conditioning} use a more lightweight setting on the left.}
\label{tab:sdxl_protocol_detail}
\end{table}

\subsection{Distillation Experiments on FLUX.2}
\label{app:flux_settings}
Training follows the DMD2 recipe~\citep{yin2024improveddistributionmatchingdistillation}, with backward simulation and a diffusion GAN whose discriminator head sits on the critic's backbone, and we use the same setup across all 4B, 9B, and 32B teacher/critic variants in Table~\ref{tab:flux_scaling}.
In all runs, we fully fine-tune a 4B FLUX.2 [klein] base student while keeping each model's default text encoder frozen; the frozen teacher and fully trainable critic are initialized from FLUX.2 [klein] base (4B or 9B) or FLUX.2 [dev] (32B).
Before distillation, we warm up the 4B student model by regressing its $x_0$ prediction with a latent Huber loss onto 10k of its own samples (50 steps, CFG 4.0, $1024\times1024$) re-noised with fresh noise to the four schedule levels.
Distillation at $1024\times1024$ uses the image--composition caption pairs from FLUX-Reason-6M~\citep{fang2025fluxreason6mprismbenchmillionscale}, where the images only serve as the real samples of the GAN branch.
A diffusion GAN discriminator head is attached to the critic's features after its first 8 single-stream blocks and receives inputs noised to $t\sim\mathcal{U}[0,1]$ without time shift.
Notably, with STAC, the step condition enters \emph{only} the critic's score-estimation branch without affecting the discriminator branch, so the discriminator sees real and generated images under the same conditioning.
Further details on the hyperparameters and training protocol are provided in Table~\ref{tab:flux_protocol_detail}.

We evaluate all rows of Table~\ref{tab:flux_scaling} at $1024\times1024$.
HPSv3 is averaged over one sample for each of 10k prompts. We obtain these prompts by recaptioning COCO 2014 validation images~\citep{lin2015microsoftcococommonobjects} with a vision-language model, using captions whose lengths are close to those in the training data.
DPG-Bench~\citep{hu2024ella} uses its official 1,065 prompts with four seeds each.
The students sample with their own 4-step schedules, and the official base models use 50 steps and CFG 4.0.

\begin{table}[!htb]
\centering
\setlength{\tabcolsep}{6pt}
\begin{tabular}{@{}>{\raggedright\arraybackslash}p{0.355\linewidth}>{\raggedright\arraybackslash}p{0.615\linewidth}@{}}
\toprule
Setting & Value \\
\midrule
Step schedule & $\{1000,948,858,668\}$ (time-shifted from $\{1000,750,500,250\}$) \\
Step input to critic & 4 in-context tokens (w/~STAC only) \\
Learning rates & Student and critic $5\times10^{-7}$; step tokens $10^{-5}$ \\
Optimizer & AdamW, $\beta=(0.9,0.999)$, weight decay 0.01; clip norm 10 \\
Batch size & 16 \\
Critic updates per student update & 5 \\
EMA & None \\
Teacher / critic guidance & 2.0 / 2.0 \\
Score-evaluation level (DM loss) & $t\sim\mathcal{U}[0.02,0.98]$, then time-shifted \\
Score-evaluation level (critic loss) & $t=\operatorname{sigmoid}(n)$, $n\sim\mathcal N(0,1)$, clipped to $[0.02,0.98]$, then time-shifted \\
GAN loss weights (student / critic) & $5\times10^{-3}$ / $10^{-2}$ \\
Backward simulation & Yes \\
Training length & 10k iterations (2k student updates) \\
\bottomrule
\end{tabular}
\caption{\textbf{Hyperparameter choices of FLUX.2 experiments} (Table~\ref{tab:flux_scaling}).}
\label{tab:flux_protocol_detail}
\end{table}

\begin{table}[!htb]
\centering
\setlength{\tabcolsep}{6pt}
\begin{tabular}{@{}>{\raggedright\arraybackslash}p{0.355\linewidth}>{\raggedright\arraybackslash}p{0.615\linewidth}@{}}
\toprule
Setting & Value \\
\midrule
Step schedule & Uniform $\{1000,750,500,250\}$, time shift 5 \\
Step input to critic & 32 in-context tokens (w/~STAC only) \\
Learning rates & $\eta_G=2\times10^{-6}$, $\eta_F=4\times10^{-7}$, constant \\
Step-token learning rate & $2\times10^{-4}\to4\times10^{-7}$, cosine decay over iterations 100--500 \\
Optimizer & AdamW, $\beta=(0.0,0.999)$, weight decay 0.01; clip norm 10 \\
Batch size & 64 \\
Critic updates per student update & 5 \\
EMA decay & 0.99 \\
Teacher / critic CFG & 4.0 / 1.0 \\
Score-evaluation level & Equation~\ref{eq:wan_score_time} \\
Backward simulation & Yes (online Self Forcing rollout) \\
Training length & 1.5k iterations \\
\bottomrule
\end{tabular}
\caption{\textbf{Hyperparameter choices of the main Wan2.1 experiments} (Table~\ref{tab:wan_main}).}
\label{tab:wan_protocol_detail}
\end{table}

\subsection{Distillation Experiments on Wan2.1}
\label{app:wan_settings}

We evaluate our method on Wan2.1-T2V \citep{wan2025wanopenadvancedlargescale} by distilling a 4-step, 1.3B student that generates 81 frames at $480\times832$ resolution, evaluated on the official VBench prompts~\citep{huang2023vbenchcomprehensivebenchmarksuite} with five samples per prompt.
Our experiments consist of two settings: (i) ablations of STAC using a 1.3B teacher and critic under bidirectional DMD and autoregressive Self Forcing (Figure~\ref{fig:wan_step_conditioning}, trained for 2k iterations), and (ii) our main scaling comparison varying teacher and critic capacities from 1.3B to 14B (Table~\ref{tab:wan_main}, trained for 1.5k iterations). The STAC ablations initialize the student from the ODE checkpoint released in the CausVid repository~\citep{yin2025slowbidirectionalfastautoregressive}, whereas the scaling comparison uses the Causal Forcing ODE initialization described below. Table~\ref{tab:wan_protocol_detail} summarizes the hyperparameters for the main comparison.

\paragraph{Training and rollout protocol.}
Following Self Forcing~\citep{huang2025selfforcingbridgingtraintest}, we train entirely data-free using prompts sampled from the filtered VidProM list, paired with the standard Wan2.1 negative prompt for the unconditional teacher branch. We optimize the DMD and flow-matching critic objectives. For STAC, we condition the critic on the generation step via 32 learnable in-context tokens.
Let $r=(1000,750,500,250)$ denote the four base generation steps (before a time shift of 5), with $r_5=0$. At each iteration, we uniformly sample a target step $k\in\{1,2,3,4\}$ across all temporal blocks: steps prior to $k$ are rolled out without gradients, and gradients are backpropagated only through step $k$. To evaluate the score difference, we sample an integer $u\sim\mathcal U\{r_{k+1},\ldots,999\}$ and apply a time shift of 5 to obtain the perturbation level:
\begin{equation}
 t=\operatorname{clip}\!\left(1000\frac{5(u/1000)}{1+4(u/1000)},20,980\right).
 \label{eq:wan_score_time}
\end{equation}
For the controlled scaling rows in Table~\ref{tab:wan_main}, the student generates 3-frame latent chunks sequentially with KV-cache context via online Self Forcing rollout~\citep{huang2025selfforcingbridgingtraintest}, while initializing from the chunk-wise Causal ODE checkpoint of Causal Forcing~\citep{zhu2026causalforcingautoregressivediffusion}.

\section{A Two-dimensional Example}
\label{app:toy}

\begin{figure}[!htb]
\centering
\includegraphics[width=\linewidth]{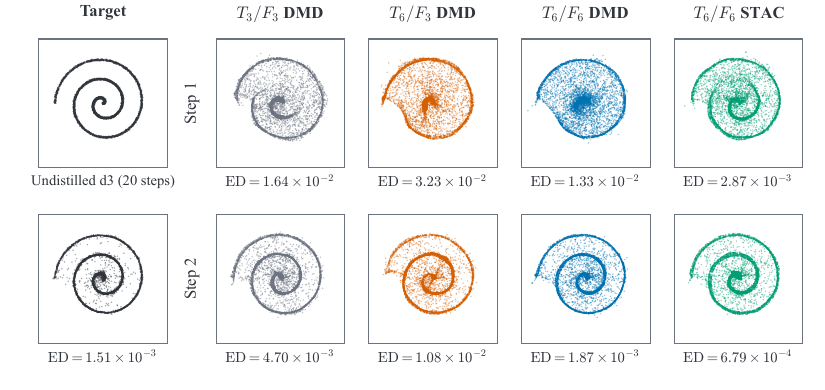}
\caption{\textbf{Joint scaling and STAC on a two-dimensional example.} Samples
of a 2-step, depth-3 student at each sampling step (rows), with energy
distance (ED) to the target. The first column shows the single-roll target
and an undistilled depth-3 teacher sampled with 20 Euler steps. The next
three columns vary the teacher ($T$) and critic ($F$) depths, and the last adds
STAC to the depth-6 pair. Deepening only the teacher makes both steps worse,
deepening the critic as well reverses this, and STAC brings both steps closer
to the single roll. (Settings: 4k DMD updates, five critic updates per student
update; Table~\ref{tab:toy_settings}.)}
\label{fig:toy}
\end{figure}

To illustrate our two core findings---the benefit of joint scaling and the effectiveness of the step-aware critic (STAC)---we construct the two-dimensional toy experiment shown in Figure~\ref{fig:toy}.

\paragraph{Setup.}
The target distribution is a single Swiss roll (2.5 turns, maximum radius 3.0) with Gaussian noise ($\sigma=0.03$). All networks are MLPs, and a \emph{depth-$d$} network has $d$ residual linear-SiLU blocks. The student has depth 3, and the teacher/critic depths are $3/3$, $6/3$, and $6/6$. For STAC, we add a separate sinusoidal step embedder to the critic. Table~\ref{tab:toy_settings} lists the remaining settings.

\begin{table}[!htb]
\centering
\begin{tabular}{@{}>{\raggedright\arraybackslash}p{0.35\linewidth}>{\raggedright\arraybackslash}p{0.60\linewidth}@{}}
\toprule
Setting & Value\\
\midrule
MLP width & 256\\
Teacher training steps & 12k with batch size 1,024\\
DMD training steps & 4k with batch size 256\\
Critic updates per student update & 5\\
Optimizer & Adam with $\beta=(0.9, 0.999)$\\
Teacher learning rate & $10^{-3}$ with cosine decay\\
DMD learning rates & Critic $10^{-3}$, student $10^{-4}$, constant\\
Step noise levels $r_k$ & $\{1,0.5\}$\\
Score-evaluation level & $t\sim\mathcal U(0,1)$\\
\bottomrule
\end{tabular}
\caption{\textbf{Hyperparameter choices of the two-dimensional toy experiment.}}
\label{tab:toy_settings}
\end{table}

\paragraph{Evaluation.}
We evaluate all checkpoints with the same initial noise, re-noising streams, and reference set. Generation quality is measured by the energy distance (ED)~\citep{sejdinovic2013equivalencedistancebasedrkhsbasedstatistics} to the target distribution,
\begin{equation}
\mathrm{ED}(X,Y)=2\E\|X-Y\|-\E\|X-X'\|-\E\|Y-Y'\|,
\end{equation}
where $X,X'$ are independent generated samples and $Y,Y'$ are independent target samples. Lower ED indicates closer agreement with the target.

\paragraph{Results.}
The samples reflect our two main findings. First, regarding network capacity (Figure~\ref{fig:toy}, columns 2--4), starting from a matched $3/3$ pair, deepening only the teacher to $6/3$ makes both steps of the student \emph{worse}. However, deepening the critic as well ($6/6$ pair) reverses this degradation.

Second, the samples also expose the step-mixture issue. For all three pairs without STAC, the first step lies farther from the target than the second, suggesting that the two steps follow different distributions. Adding STAC to the $6/6$ pair (Figure~\ref{fig:toy}, column 5) makes the critic step-aware, and both steps then follow the single Swiss roll more closely, with the weaker first step improving the most.

\section{More Results on Joint Scaling}
\label{app:grids}

\subsection{The Full Teacher--Critic Grid}
\label{app:full_grids}

\begin{figure}[!htb]
\centering
\includegraphics[width=\linewidth]{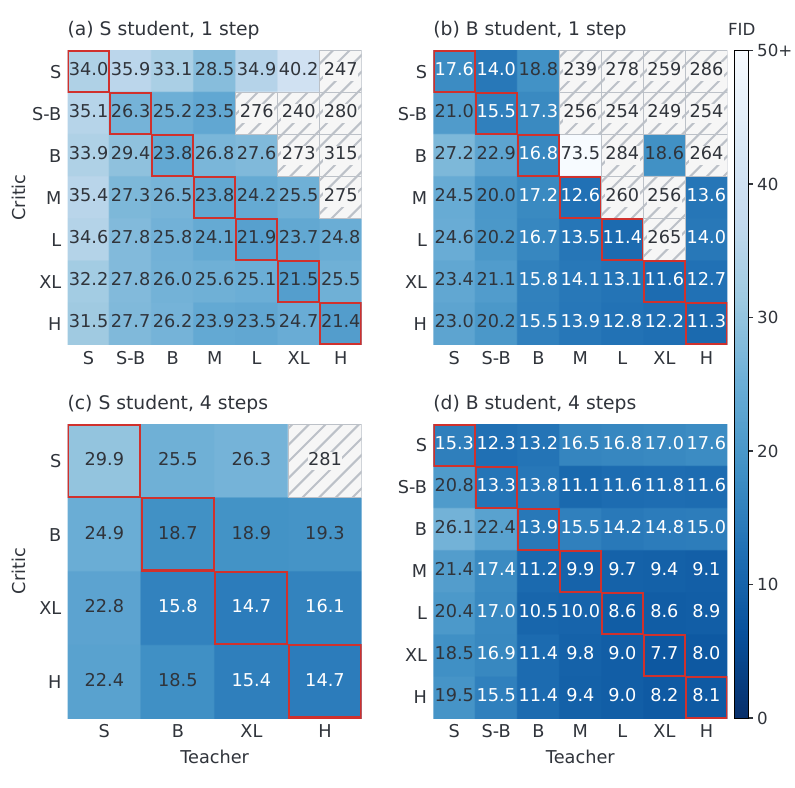}
\caption{\textbf{The expanded teacher--critic scaling grid.} Best FID per teacher/critic pair for S and B students at 1 and 4 steps. Red boxes mark the matched pairs, and hatched cells mark collapsed runs, whose best FID is 100 or higher under all nine learning-rate pairs. Every collapse occurs with a critic smaller than its teacher. (Settings: ImageNet $256\times256$, w/o CFG, 200k updates; learning-rate selection as in Figure~\ref{fig:scaling}b.)}
\label{fig:full_grids}
\end{figure}

\paragraph{Extending the FID grid.}
Figure~\ref{fig:full_grids} extends the teacher--critic scaling grid of Figure~\ref{fig:scaling} to encompass more combinations and student configurations (S and B students, one and four sampling steps). As shown in the figure, a clear \emph{asymmetry} emerges: cells above the diagonal, where the critic is smaller than the teacher, often degrade and sometimes collapse. Conversely, in cells below the diagonal, where the critic is larger than the teacher, no run collapses in our experiments. Empirically, increasing the number of student sampling steps from one to four removes most collapses of the mismatched pairs. These experiments suggest that the critic should \emph{not} be smaller than the teacher, which makes joint scaling a safe default.

\paragraph{More evaluations on quality and coverage.}
To confirm that the benefits of joint scaling extend beyond FID, we evaluate the four cornerstone runs of Figure~\ref{fig:scaling}a (a 1-step B student under B and XL teachers and critics) using Inception Score (IS), precision, and recall. As shown in Table~\ref{tab:b1_metrics}, the jointly scaled XL/XL pair leads on every metric except precision, suggesting better generation quality and coverage.

\paragraph{Training stability.}
Furthermore, tracking the training dynamics reveals a critical stability advantage. Comparing the lowest FID over training against the FID at 100k and 200k updates shows that only the matched pairs reach their lowest FID at the end of training, whereas the mismatched pairs reach it earlier and then degrade. Figure~\ref{fig:imagenet_qualitative} in Appendix~\ref{app:imagenet_visuals} shows the samples at 200k updates, where the XL/B student in particular loses diversity.

\paragraph{Are the collapses merely optimization artifacts?}
A natural question is whether the training collapses observed in mismatched pairs are capacity failures or simply artifacts of suboptimal learning rates.
In Figure~\ref{fig:full_grids}, every marked cell fails to achieve an FID below 100 across all nine learning-rate pairs in our search (Appendix~\ref{app:imagenet_settings}).
Thus, these failures are not specific to a single learning-rate choice within the tested range, even though this does not establish an intrinsic capacity limit or rule out stabilization through other optimization settings.

To isolate capacity from optimization, we increased the critic's learning rate by a factor of 16 for the H/S pair with an S student, so that the small critic can track the student faster. As Table~\ref{tab:lr_ablation} shows, while this larger learning rate avoids the collapse, the resulting student still performs poorly compared to the matched teacher--critic pairs.
Thus, increasing the critic learning rate can prevent collapse in this setting, but does not recover the quality achieved by the matched teacher--critic configuration.

\begin{table}[!htb]
\centering
\setlength{\tabcolsep}{6pt}
\begin{tabular}{@{}lcc>{\columncolor{ourgray}}cc@{}}
\toprule
Teacher / critic & S / S & H / S & H / S & H / H \\
Critic learning rate & searched & searched & $16\times$ default & searched \\
\midrule
S student, 1 step & \underline{33.96} & 247.18 & 52.53 & \textbf{21.41} \\
S student, 4 steps & \underline{29.91} & 281.41 & 42.89 & \textbf{14.65} \\
\bottomrule
\end{tabular}
\caption{\textbf{Forcing a small critic with an extreme learning rate.} Best FID of 1- and 4-step S students under the H teacher with an S critic. We compare the searched learning rates against an aggressively scaled critic learning rate ($16\times$ default), alongside the matched S/S and H/H baselines. Although the larger learning rate prevents the collapse, the generation quality stays far behind that of the matched H/H pair. (Settings: ImageNet $256\times256$, w/o CFG, best FID over training; $16\times$ runs use $(\eta_G,\eta_F)=(8\times10^{-6},1.6\times10^{-5})$; other cells use the best of the nine searched pairs, as in Figure~\ref{fig:full_grids}.)}
\label{tab:lr_ablation}
\end{table}

\subsection{Classifier-free Guidance}
\label{app:guidance}

In this section, we investigate how classifier-free guidance (CFG)~\citep{ho2022classifierfreediffusionguidance} interacts with the DMD framework.

\begin{figure}[!htb]
\centering
\includegraphics[width=\linewidth]{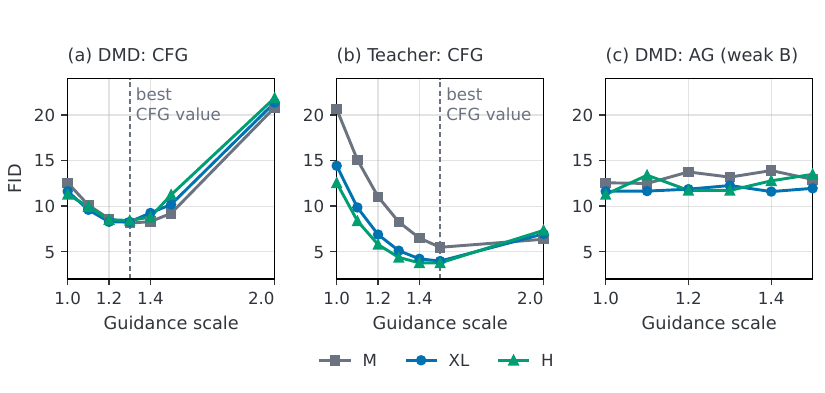}
\caption{\textbf{Guidance for DMD students and their teachers.} FID against guidance scale for (a) 1-step B students distilled under CFG, (b) their teachers sampled with CFG, and (c) students distilled with a weak B model as the autoguidance reference. The students' best CFG scale, about 1.3, lies below the teachers' 1.5, and autoguidance does not improve on CFG. (Settings: ImageNet $256\times256$, $F=T$ for M, XL, and H; students report their best FID over training.)}
\label{fig:guidance_ablation}
\end{figure}

\paragraph{DMD prefers smaller guidance scales.}
We investigate the optimal CFG value for a 1-step B student and for direct sampling from its teacher. We use M, XL, and H teacher--critic pairs, with $F=T$ throughout. As shown in Figure~\ref{fig:guidance_ablation}, the optimal CFG value for the students across sizes is $\sim$1.3, which is smaller than the optimal value for the teachers ($\sim$1.5). We hypothesize that DMD's reverse-KL-style objective may favor high-density regions of the teacher distribution, partially reducing the guidance required by the student. This interpretation is motivated by our ablation rather than established as a general mechanism; related mode-seeking behavior has been discussed in adversarial distribution matching~\citep{lu2025adversarialdistributionmatchingdiffusion}.

Note that in DMD each student guidance scale requires a separate distillation run. Inspired by the explicit guidance-scale conditioning used during from-scratch training in Improved Mean Flows~\citep{geng2026improvedmeanflowschallenges}, a natural extension is to condition the DMD student on the guidance scale during distillation. We leave this continuous guidance conditioning to future work.

\paragraph{Investigating autoguidance in DMD.}
As an alternative to standard CFG, autoguidance replaces the unconditional reference with a weaker model evaluated under the same class condition~\citep{karras2024guidingdiffusionmodelbad}. In DMD, we use the B model as the weak reference for the M, XL, and H teachers. As shown in Figure~\ref{fig:guidance_ablation}c, autoguidance yields no measurable improvement over standard CFG in this distillation setting.

\paragraph{Joint scaling under CFG.}
It is natural to ask whether the benefit of joint scaling, shown without CFG in the main text, persists under CFG. To study this, we fix the CFG scale at 1.2 and the critic at XL, and vary the teacher across the full scaling ladder, for S students at 1, 2, and 4 steps and a B student at 4 steps.

As shown in Table~\ref{tab:cfg_fixed_critic}, in every student configuration the XL teacher, which matches the critic, gives the best student. Scaling the teacher further to H degrades every row, suggesting that joint scaling also holds under CFG.

\begin{table}[!htb]
\centering
\setlength{\tabcolsep}{6pt}
\begin{tabular}{@{}llrrrrrrr@{}}
\toprule
$G$ & Steps & S & S-B & B & M & L & XL & H \\
\midrule
S & 1 & 24.07 & 20.75 & 18.55 & 16.98 & 19.10 & \textbf{16.19} & 20.17 \\
S & 2 & 18.17 & 15.85 & 12.83 & 10.92 & 11.69 & \textbf{10.34} & 11.91 \\
S & 4 & 17.47 & 14.98 & 11.09 & 9.72 & 9.95 & \textbf{8.64} & 10.54 \\
B & 4 & 13.81 & 11.31 & 7.05 & 6.40 & 6.03 & \textbf{5.52} & 5.76 \\
\bottomrule
\end{tabular}

\caption{\textbf{Joint scaling with classifier-free guidance.} Best FID for each teacher (columns) with the critic fixed at XL. Rows vary the student and its step count. Every row is best at the XL teacher, the one matching the critic, and worse again at the larger H teacher. (Settings: ImageNet $256\times256$, CFG $=1.2$; other settings follow Appendix~\ref{app:imagenet_settings}.)}
\label{tab:cfg_fixed_critic}
\end{table}

\subsection{Exploring Teacher Scaling on Other Distillation Methods}
\label{app:other_methods}

To put our scaling experiments in a broader context, we investigate teacher scaling under two alternative distillation methods: velocity knowledge distillation (KD) and MeanFlow distillation (MFD).

\begin{figure}[!htb]
\centering
\includegraphics[width=\linewidth]{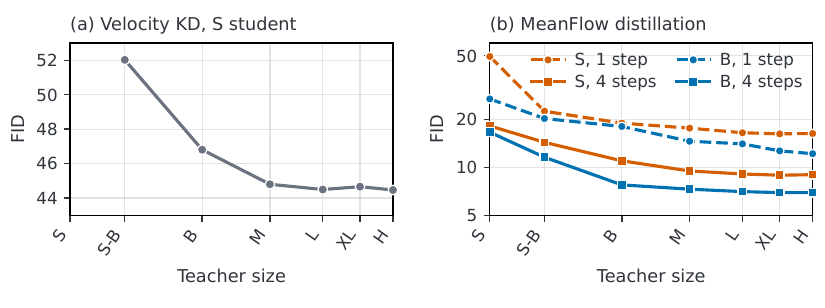}
\caption{\textbf{Teacher scaling under velocity KD and MeanFlow distillation.} Best FID against teacher size for (a) an S student regressed onto each larger teacher's velocity (the S teacher was not evaluated) and (b) S and B students distilled with MeanFlow at 1 and 4 steps. No teacher collapses these students, and the gain from a larger teacher shrinks as it grows. (Settings: ImageNet $256\times256$; KD: learning rate $10^{-4}$, 200k updates; MFD: official MeanFlow recipe with default guidance coefficients, 200k updates for S and 60k for B students.)}
\label{fig:other_methods}
\end{figure}

\paragraph{Knowledge Distillation.} The most straightforward approach to distilling a diffusion teacher is velocity KD, which directly regresses the student's velocity prediction onto that of the frozen teacher. We train an S student using this method from every teacher in our ladder (except S itself) without CFG. As shown in Figure~\ref{fig:other_methods}a, the student's FID improves substantially as the teacher scales up to the M size, but plateaus thereafter. While the student initially benefits from a stronger teacher, this advantage saturates early.

\paragraph{MeanFlow Distillation.} MFD is a few-step distillation framework that learns jumps along the teacher's diffusion trajectory. A MeanFlow model~\citep{geng2025meanflowsonestepgenerative,geng2026improvedmeanflowschallenges,lu2026onesteplatentfreeimagegeneration} predicts the average velocity $u(x_t,r,t)$ over an interval $[r,t]$ and is trained via the identity $u(x_t,r,t)=v(x_t,t)-(t-r)\,\tfrac{\mathrm{d}}{\mathrm{d}t}u(x_t,r,t)$, tying the average velocity to the instantaneous one. During distillation, this instantaneous velocity $v$ is supplied by the frozen teacher rather than the dataset.

We adopt the standard hyperparameter configurations from the official MeanFlow recipe~\citep{geng2025meanflowsonestepgenerative}. Figure~\ref{fig:other_methods}b illustrates the scaling behavior across S and B student configurations, evaluated at 1 and 4 steps. The pattern closely mirrors velocity KD: an oversized teacher never collapses the student, and the marginal gains from scaling the teacher quickly shrink as the teacher grows.

\paragraph{Discussion.} Both KD and MFD exhibit standard scaling behaviors characterized by diminishing returns, contrasting with DMD, where an overly capable teacher can induce training collapse (Section~\ref{sec:scaling}). We attribute this discrepancy to \emph{the nature of the objectives}. KD strictly matches the teacher's velocity, and MFD learns its integral; both are \emph{trajectory-based} methods regressing onto fixed, deterministic targets. Conversely, DMD relies on coupled online student--critic dynamics: the critic must track the shifting student distribution while its score estimate updates the student~\citep{yin2024onestepdiffusiondistributionmatching}. This coupling makes DMD acutely sensitive to critic capacity, so joint network scaling matters in a way that trajectory-based methods bypass.

\section{Exploring the Critic}
\label{app:critic_share}

\subsection{Critic Update Ratio}
\label{app:update_ratio}

\begin{figure}[!htb]
\centering
\includegraphics[width=\linewidth]{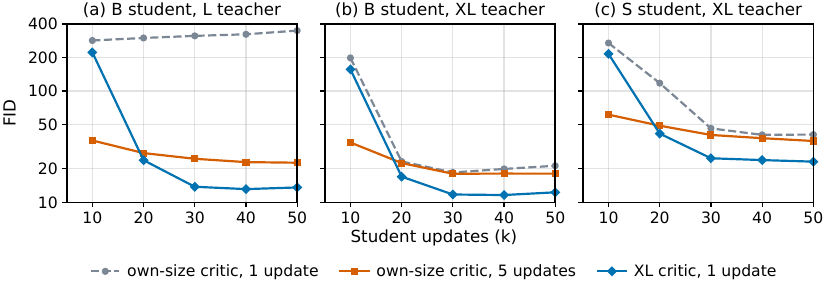}
\caption{\textbf{More critic updates for a small critic.} FID during training with the student-size critic at one and five updates per student update, and with an XL critic at one update, for (a) a B student with an L teacher, (b) a B student with an XL teacher, and (c) an S student with an XL teacher. Five updates speed up early training and avoid the collapse in (a), though the student-size critic stays at a higher FID than the single-update XL critic. (Settings: ImageNet $256\times256$, 1-step students, w/o CFG.)}
\label{fig:update_ratio}
\end{figure}

A common and effective strategy to enhance the critic's ability to track the student distribution is to increase its update frequency, as successfully demonstrated in DMD2~\citep{yin2024improveddistributionmatchingdistillation} with five critic updates per student update. To investigate how this temporal remedy interacts with network capacity, we apply five critic updates per student update to a student-sized critic and compare it with a larger (XL) critic updated only once per step (Figure~\ref{fig:update_ratio}).

As shown in the figure, increasing the update ratio clearly accelerates early training and helps stabilize mismatched pairs (e.g., avoiding the collapse of a B student under an L teacher). However, while more frequent updates allow a small critic to maximize its potential and reach its performance plateau much faster, they do not fully close the capacity gap. In the later stages of training, the larger XL critic, even with a single update, consistently converges to a lower FID. This suggests that while increasing the update ratio is a beneficial optimization strategy, scaling the critic's capacity provides complementary advantages that are crucial for pushing the limits of final generation quality.

\paragraph{Implementation details.}
The student EMA (with decay 0.9999) is updated at every training iteration. Consequently, performing five critic updates per student update shortens the effective EMA horizon from $\sim$10k to $\sim$2k student updates, which shifts the moving average dynamics. To ensure a fair comparison, we swept the EMA decay over $\{0.9999, 0.99998, 0.99999\}$. Furthermore, the learning rates of the student and the critic must be carefully re-balanced. Empirically, raising the critic update ratio from 1 to 5 favors a reduced ratio between the student and critic learning rates (from $\sim$8 down to $\sim$1). For the five-update runs in Figure~\ref{fig:update_ratio}, we use $\eta_G=\eta_F=2\times 10^{-6}$, which aligns with the default setting of DMD2~\citep{yin2024improveddistributionmatchingdistillation}, and the one-update runs use the learning rates of the corresponding grid cells.

\subsection{Critic Warm-up}
\label{app:critic_warmup}

Another intuitive hypothesis for why a small critic collapses under a large teacher (Figure~\ref{fig:scaling}b) is a poor initial fit. Because the critic is initialized from a pretrained checkpoint rather than tailored to the student, early gradient inaccuracies might push the student irreversibly off course. If true, fitting the critic \emph{before} updating the student should prevent the collapse.

\begin{figure}[!htb]
\centering
\includegraphics[width=\linewidth]{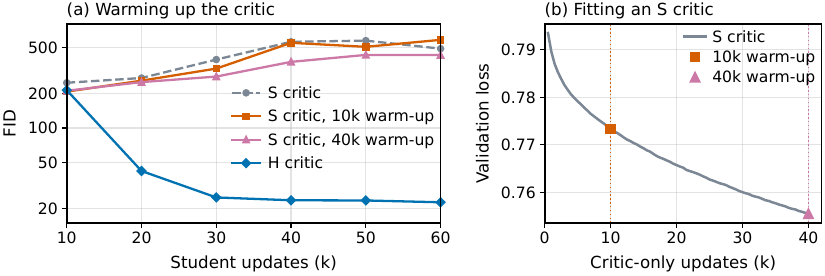}
\caption{\textbf{Critic warm-up.} (a) FID of a 1-step S student under an H teacher with an S critic. The S critic is either trained concurrently from the start or warmed up for 10k/40k updates before student training begins. Both warm-ups fail to prevent collapse compared to the matched H critic. (b) Validation loss of an S critic fitted to a frozen student. (Settings: ImageNet $256\times256$, w/o CFG.)}
\label{fig:critic_warmup}
\end{figure}

To test this, we freeze the student at initialization and ``warm up'' the critic alone for 10k or 40k updates before resuming the standard alternating recipe (Figure~\ref{fig:critic_warmup}b tracks the critic's validation denoising loss). We apply this intervention to the previously unstable configuration of a 1-step S student, an H teacher, and an S critic. As shown in Figure~\ref{fig:critic_warmup}a, neither warm-up duration stabilizes the training.

\paragraph{Implementation details.}
The four runs in Figure~\ref{fig:critic_warmup}a share the setting of the S-student grid: 1-step S student, H teacher, $(\eta_G,\eta_F)=(1.2\times10^{-5},5\times10^{-7})$, batch size 64, one critic update per student update, w/o CFG. The S critic starts from its pretrained checkpoint and trains for 40k updates at a learning rate of $10^{-6}$. The validation loss is measured as the denoising loss on a fixed set of the frozen student's outputs at eight score-evaluation levels.

\subsection{Critic Initialization}
\label{app:initialization}

In our primary experiments, a matched critic always starts as an \emph{exact copy} of the teacher. Consequently, it inherits the teacher's exact initial weights, its level of pretraining convergence, and its intrinsic architectural capacity. To isolate which of these factors is responsible for avoiding the degradation of mismatched pairs, we systematically decouple them.

\paragraph{Do the teacher and critic need to share exact weights?}
We first hypothesize that the critic might rely on sharing the teacher's exact starting weights to effectively track the target distribution. To test this, we distill a 1-step S student using two \emph{independently pretrained} H checkpoints (Model 1 and Model 2). We cross-test them as teachers and critics, removing the shared-weight advantage:

\begin{center}
\setlength{\tabcolsep}{8pt}
\begin{tabular}{@{}lcc@{}}
\toprule
& \multicolumn{2}{c@{}}{Critic}\\
\cmidrule(l){2-3}
Teacher & Model 1 & Model 2\\
\midrule
Model 1 & \cellcolor{ourgray}21.41 & 21.14\\
Model 2 & 21.47 & \cellcolor{ourgray}20.97\\
\bottomrule
\end{tabular}
\end{center}

As shown above (with shaded cells indicating shared initial weights), the off-diagonal pairs with independent weights perform practically identically to the diagonal pairs. This confirms that the critic does \emph{not} need to share the teacher's specific initialization to provide comparable performance.

\paragraph{Does extended pretraining act like larger capacity?}
If intrinsic capacity is the key, how does pretraining duration factor in? One might suspect that a teacher trained for an extraordinarily long time, reaching near-perfect convergence, would act like a ``larger'' model and overwhelm a matched-size critic. To investigate this, we use the official SiT-XL model heavily pretrained for 7M updates~\citep{ma2024sitexploringflowdiffusionbased} (far longer than our 800k updates, with FID 9.35 against 14.43 for ours) and cross-test it against our standard B and XL checkpoints (Table~\ref{tab:critic_checkpoints}). The results help separate architectural capacity from pretraining convergence: under the converged 7M XL teacher, our 800k XL critic stabilizes the student and achieves highly competitive performance. Conversely, a B critic degrades under the 7M XL teacher (FID 35.96).

These results tell a cohesive story: a critic's ability to track the student without degrading is largely \emph{not determined by shared initial weights or the teacher's pretraining duration}. It is instead greatly influenced by the critic's \emph{intrinsic architectural capacity}.

\begin{table}[!htb]
\centering
\setlength{\tabcolsep}{8pt}
\begin{tabular}{@{}lccc@{}}
\toprule
& \multicolumn{3}{c@{}}{Critic} \\
\cmidrule(l){2-4}
Teacher & B & XL & XL, 7M \\
\midrule
B & \cellcolor{ourgray}17.94 & 16.61 & 17.64 \\
XL & 18.56 & \cellcolor{ourgray}13.71 & 11.49 \\
XL, 7M & 35.96 & 12.01 & \multicolumn{1}{>{\columncolor{ourgray}[\tabcolsep][0pt]}c@{}}{11.90} \\
\bottomrule
\end{tabular}
\caption{\textbf{Teachers and critics from different checkpoints.} Best FID of a 1-step B student. Teachers (rows) and critics (columns) come from our B and XL models pretrained for 800k updates, alongside the highly converged XL model from \citet{ma2024sitexploringflowdiffusionbased} trained for 7M updates (XL, 7M). Shaded cells denote shared initial weights. Regardless of the teacher's training duration, matching the critic's architectural size (XL) gives stable and strong performance, whereas a B critic degrades under both XL teachers. (Settings: ImageNet $256\times256$, w/o CFG, default learning rates, 100k updates.)}
\label{tab:critic_checkpoints}
\end{table}

\subsection{Low-rank Adaptation on Critics}
\label{app:lora_critic}

When an undersized critic falls behind a large teacher, it suffers from two separate deficits: a smaller architectural backbone and fewer trainable parameters. This raises a question: is this degradation driven by insufficient \emph{trainable parameters} to track the moving student, or by the limited \emph{expressive capacity} of the smaller backbone?

To decouple parameter count from architectural capacity, we build the critic directly on the frozen XL teacher and train only Low-Rank Adapters (LoRA)~\citep{hu2021loralowrankadaptationlarge}. We inject adapters into the attention projections, MLPs, and AdaLN layers across all blocks, paired with rank-stabilized scaling~\citep{kalajdzievski2023rankstabilizationscalingfactor}. Initializing the adapters to zero ensures the critic starts identically to the teacher, while the rank restricts each adapter update to a low-rank subspace. Crucially, an $r=1$ critic introduces less than 1\% of the trainable parameters of a full B-sized critic, yet it operates entirely within the high-capacity representation space of the XL backbone.

As shown in Table~\ref{tab:lora_critic}, evaluating these configurations on a 1-step B student under an XL teacher (the setting where a full B critic degrades in Figure~\ref{fig:scaling}a) yields two findings:
First, \emph{backbone capacity matters more than the number of trainable parameters}. Every LoRA variant avoids this degradation and outperforms the full B critic, and even at $r=1$, with few trainable parameters, the adapter-based critic remains stable. This suggests that the degradation of small critics stems mainly from their smaller backbones.
Second, \emph{LoRA provides a practical, memory-efficient surrogate for joint scaling}. While a fully fine-tuned XL critic still achieves the lowest FID, low-rank adaptation recovers about 70\% of the joint-scaling gain at a fraction of the optimizer memory. Adapting the frozen teacher thus offers an attractive, parameter-efficient recipe for scaling critics under constrained academic budgets.

\begin{table}[!htb]
\centering
\setlength{\tabcolsep}{6pt}
\begin{tabular}{@{}lrrrr@{}}
\toprule
& Trainable & \multicolumn{3}{c@{}}{Critic learning rate} \\
\cmidrule(l){3-5}
Critic & params (M) & $3\times10^{-5}$ & $10^{-4}$ & $3\times10^{-4}$ \\
\midrule
LoRA on XL, $r=1$ & 0.74 & 14.61 & 13.78 & 14.99 \\
LoRA on XL, $r=4$ & 2.97 & \underline{13.66} & 14.97 & 16.34 \\
LoRA on XL, $r=16$ & 11.87 & 14.82 & 16.92 & 17.28 \\
LoRA on XL, $r=64$ & 47.48 & 13.84 & 17.68 & 17.75 \\
\midrule
Full B (student size) & 130.5 & \multicolumn{3}{c@{}}{18.56} \\
\rowcolor{ourgray}Full XL (matched) & 675.1 & \multicolumn{3}{c@{}}{\textbf{11.63}} \\
Full H & 951.3 & \multicolumn{3}{c@{}}{12.23} \\
\bottomrule
\end{tabular}
\caption{\textbf{Low-rank critics on the frozen teacher.} Best FID of a 1-step B student under an XL teacher. LoRA critics adapt the frozen XL teacher, and the full-parameter critics are the corresponding cells of Figure~\ref{fig:full_grids}b, with the learning rates selected per cell. The best LoRA result is underlined. Even at rank 1, every LoRA critic beats the full B critic, though none reaches the full XL critic. (Settings: ImageNet $256\times256$, w/o CFG; LoRA runs: $\eta_G=1.2\times10^{-5}$, 100k updates, batch size 64, scaling $\alpha/\sqrt{r}$ with $\alpha=8$, zero-initialized up-projection.)}
\label{tab:lora_critic}
\end{table}

\section{More Results on STAC}
\label{app:conditional}

\subsection{Pooled and Stepwise Objectives}

In standard multi-step DMD, the critic estimates the score of a mixture over student generation steps. STAC instead conditions the critic on the generation step. To motivate this design, we compare idealized pooled and stepwise KL objectives.

For a fixed condition $c$ and score-evaluation level $t$, let $q_k=q_t^k$ be the distribution of step $k$'s re-noised prediction, $p=p_t$ the target, and $\bar q=\sum_k\pi_k q_k$ their pool, where $\pi_k$ are the fixed sampling probabilities. Assuming densities exist, the stepwise and pooled KL divergences are related by
\begin{align}
\sum_k\pi_k\KL(q_k\|p)
&=\sum_k\pi_k\int q_k(x) \log\frac{q_k(x)}{\bar q(x)}\,dx + \sum_k\pi_k\int q_k(x) \log\frac{\bar q(x)}{p(x)}\,dx\nonumber\\
&=I(K;X_t)+\KL(\bar q\|p),
\label{eq:pooled_identity}
\end{align}
where $K \sim \pi$ and $X_t \mid K=k \sim q_k$. The stepwise objective (left side) matches every step to the target individually. It differs from the pooled objective (last term) by the mutual information $I(K;X_t)$ between the step index and the re-noised prediction.
This identity provides a distribution-level motivation for STAC. Our practical update uses estimated scores, sample-dependent normalization, and detached rollouts.

\subsection{Step Information and Optimization Stability}
\label{app:step_info}

Equation~\ref{eq:pooled_identity} implies that the stepwise and pooled objectives differ only when the predictions from different steps are distinguishable, that is, when $I(K;X_t)>0$. To measure this, we trained a classifier to predict the step index $k$ from the re-noised prediction $x_t^k$ of two 4-step B students, one without and one with STAC.

\begin{figure}[!htb]
\centering
\includegraphics[width=\linewidth]{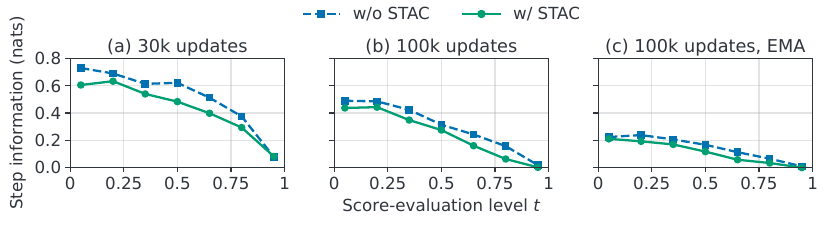}
\caption{\textbf{Step information in re-noised predictions.} Lower bound on the mutual information $I(K;X_t)$ from a classifier predicting the step index of a 4-step B student, evaluated at (a) 30k and (b) 100k updates with online weights, and (c) at 100k updates with EMA weights. The step is readable at low and moderate score-evaluation levels, and STAC lowers this classifier-derived bound at nearly every level. (Settings: ImageNet $256\times256$, XL teacher and critic, w/o CFG; 10k fixed noise--class pairs, 8k/2k training/held-out split.)}
\label{fig:step_separability}
\end{figure}

As shown in Figure~\ref{fig:step_separability}, the classifier identifies the step well above chance at low and moderate score-evaluation levels, suggesting a clear distribution shift across generation steps.
Moreover, STAC yields a lower classifier-derived bound on step
information at nearly every level. This is consistent with
reduced distinguishability across steps, but does not establish
a decrease in the true mutual information.

\begin{figure}[!htb]
\centering
\includegraphics[width=\linewidth]{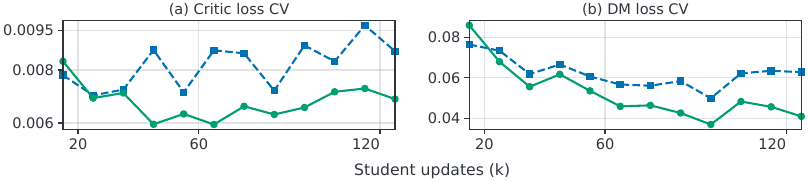}
\caption{\textbf{Loss variability with and without STAC.} Coefficients of variation of (a) the critic loss and (b) the DM loss during the training of a 4-step student. STAC (solid) has a lower CV than the step-agnostic baseline (dashed) at every evaluation after the first. (Settings: XL student, teacher, and critic, ImageNet $256\times256$, CFG $=1.3$.)}
\label{fig:stac_cv}
\end{figure}

\paragraph{Loss variability.}
A single step-agnostic critic has to fit a mixture of step distributions whose composition shifts as the student changes. STAC gives the critic the step index, so it can model each step's distribution separately. Consistent with this, STAC lowers the coefficient of variation (CV) of both the critic loss and the distribution matching (DM) loss at every evaluation after the first (Figure~\ref{fig:stac_cv}). Together with its FID gains in the same runs (Figure~\ref{fig:stac_dynamics}), this suggests that a step-aware critic also makes training smoother.

\subsection{Ablations on Step Conditioning for SDXL}
\label{sec:external_implementation}

Our ImageNet experiments (Table~\ref{tab:cumulative}) show that how the generation step enters the critic matters: in-context tokens outperform an AdaLN branch~\citep{peebles2023scalablediffusionmodelstransformers}. Here, we extend this comparison to the UNet-based SDXL backbone~\citep{podell2023sdxlimprovinglatentdiffusion}. Because the SDXL UNet incorporates text prompts via cross-attention~\citep{rombach2022highresolutionimagesynthesislatent}, we compare injecting the step index via adaptive group normalization (AdaGN)~\citep{dhariwal2021diffusionmodelsbeatgans} (the UNet counterpart of AdaLN) against appending learnable step tokens to the cross-attention text sequence (see Appendix~\ref{app:sdxl_settings} for full settings). As shown in Table~\ref{tab:sdxl_conditioning}, the results mirror our ImageNet findings: four in-context step tokens achieve the best FID, whereas AdaGN yields only a marginal improvement over the step-agnostic baseline, and increasing the token count to 8 or 16 brings diminishing returns.

\begin{center}
\begin{minipage}{\linewidth}
\centering
\makeatletter\def\@captype{table}\makeatother
\setlength{\tabcolsep}{5pt}
\begin{tabular}{@{}llcrrr@{}}
\toprule
Conditioning & Channel & Tokens & FID$\downarrow$ & CLIP-S$\uparrow$ & IS$\uparrow$ \\
\midrule
None & None & 0 & 34.71 & 32.65 & 34.78 \\
Step index & AdaGN & 0 & 34.41 & 32.65 & 36.25 \\
\rowcolor{ourgray}Step tokens & Cross-attention & 4 & \textbf{32.73} & 32.64 & 35.29 \\
Step tokens & Cross-attention & 8 & 33.72 & 32.59 & 37.15 \\
Step tokens & Cross-attention & 16 & 34.01 & 32.60 & 36.51 \\
\bottomrule
\end{tabular}
\caption{\textbf{Step conditioning of the SDXL critic.} Four cross-attention step tokens yield the best FID, outperforming both AdaGN and larger token counts (8 or 16). (Settings: SDXL student and critic, 4 steps, 8k updates, 10k COCO prompts; pure DMD without GAN loss.)}
\label{tab:sdxl_conditioning}
\end{minipage}
\end{center}

\paragraph{Implementation of step tokens.}
Across all backbones, we construct the $N$ step tokens ($N=4$ for SiT, SDXL, and FLUX.2; $N=32$ for Wan2.1) using a shared parameterization. The scalar step $r_k$ is mapped to a 256-dim sinusoidal embedding~\citep{vaswani2023attentionneed} and projected to the context dimension $D$ via a two-layer SiLU MLP. We then add this shared step embedding to $N$ learnable token embeddings $\{\mathbf{e}_n\}_{n=1}^N \subset \mathbb{R}^D$, forming $\mathbf{t}_n = \mathbf{e}_n + \operatorname{MLP}(\operatorname{sinu}(r_k))$, and append $\{\mathbf{t}_n\}_{n=1}^N$ to the conditioning sequence. These randomly initialized parameters are trained alongside a pretrained critic. For SiT, SDXL, and FLUX.2, we assign them a higher learning rate ($20\times$ to $200\times$ the critic backbone rate). For Wan2.1, the step-token rate follows a cosine decay from $2\times10^{-4}$ to $4\times10^{-7}$ over iterations 100--500. See Appendices~\ref{app:ladder}--\ref{app:wan_settings} for details.

\subsection{Cost Analysis}
\label{sec:external_overhead}

\paragraph{Zero inference overhead.}
Because STAC modifies only the critic, which is discarded after training, the deployed student keeps its architecture and sampling procedure. STAC therefore adds \emph{no computational cost at inference}.

\paragraph{Negligible training overhead.}
During training, the step tokens and their small projection MLP add little overhead. In the SDXL and Wan2.1 experiments (Sections~\ref{sec:external_sdxl} and~\ref{sec:external_wan}), STAC adds less than 0.2\% to the critic's parameters and at most 0.7\% to peak memory (Table~\ref{tab:external_overhead}), with no measurable change in iteration time.

\paragraph{Separate critics per step.}
A direct alternative to STAC is to train $K$ separate step-specific critics. However, this multiplies the critic's parameters and optimizer state by $K$. Assuming roughly 16 bytes per parameter for weights, gradients, and optimizer moments, four step-specific SDXL critics would need about 166\,GB of training memory, compared with 42\,GB for a single critic with STAC. For the 14B Wan2.1 critic, the requirement grows to 0.9\,TB against 224\,GB, which makes this design far more costly to train.

\begin{center}
\begin{minipage}{\linewidth}
\centering
\makeatletter\def\@captype{table}\makeatother
\begin{tabular}{@{}lrrr@{}}
\toprule
System & Added parameters & Iteration time & Peak memory \\
\midrule
SDXL & $+0.18\%$ & $\approx0\%$ & $+0.70\%$ \\
Wan2.1, autoregressive & $+0.19\%$ & $\approx0\%$ & $+0.15\%$ \\
\bottomrule
\end{tabular}
\caption{\textbf{Cost analysis of STAC on SDXL and Wan2.1.} Changes relative to each system's baseline without STAC, with the inference architecture unchanged. STAC adds a negligible amount of memory and iteration time.}
\label{tab:external_overhead}
\end{minipage}
\end{center}

\clearpage
\section{More Visualizations}
\label{app:visuals}

\subsection{Qualitative Results on ImageNet}
\label{app:imagenet_visuals}
Figure~\ref{fig:imagenet_qualitative} shows samples of the four 1-step B students of Figure~\ref{fig:scaling}a and Table~\ref{tab:b1_metrics} at 200k updates, next to the B teacher, whose checkpoint also initializes every student. Compared with the B/B pair, scaling both the teacher and the critic to XL gives more coherent objects, whereas scaling only the critic leaves the sample quality largely unchanged. In contrast, the XL/B student produces nearly a single image per class, sometimes with distorted colors or textures.

\par\noindent\begin{minipage}{\linewidth}
\begingroup
\centering
\makeatletter\def\@captype{figure}\makeatother
\captionsetup{skip=4pt}
\setlength{\tabcolsep}{1.5pt}
\renewcommand{\arraystretch}{0.3}
\definecolor{imnetgood}{HTML}{009E73}
\definecolor{imnetflat}{HTML}{C98A00}
\definecolor{imnetbad}{HTML}{C62828}
\newcommand{\imnet}[2]{\includegraphics[width=\linewidth]{figures/imagenet_samples_c#1_#2.jpg}}
\newcommand{\imnethead}[3]{\makebox[\linewidth]{\shortstack{\footnotesize #1\\[1pt]\scriptsize FID #2\\[1pt]\scriptsize #3}}}
\begin{tabular}{@{}>{\centering\arraybackslash}m{.03\linewidth}>{\centering\arraybackslash}m{.145\linewidth}@{\hspace{3.5pt}{\color{black!45}\vrule width .5pt}\hspace{3.5pt}}*{4}{>{\centering\arraybackslash}m{.145\linewidth}}@{}}

 & \imnethead{B teacher}{29.50}{\strut} & \imnethead{B/B}{16.82}{baseline}
 & \imnethead{XL/XL}{13.50}{\textcolor{imnetgood}{\ding{51}\,visible improvement}}
 & \imnethead{B/XL}{19.12}{\textcolor{imnetflat}{little change}}
 & \imnethead{XL/B}{41.75}{\textcolor{imnetbad}{\ding{55}\,low diversity}} \\[12pt]

\rotatebox{90}{\footnotesize Bald eagle} &
\imnet{022}{teacher_B} & \imnet{022}{B_B} & \imnet{022}{XL_XL} & \imnet{022}{B_XL} & \imnet{022}{XL_B} \\[2pt]
\rotatebox{90}{\footnotesize Corgi} &
\imnet{263}{teacher_B} & \imnet{263}{B_B} & \imnet{263}{XL_XL} & \imnet{263}{B_XL} & \imnet{263}{XL_B} \\[2pt]
\rotatebox{90}{\footnotesize Monarch} &
\imnet{323}{teacher_B} & \imnet{323}{B_B} & \imnet{323}{XL_XL} & \imnet{323}{B_XL} & \imnet{323}{XL_B} \\[2pt]
\rotatebox{90}{\footnotesize Husky} &
\imnet{250}{teacher_B} & \imnet{250}{B_B} & \imnet{250}{XL_XL} & \imnet{250}{B_XL} & \imnet{250}{XL_B} \\[2pt]
\rotatebox{90}{\footnotesize Lion} &
\imnet{291}{teacher_B} & \imnet{291}{B_B} & \imnet{291}{XL_XL} & \imnet{291}{B_XL} & \imnet{291}{XL_B} \\[2pt]
\rotatebox{90}{\footnotesize Robin} &
\imnet{015}{teacher_B} & \imnet{015}{B_B} & \imnet{015}{XL_XL} & \imnet{015}{B_XL} & \imnet{015}{XL_B} \\[2pt]
\rotatebox{90}{\footnotesize Red panda} &
\imnet{387}{teacher_B} & \imnet{387}{B_B} & \imnet{387}{XL_XL} & \imnet{387}{B_XL} & \imnet{387}{XL_B} \\
\end{tabular}
\par
\caption{\textbf{\emph{Uncurated} ImageNet 256$\times$256 samples of 1-step B students.} Each $3\times3$ grid shows nine samples of one selected class. The first column is the B teacher without guidance, with its FID from Table~\ref{tab:ladder}, and the other columns are the students of Figure~\ref{fig:scaling}a at 200k updates, labeled by their teacher/critic pair, their FID (Table~\ref{tab:b1_metrics}), and the change from B/B. Scaling both networks (XL/XL) gives more coherent objects than B/B, and scaling only the critic (B/XL) changes little, whereas scaling only the teacher (XL/B) produces near-duplicate samples within each class, indicating low diversity. (Settings: w/o CFG.)}
\label{fig:imagenet_qualitative}
\endgroup
\end{minipage}
\par

\vspace{1.8ex plus .5ex minus .2ex}%
\par\noindent\begin{minipage}{\linewidth}
\subsection{Qualitative Results on FLUX.2}
\label{app:flux_visuals}

Figure~\ref{fig:flux_qualitative} provides visual comparisons of 4-step FLUX.2 4B students across different distillation configurations. We note that, because DMD matches the teacher's output distribution without following its sampling trajectory, the layouts of the student samples differ from those of the teacher samples.
\par\vspace{10pt}

\begingroup
\centering
\makeatletter\def\@captype{figure}\makeatother
\captionsetup{skip=9pt}
\setlength{\tabcolsep}{2pt}
\renewcommand{\arraystretch}{0.3}
\newcommand{\fluxcell}[2]{\includegraphics[width=\linewidth]{figures/flux_samples_#1_#2.jpg}}
\newcommand{\fluxrow}[1]{\fluxcell{#1}{teacher_4b} & \fluxcell{#1}{teacher_9b} & \fluxcell{#1}{t4f4} & \fluxcell{#1}{t9f4} & \fluxcell{#1}{t9f9} & \fluxcell{#1}{t9f9_stac}}
\newcommand{\fluxprompt}[1]{\multicolumn{6}{@{}c@{}}{\parbox{0.8\linewidth}{\centering\scriptsize #1}}}
\newcommand{\fluxhead}[2]{\shortstack{\footnotesize #1\\[1pt]\scriptsize #2}}
\begin{tabular}{@{}*{2}{>{\centering\arraybackslash}m{.145\linewidth}}@{\hspace{3.5pt}{\color{black!45}\vrule width .5pt}\hspace{3.5pt}}*{4}{>{\centering\arraybackslash}m{.145\linewidth}}@{}}

\fluxhead{4B teacher}{} & \fluxhead{9B teacher}{} & \fluxhead{4B/4B}{} & \fluxhead{9B/4B}{} & \fluxhead{9B/9B}{} & \fluxhead{9B/9B w/ STAC}{} \\[6pt]

\fluxrow{animal_texture_10} \\\noalign{\vskip 3pt}
\fluxprompt{A \textbf{horse} galloping along a beach at sunrise, mane flowing, water splashing under its hooves} \\[10pt]
\fluxrow{color_attr_10} \\\noalign{\vskip 3pt}
\fluxprompt{A \textbf{golden crown} resting on a purple velvet cushion, dark background, dramatic spotlight} \\[10pt]
\fluxrow{style_02} \\\noalign{\vskip 3pt}
\fluxprompt{A cute 3D render of a tiny astronaut riding a \textbf{snail} on the moon, soft studio lighting, pastel colors} \\[10pt]
\fluxrow{scene_light_00} \\\noalign{\vskip 3pt}
\fluxprompt{A cozy \textbf{bookstore} interior at night, warm lamps, tall wooden shelves filled with books, a cat asleep on a reading chair, rain outside the window} \\[10pt]
\fluxrow{knowledge_03} \\\noalign{\vskip 3pt}
\fluxprompt{A Scottish \textbf{highland cow} with long shaggy orange hair and curved horns standing in a misty glen} \\[10pt]
\fluxrow{style_08} \\\noalign{\vskip 3pt}
\fluxprompt{A \textbf{stained glass} window design depicting a phoenix rising from flames, vivid reds and golds} \\
\end{tabular}
\par
\caption{\textbf{FLUX.2 samples of 4-step 4B students across teacher/critic pairs.} Each row shares a prompt (listed below) and initial noise. The first two columns display the 50-step 4B and 9B teachers for reference. The subsequent columns show 4-step students distilled under the 4B/4B, 9B/4B, and 9B/9B teacher/critic pairs, alongside the 9B/9B pair combined with STAC. (Settings: students initialized from FLUX.2 [klein] base 4B, 4 steps; teachers sampled with 50 Euler steps at CFG 2.0; resolution $512\times512$.)}
\label{fig:flux_qualitative}
\endgroup
\end{minipage}
\par

\vspace{1.8ex plus .5ex minus .2ex}%
\par\noindent\begin{minipage}{\linewidth}
\subsection{Qualitative Results on SDXL}
\label{sec:external_qualitative}

\vspace{4pt}
\begingroup
\centering
\makeatletter\def\@captype{figure}\makeatother
\captionsetup{skip=9pt}
\setlength{\tabcolsep}{2pt}
\renewcommand{\arraystretch}{0.3}
\newcommand{\sdxlprompt}[1]{\multicolumn{4}{@{}c@{}}{\parbox{0.92\linewidth}{\centering\scriptsize #1}}}
\begin{tabular}{@{}*{4}{>{\centering\arraybackslash}m{.23\linewidth}}@{}}

\footnotesize PCM & \footnotesize SDXL-Lightning
 & \footnotesize DMD2 & \footnotesize DMD2 + STAC (\textbf{Ours}) \\[4pt]

\includegraphics[width=\linewidth]{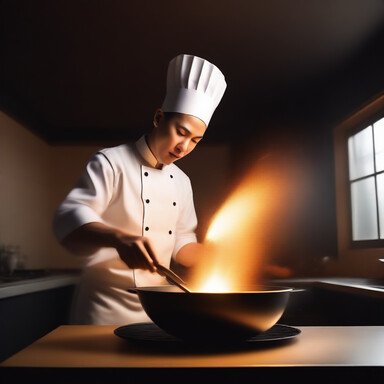} &
\includegraphics[width=\linewidth]{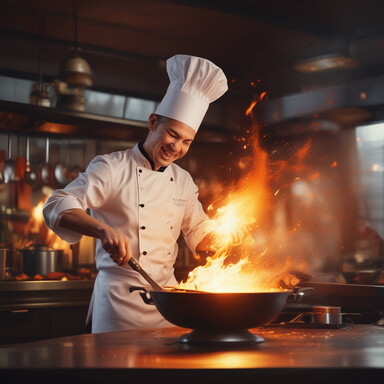} &
\includegraphics[width=\linewidth]{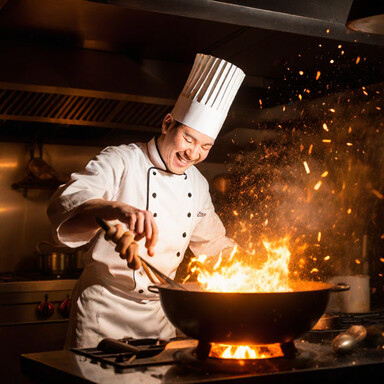} &
\includegraphics[width=\linewidth]{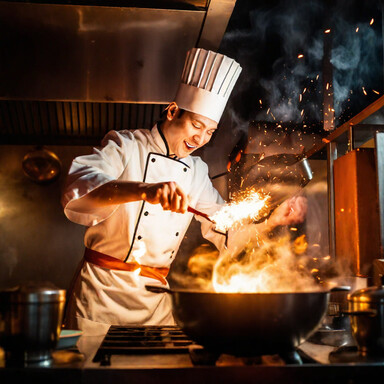} \\\noalign{\vskip 3pt}
\sdxlprompt{\textbf{A joyful chef} in whites tossing flames in a wok, sparks flying, dark restaurant kitchen, dramatic orange light} \\[10pt]
\includegraphics[width=\linewidth]{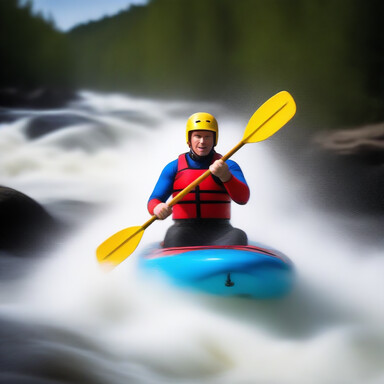} &
\includegraphics[width=\linewidth]{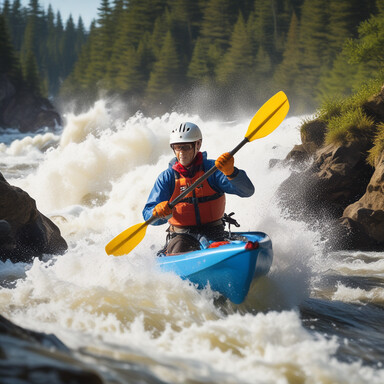} &
\includegraphics[width=\linewidth]{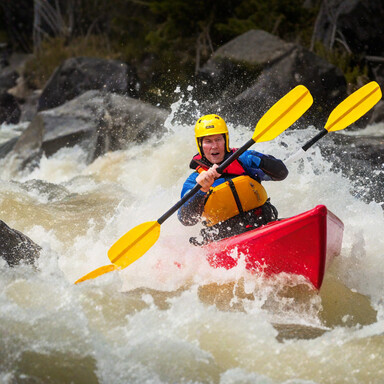} &
\includegraphics[width=\linewidth]{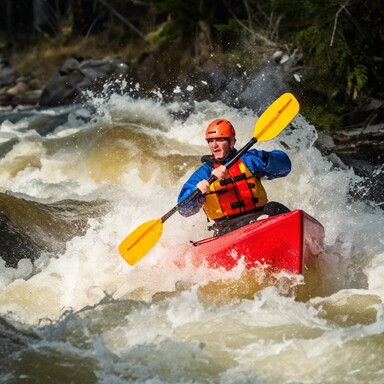} \\\noalign{\vskip 3pt}
\sdxlprompt{\textbf{A kayaker} paddling through white-water rapids, splashing spray, helmet and gear, high-energy adventure} \\[10pt]
\includegraphics[width=\linewidth]{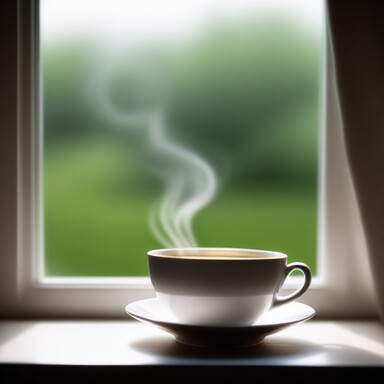} &
\includegraphics[width=\linewidth]{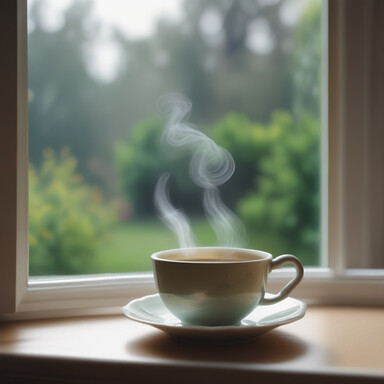} &
\includegraphics[width=\linewidth]{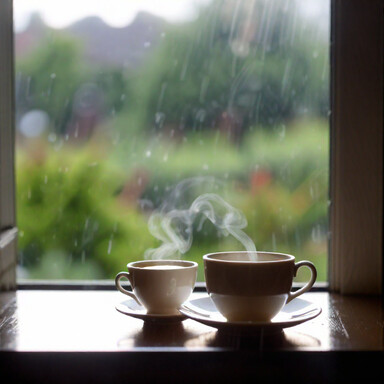} &
\includegraphics[width=\linewidth]{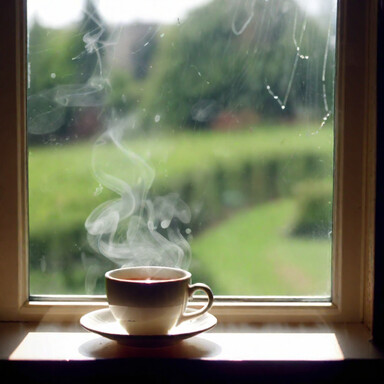} \\\noalign{\vskip 3pt}
\sdxlprompt{\textbf{A hot cup of tea} with steam forming a delicate swirl, on a windowsill overlooking a rainy garden, soft light, tranquil} \\[10pt]
\includegraphics[width=\linewidth]{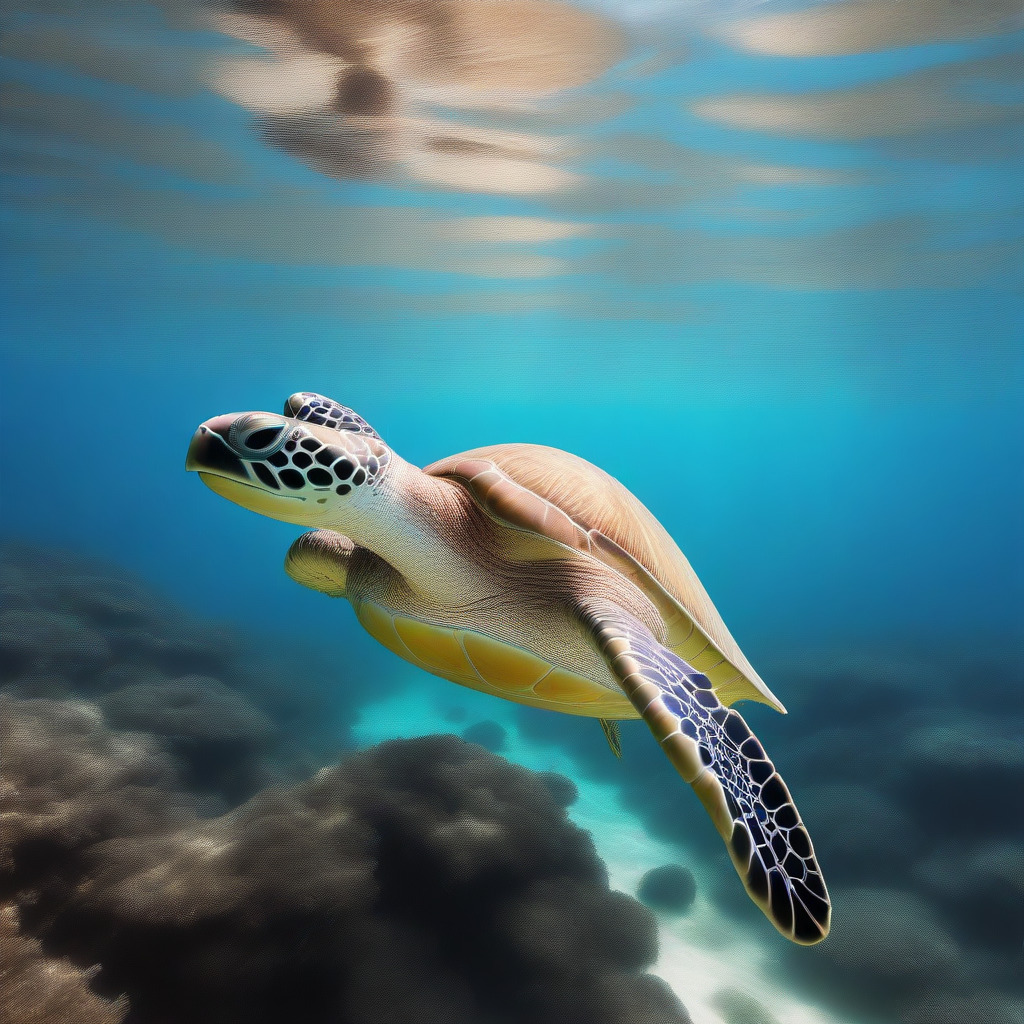} &
\includegraphics[width=\linewidth]{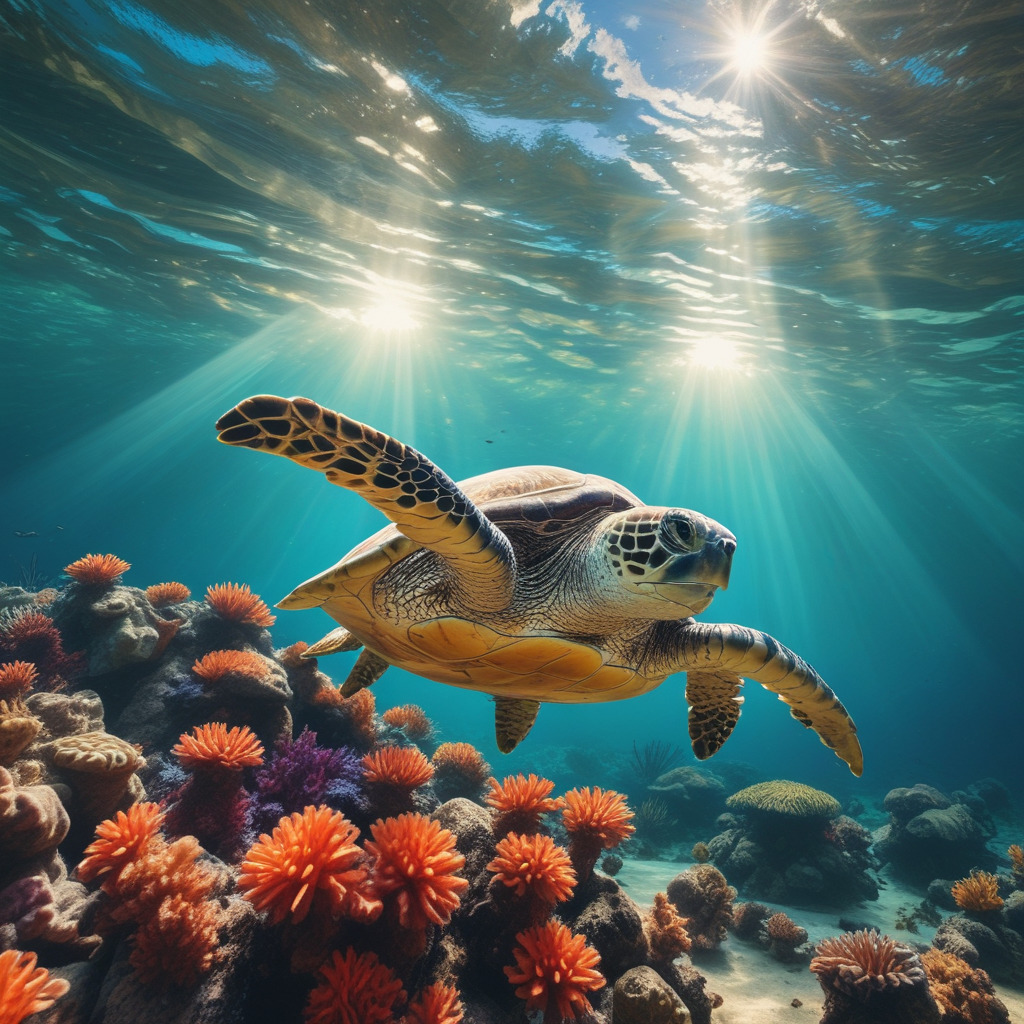} &
\includegraphics[width=\linewidth]{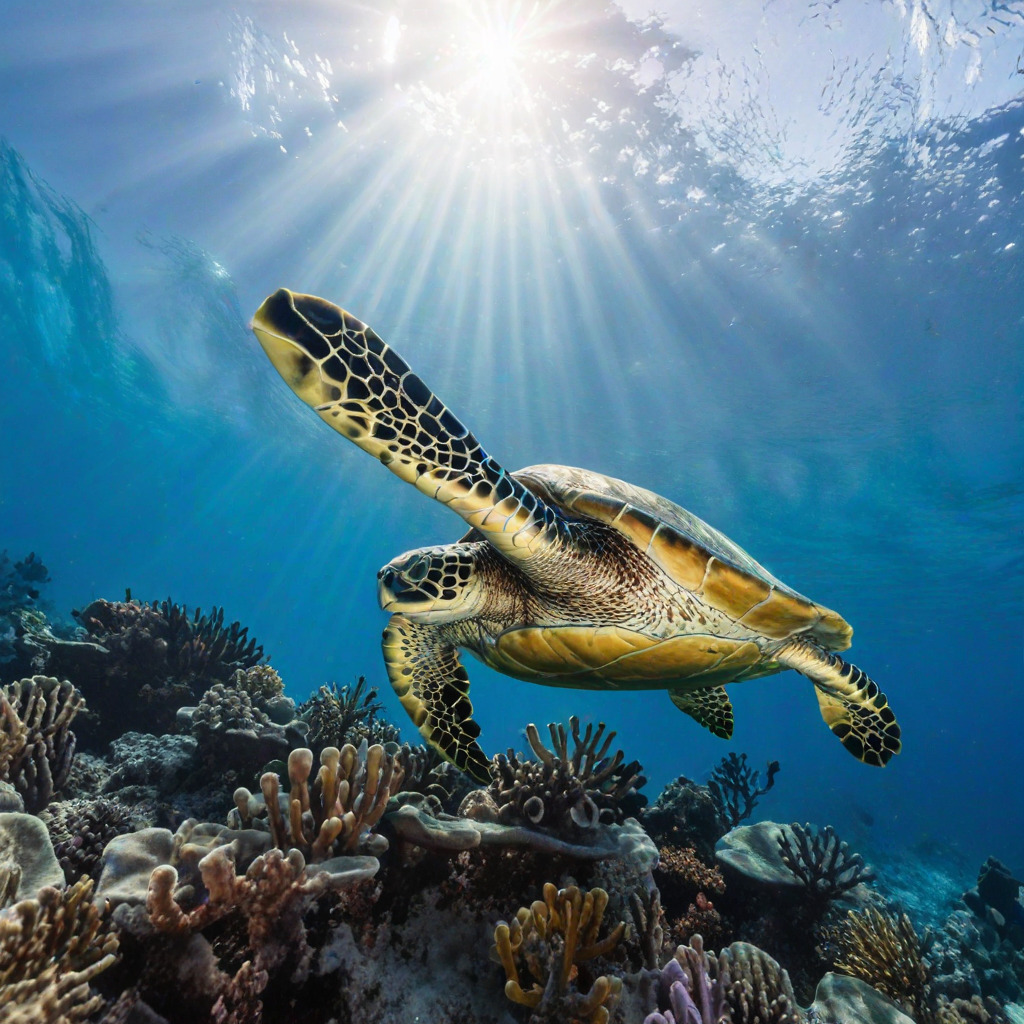} &
\includegraphics[width=\linewidth]{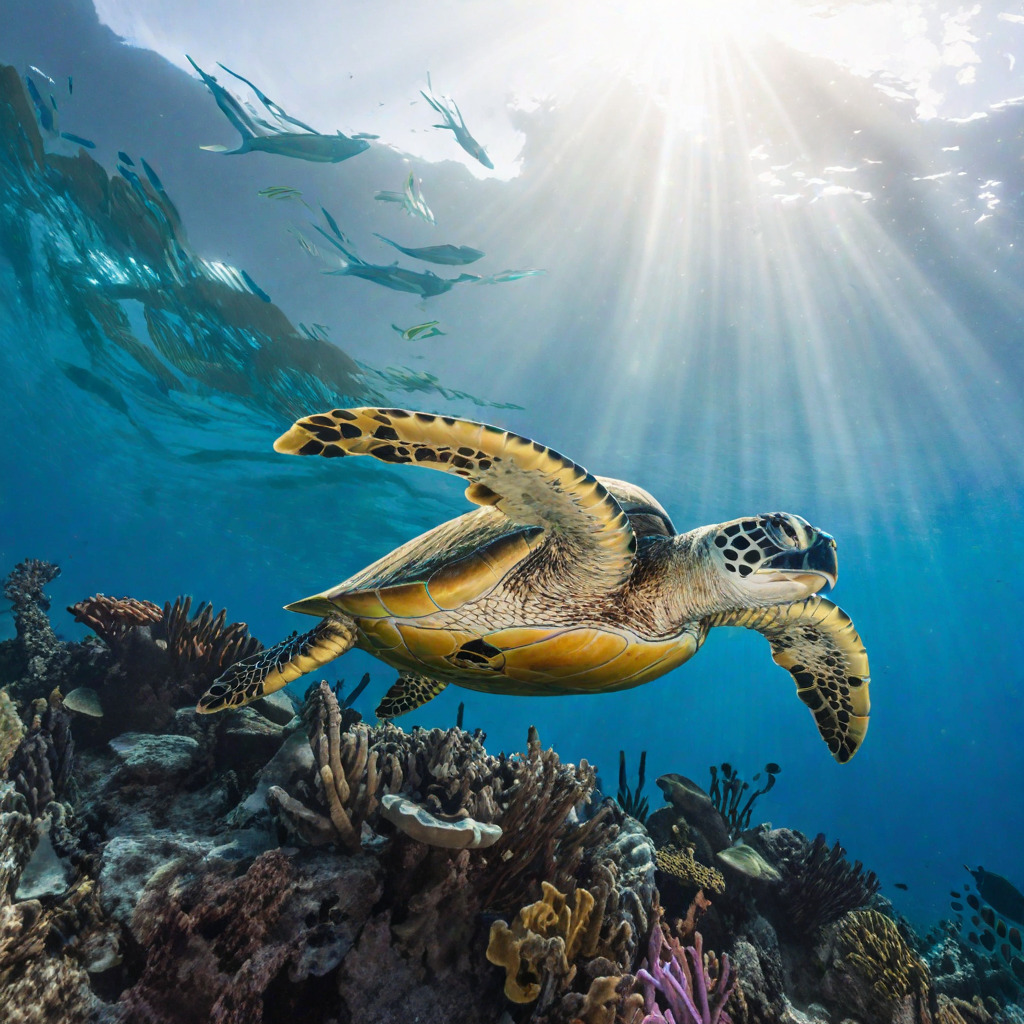} \\\noalign{\vskip 3pt}
\sdxlprompt{\textbf{An underwater scene} of a sea turtle gliding gracefully over a vibrant coral reef, sunbeams filtering through the water} \\[10pt]
\includegraphics[width=\linewidth]{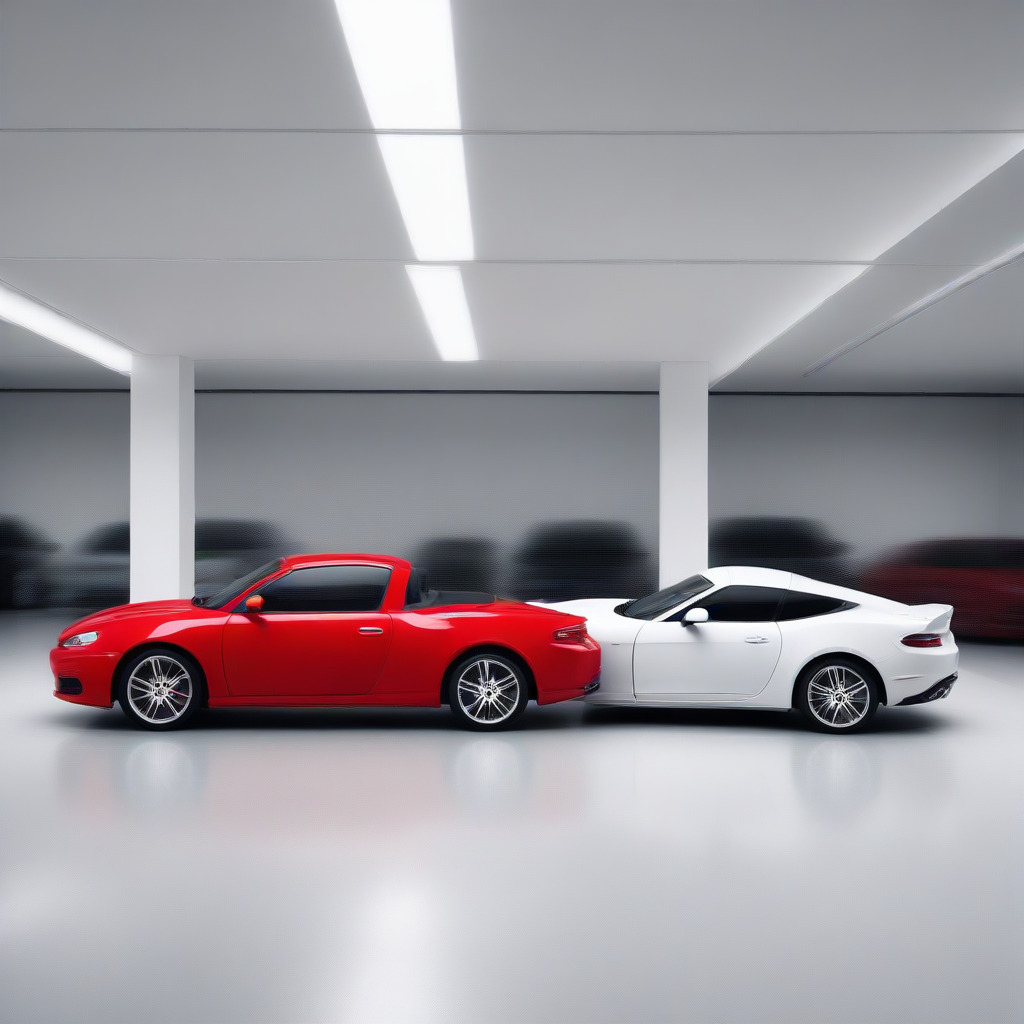} &
\includegraphics[width=\linewidth]{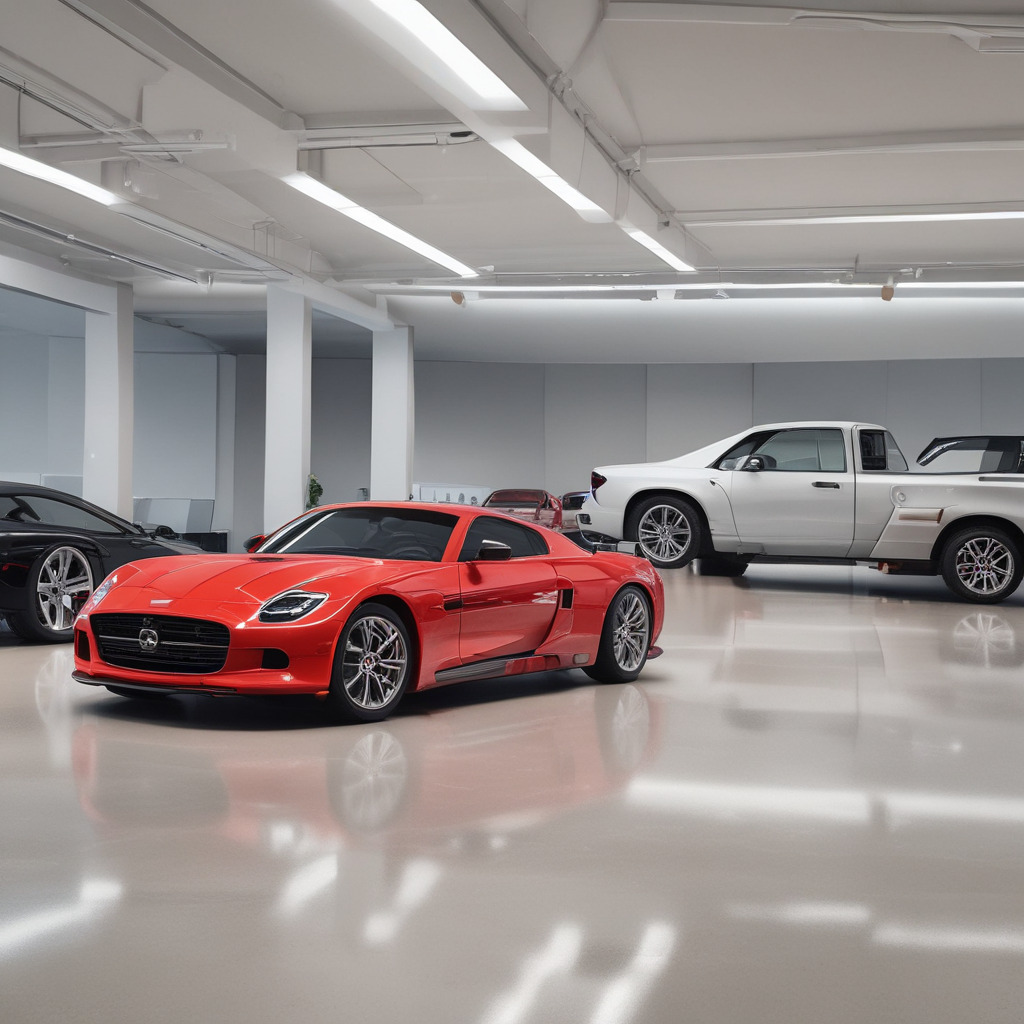} &
\includegraphics[width=\linewidth]{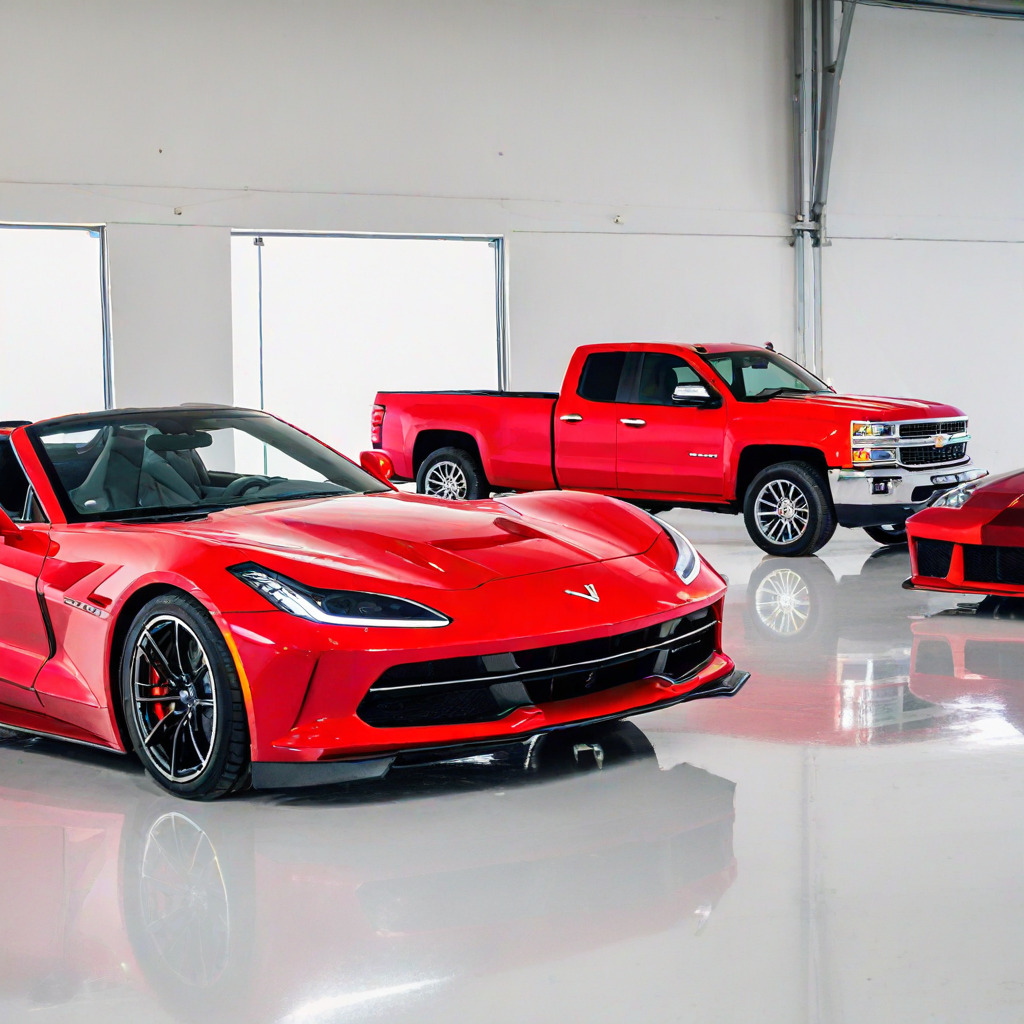} &
\includegraphics[width=\linewidth]{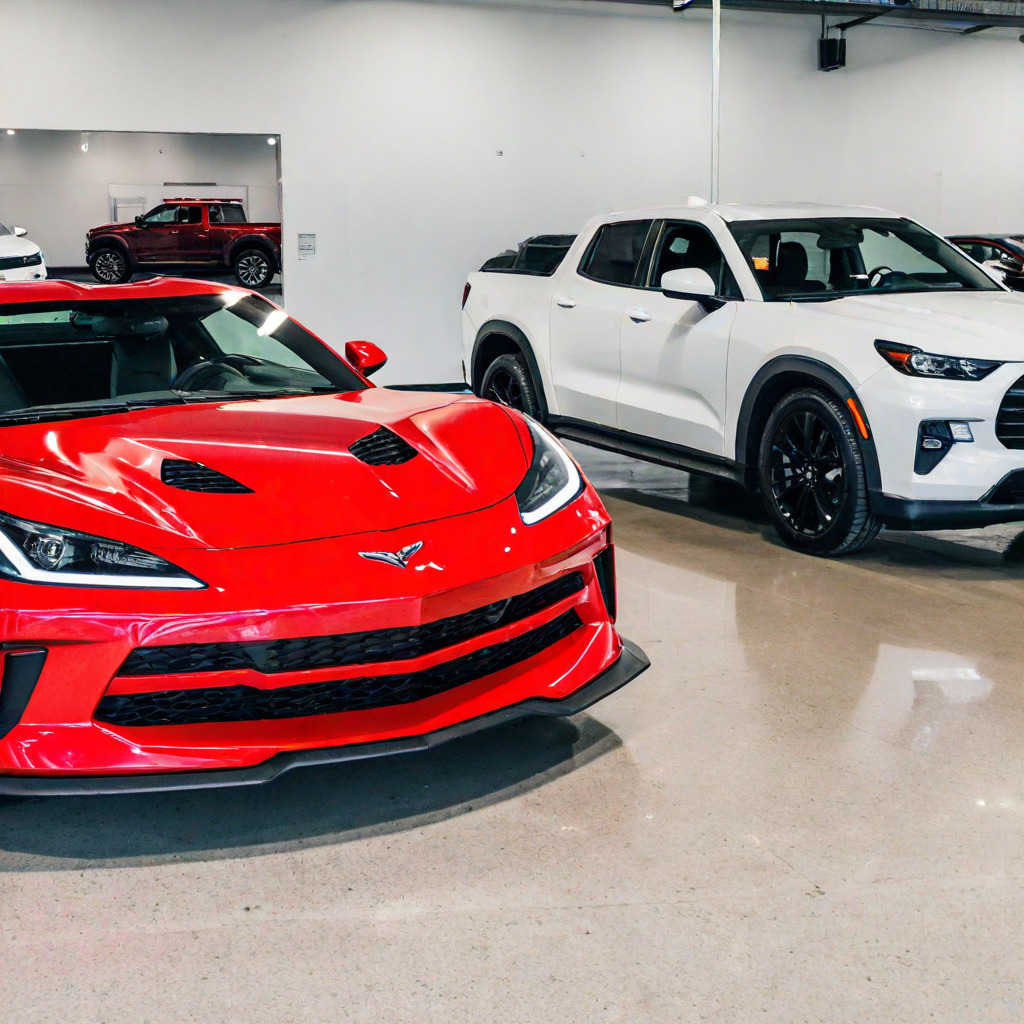} \\\noalign{\vskip 3pt}
\sdxlprompt{Two cars parked side by side in a showroom, the left one is a \textbf{red sports car}, the right one is a \textbf{white pickup truck}} \\
\end{tabular}
\par
\caption{\textbf{SDXL samples of 4-step students across distillation methods.} Each row shares a prompt and seed across PCM, SDXL-Lightning, DMD2, and its STAC continuation, with the prompt given below it. Compared with the other methods in this figure, DMD2 + STAC gives sharp images with coherent objects and faithful attributes. (Settings: 4 steps, $1024\times1024$, shared seed per prompt.)}
\label{fig:sdxl_qualitative_all}
\endgroup
\end{minipage}
\par

\subsection{Qualitative Results on Wan2.1}
\label{app:wan_visuals}

\begingroup
\newcommand{\wanframe}[1]{\includegraphics[width=.205\linewidth]{#1}}
\newcommand{\wanconfig}[3][15pt]{\raisebox{\dimexpr.05915\linewidth-.5\totalheight+#1\relax}[0pt][0pt]{\rotatebox[origin=c]{90}{\shortstack{\scriptsize #2 teacher\\[-1pt]\scriptsize #3 critic}}}}
\newcommand{\wanprompt}[1]{\par\parbox{.88\linewidth}{\centering\scriptsize #1\par}\par}
\newcommand{\wanchange}[1]{\raisebox{\dimexpr.05915\linewidth-.5\height\relax}{\makebox[3.4em]{\small #1}}}
\begin{center}
\begin{minipage}{\linewidth}
\centering
\makeatletter\def\@captype{figure}\makeatother
\setlength{\tabcolsep}{1.5pt}
\begin{tabular}{@{}cccccc@{}}
 & \small Start & \small $1/3$ duration & \small $2/3$ duration & \small End & \makebox[3.4em]{\shortstack{\small Frame\\\small change$\uparrow$}} \\
\wanconfig{Small}{and} &
\wanframe{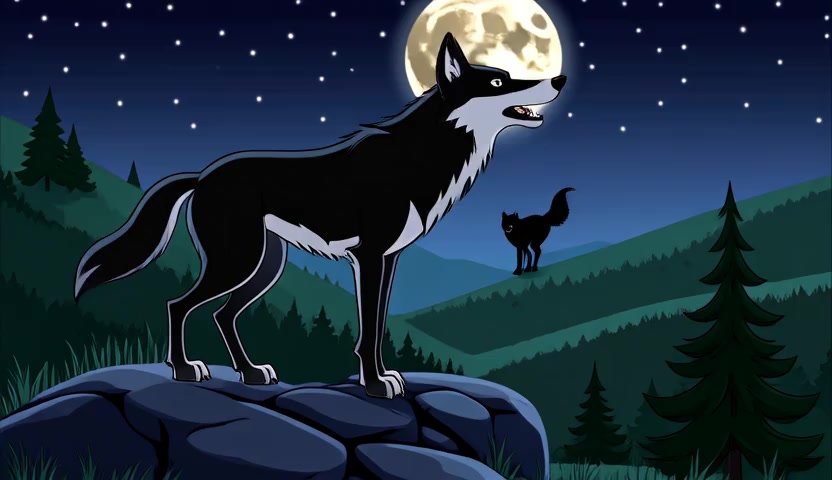} &
\wanframe{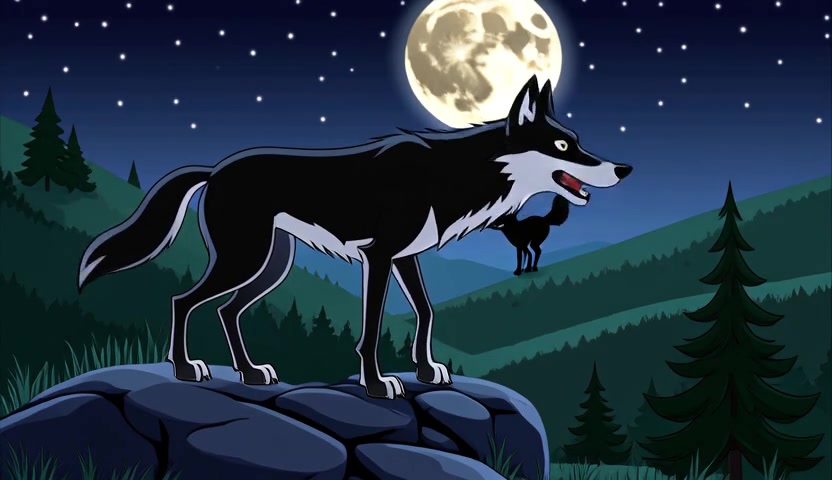} &
\wanframe{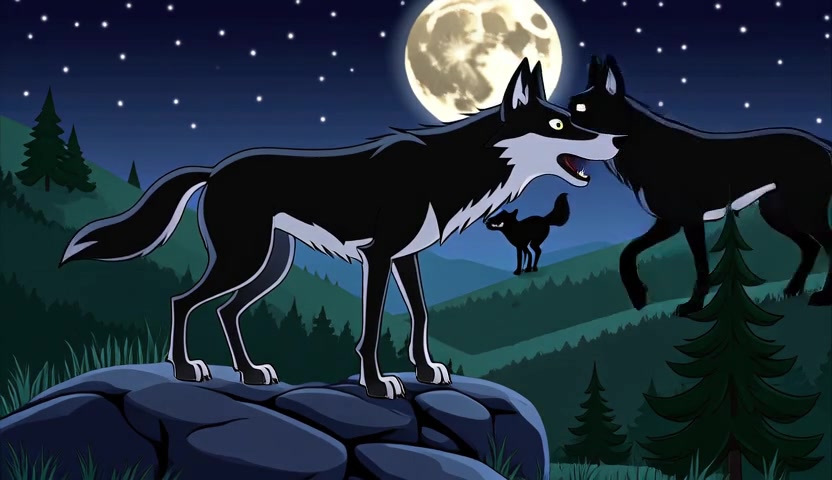} &
\wanframe{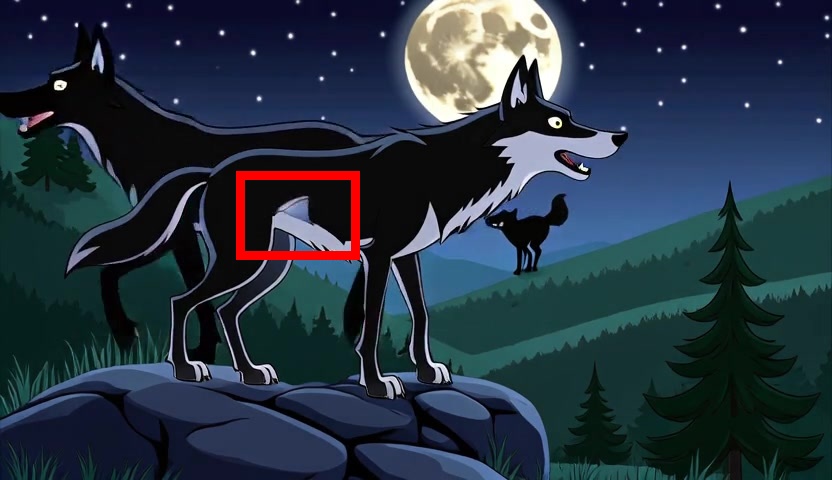} &
\wanchange{4.0} \\[3pt]
\wanconfig[11pt]{Big}{small} &
\wanframe{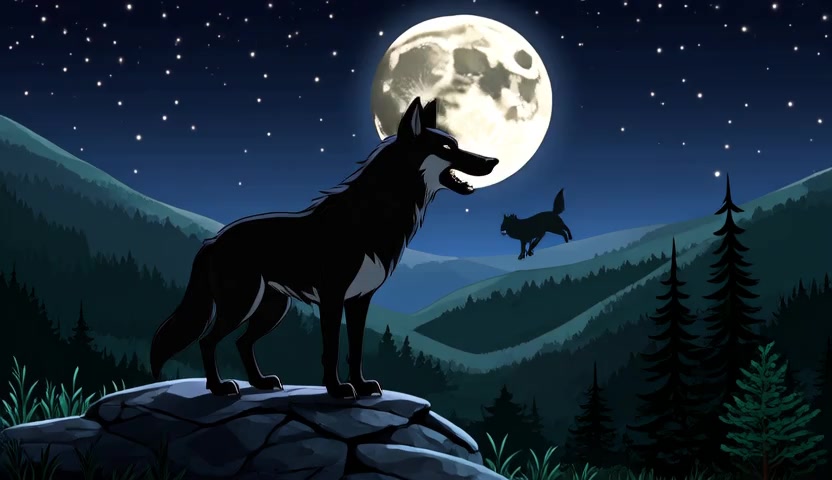} &
\wanframe{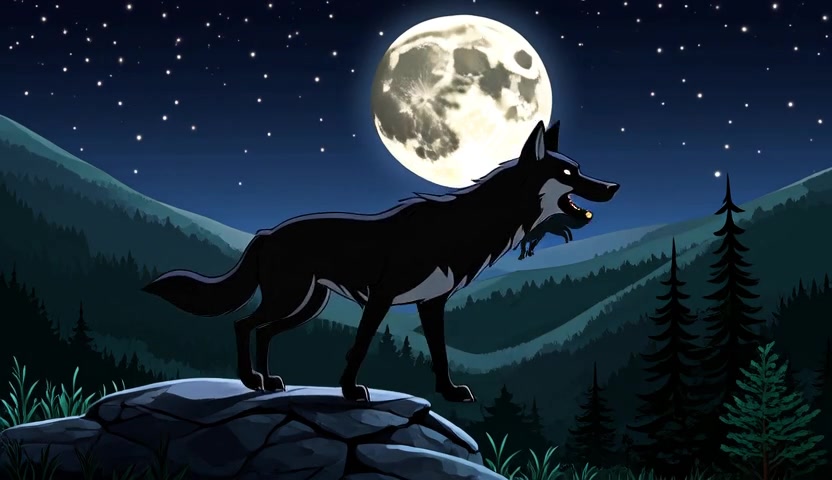} &
\wanframe{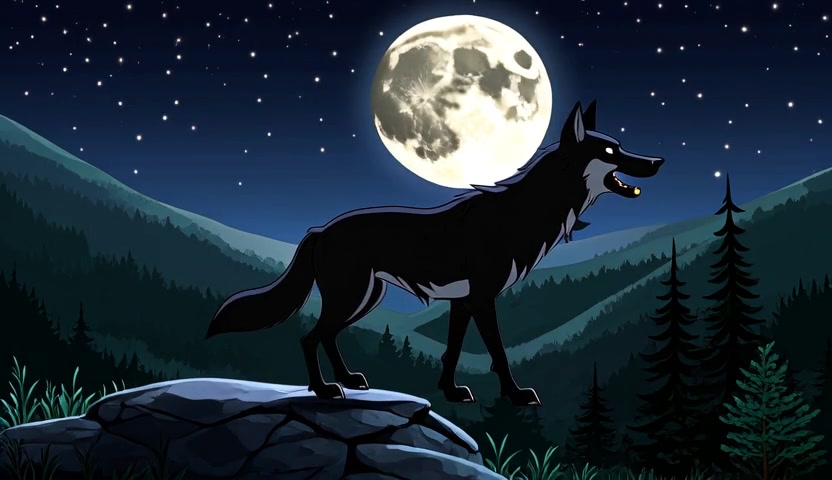} &
\wanframe{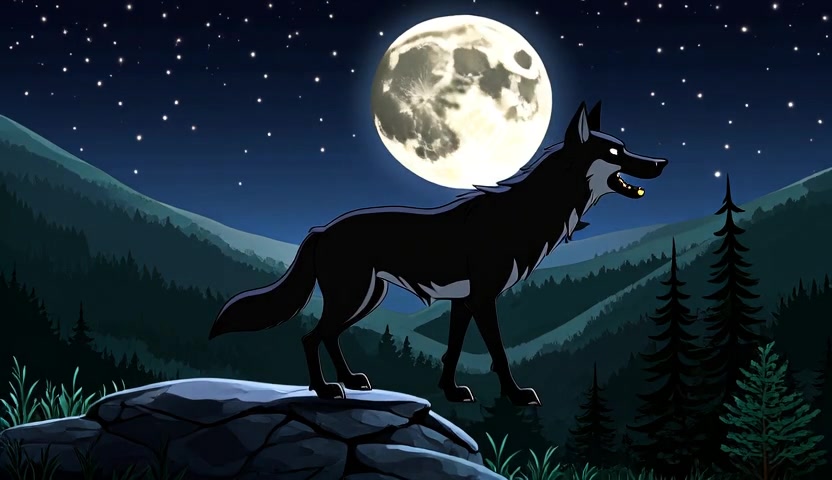} &
\wanchange{2.3} \\[3pt]
\wanconfig[11pt]{Big}{and} &
\wanframe{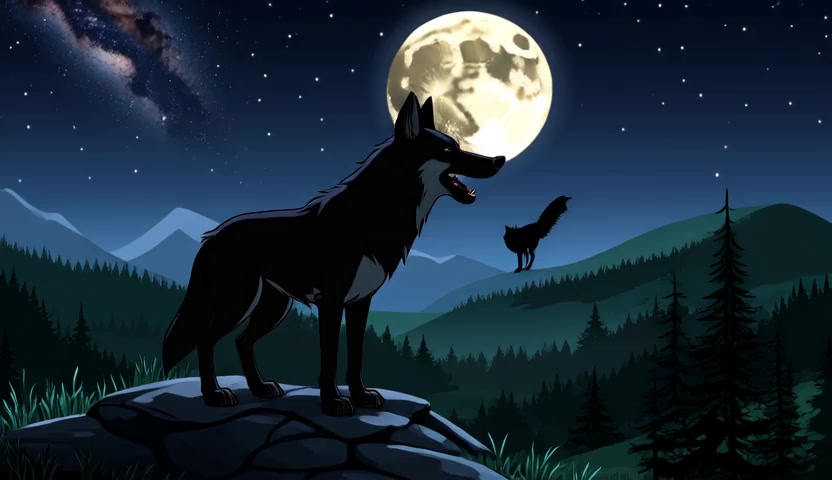} &
\wanframe{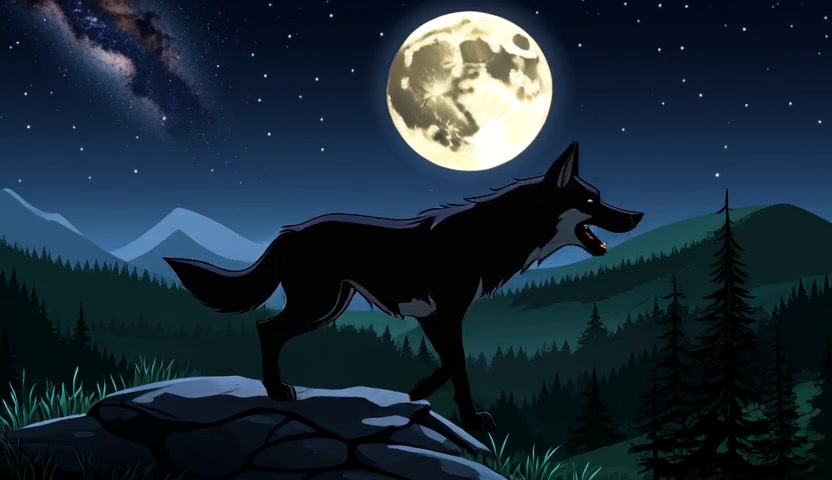} &
\wanframe{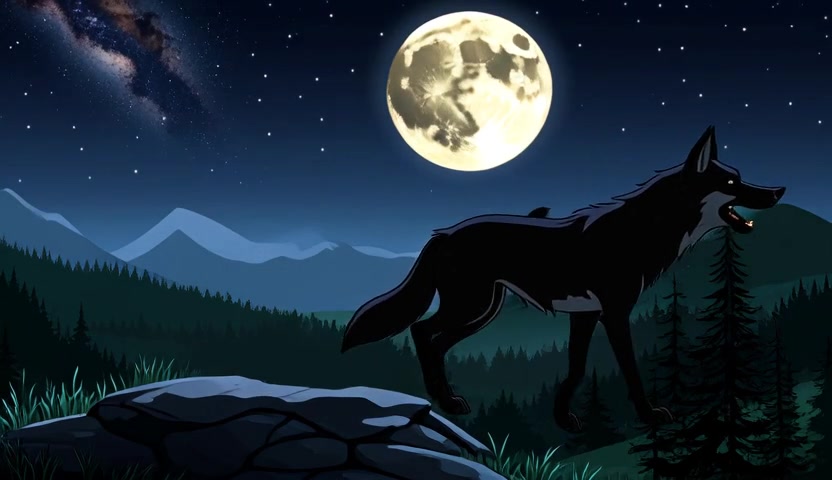} &
\wanframe{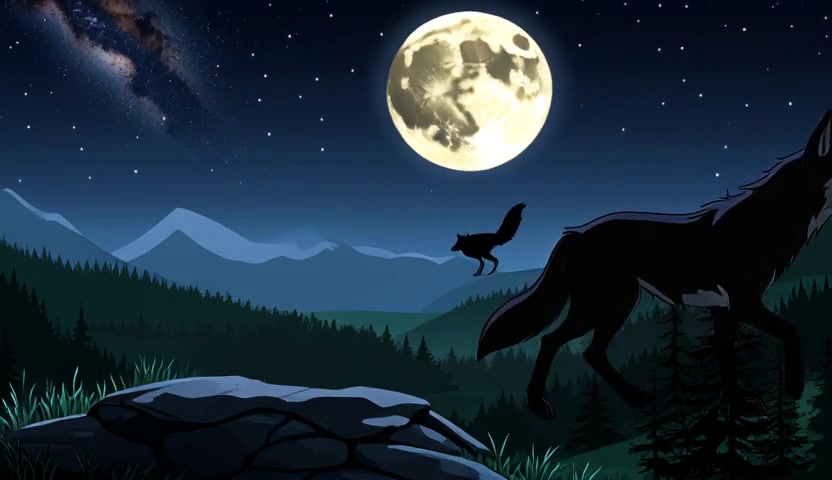} &
\wanchange{3.2} \\
\end{tabular}
\wanprompt{A beautiful silhouette animation shows a wolf howling at the moon, feeling lonely, until it finds its pack.}
\vspace{10pt}

\begin{tabular}{@{}cccccc@{}}
\wanconfig{Small}{and} &
\wanframe{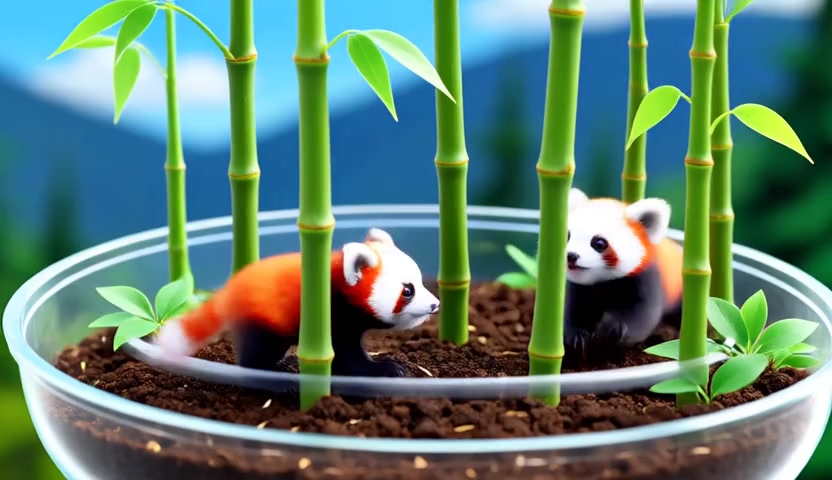} &
\wanframe{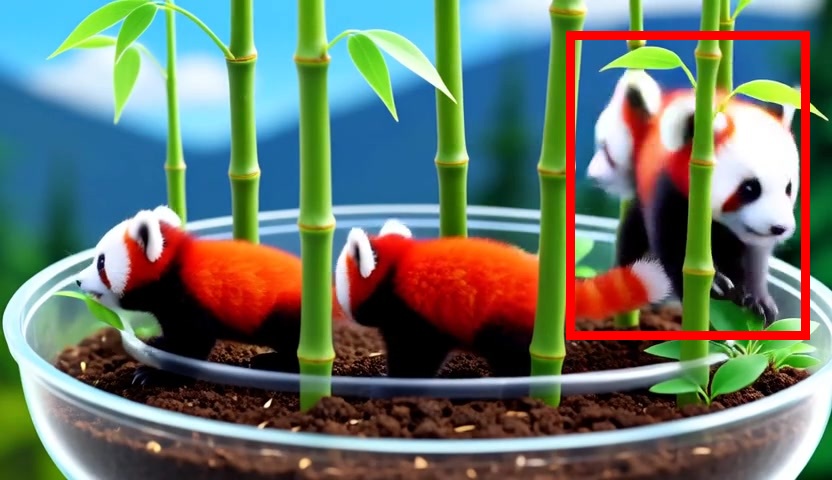} &
\wanframe{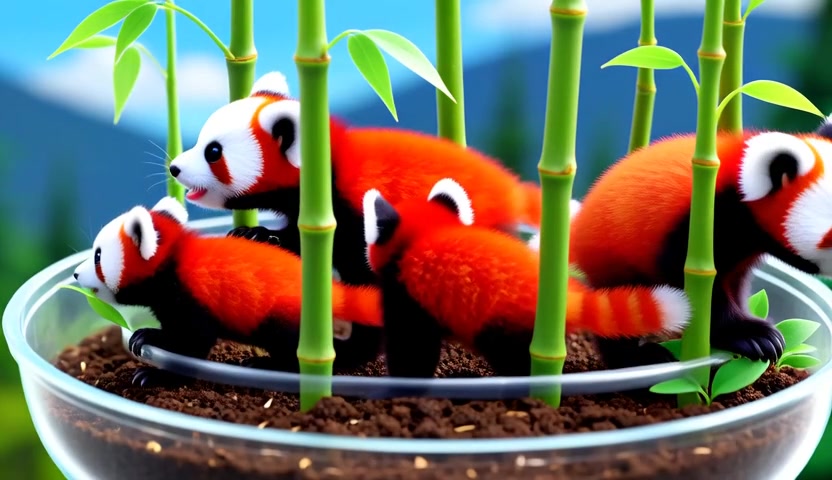} &
\wanframe{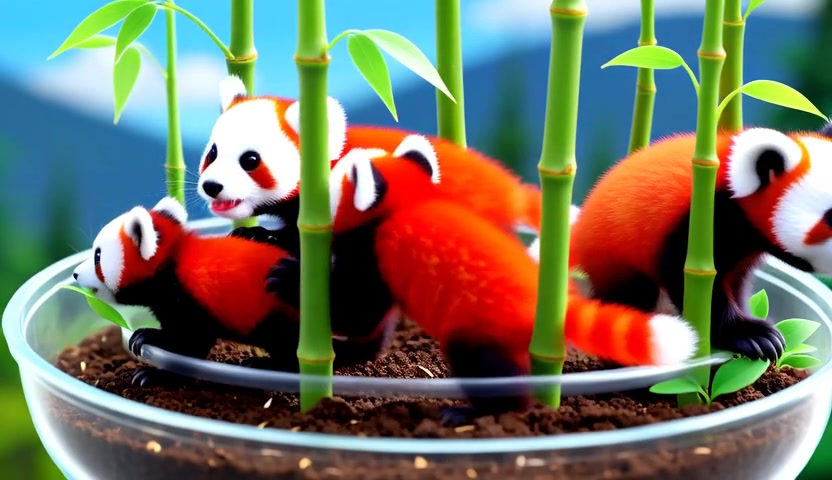} &
\wanchange{8.9} \\[3pt]
\wanconfig[11pt]{Big}{small} &
\wanframe{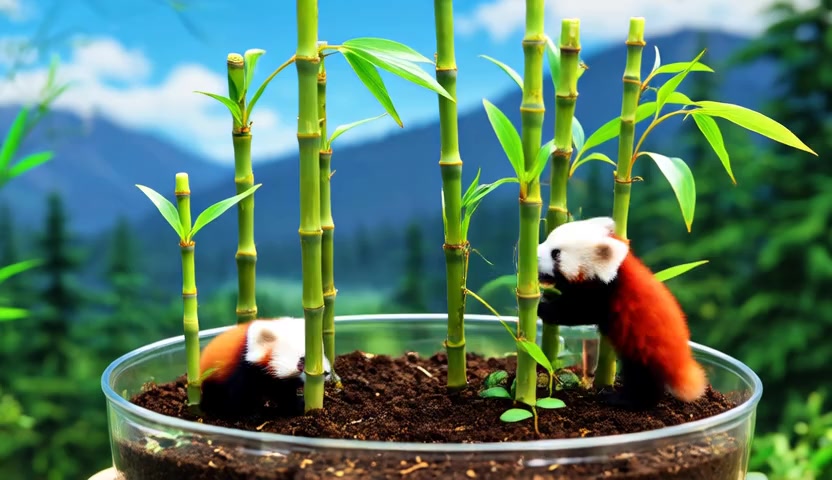} &
\wanframe{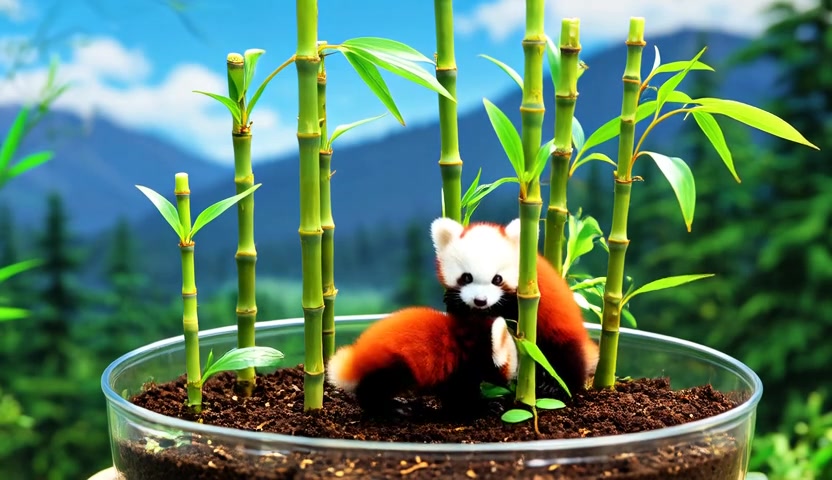} &
\wanframe{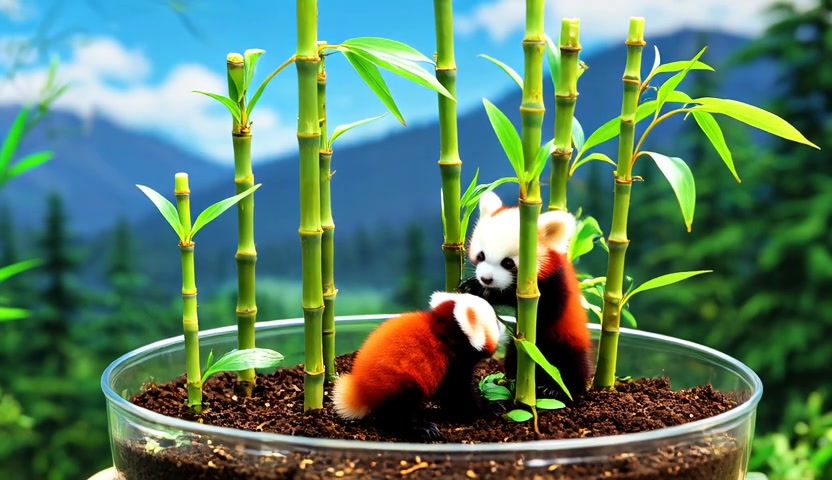} &
\wanframe{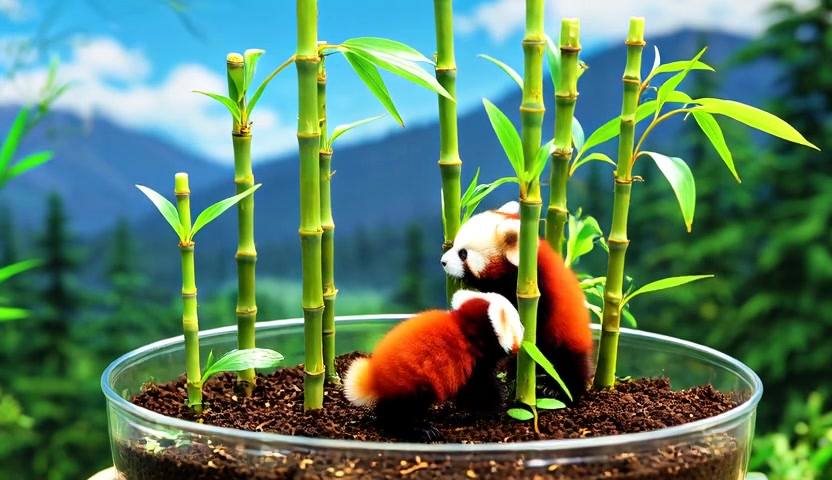} &
\wanchange{3.9} \\[3pt]
\wanconfig[11pt]{Big}{and} &
\wanframe{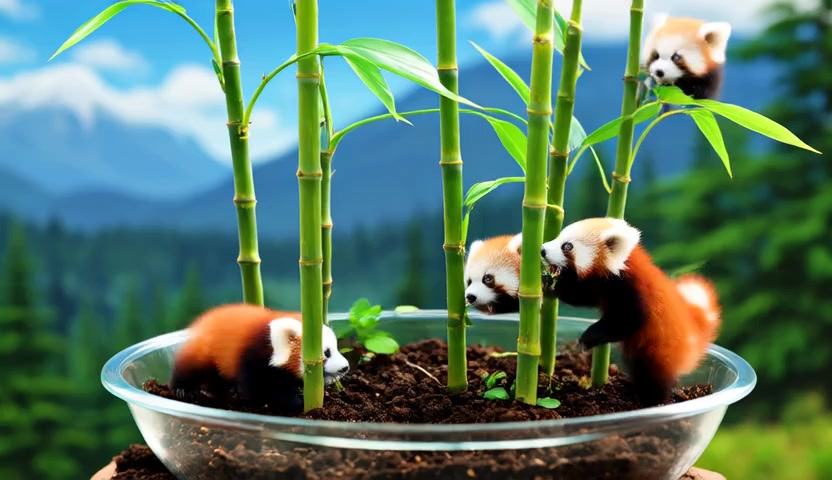} &
\wanframe{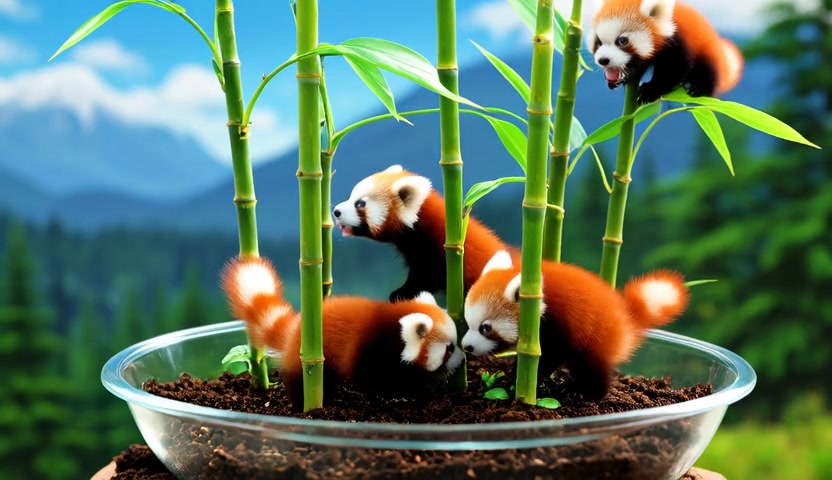} &
\wanframe{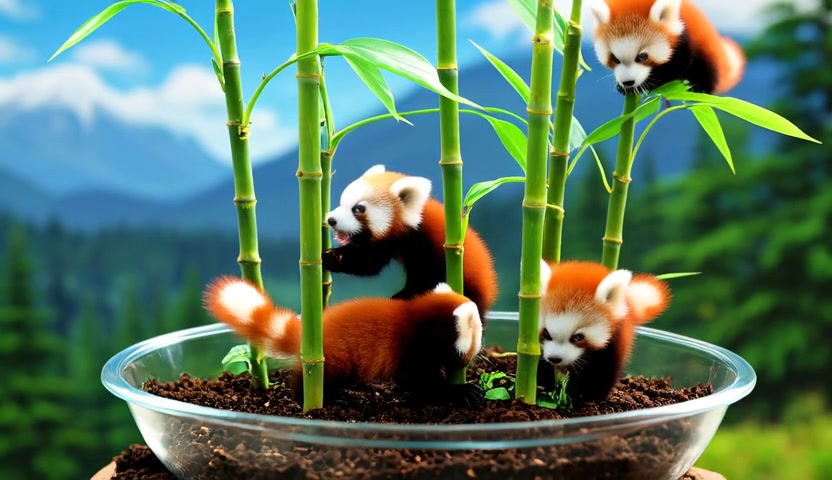} &
\wanframe{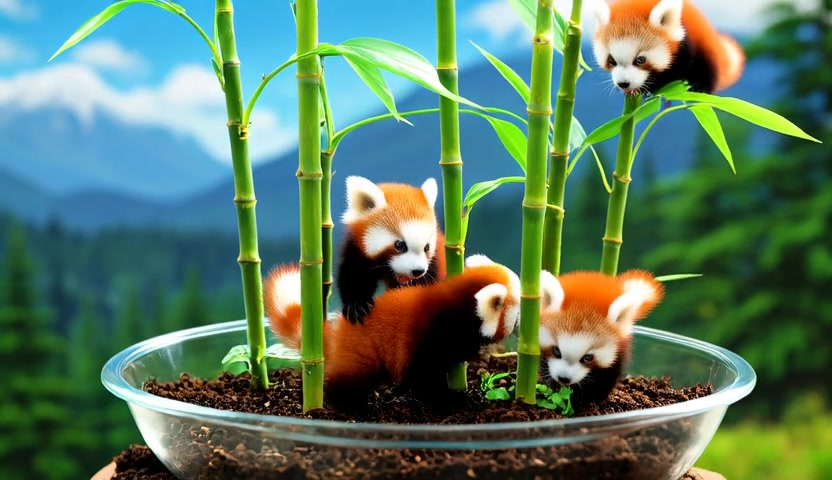} &
\wanchange{5.4} \\
\end{tabular}
\wanprompt{A petri dish with a bamboo forest growing within it that has tiny red pandas running around.}
\vspace{3.5pt}
\caption{\textbf{Wan2.1 samples of 4-step 1.3B students across teacher/critic pairs.} Each row shows four frames of one video, and the three rows of a block share the prompt given below them. \emph{Big} and \emph{small} denote the 14B and 1.3B models. Frame change is the mean absolute luminance difference between consecutive shown frames, expressed as a percentage of the intensity range; higher values indicate greater temporal change, which may reflect either motion or artifacts. Red boxes mark artifacts. Relative to the big-teacher/small-critic pair, matching the critic to the big teacher recovers part of the temporal change and yields more plausible videos. (Settings: autoregressive Wan2.1-T2V-1.3B students with the Self Forcing rollout, 4 steps, 480p; teacher/critic pairs of Table~\ref{tab:wan_main}.)}
\label{fig:wan_qualitative}
\end{minipage}
\end{center}
\endgroup

\end{document}